\documentclass[11pt]{youtu}

\usepackage{mathpazo}
\usepackage{XCharter}
\usepackage[scaled=1.1]{zlmtt}
\usepackage{graphicx}
\usepackage[numbers,sort]{natbib}
\usepackage{textcomp}
\usepackage{tikz}
\usepackage{gensymb}
\usepackage[utf8]{inputenc}
\usepackage[T1]{fontenc}
\usepackage{adjustbox}
\usepackage{siunitx}
\usepackage{amsmath}
\usepackage{pifont}
\usepackage{breakurl}
\usepackage{fontawesome5}
\usepackage[toc,page]{appendix}
\usepackage{tocloft}
\usepackage{marvosym}
\usepackage{booktabs}
\usepackage{multirow}
\usepackage{supertabular}
\usepackage{setspace}
\usepackage{multicol}
\usepackage{array}
\usepackage{longtable}
\usepackage{ragged2e}
\usepackage{makecell}
\usepackage{float}
\usepackage[edges]{forest}
\usepackage{subcaption}
\usepackage{soul}
\usepackage{tabularx}
\usepackage{enumitem}
\usepackage{amssymb}
\usepackage{wrapfig}
\usepackage{courier}
\usepackage{bxcoloremoji}
\usepackage{CJKutf8}
\usepackage{threeparttable}

\tcbuselibrary{most}

\definecolor{TitlePurpleStart}{RGB}{55, 24, 190}
\definecolor{TitlePurpleMid}{RGB}{92, 48, 220}
\definecolor{TitlePurpleEnd}{RGB}{142, 73, 245}

\newcommand{\papertitle}{From Chatbot to Digital Colleague:\\ The Paradigm Shift Toward Persistent Autonomous AI}
\fancypagestyle{firstpage}{
  \fancyfoot[C]{}
}

\def\method{OpenClaw}

\definecolor{hidden-red}{RGB}{205, 44, 36}
\definecolor{hidden-blue}{RGB}{194,232,247}
\definecolor{hidden-orange}{RGB}{243,202,120}
\definecolor{hidden-green}{RGB}{34,139,34}
\definecolor{hidden-pink}{RGB}{255,245,247}
\definecolor{hidden-black}{RGB}{20,68,106}
\definecolor{purple}{RGB}{144,153,196}
\definecolor{yellow}{RGB}{255,228,123}
\definecolor{hidden-yellow}{RGB}{255,248,203}
\definecolor{tkcolor}{RGB}{224,223,255}
\definecolor{darkblue}{rgb}{0, 0.40, 0.75}
\definecolor{FigureBackground}{RGB}{237,245,254}
\colorlet{youtuLightBlue}{FigureBackground}
\definecolor{AIboxBackground}{RGB}{238,245,253}
\definecolor{AIboxFrame}{RGB}{54,85,140}

\definecolor{blue}{RGB}{55,83,156}

\definecolor{myblue}{RGB}{215,226,240}
\definecolor{lightcream}{RGB}{252,249,242}

\hypersetup{colorlinks=true, citecolor=darkblue, linkcolor=darkblue, urlcolor=darkblue}

\tcbset{
  aibox/.style={
    width=\linewidth,
    top=8pt,
    bottom=4pt,
    colback=AIboxBackground,
    colframe=AIboxFrame,
    colbacktitle=AIboxFrame,
    enhanced,
    center,
    attach boxed title to top left={yshift=-0.1in,xshift=0.15in},
    boxed title style={boxrule=0pt,colframe=AIboxFrame,},
  }
}

\tikzstyle{my-box}=[
rectangle,
draw=hidden-black,
rounded corners,
text opacity=1,
minimum height=1.5em,
minimum width=5em,
inner sep=2pt,
align=center,
fill opacity=.5,
]
\tikzstyle{leaf}=[my-box, minimum height=1.5em, fill=hidden-red!20, text=black, align=left, font=\normalsize, inner xsep=5pt, inner ysep=4pt]
\tikzstyle{leaf2}=[my-box, minimum height=1.5em, fill=hidden-green!20, text=black, align=left, font=\normalsize, inner xsep=5pt, inner ysep=4pt]
\tikzstyle{leaf3}=[my-box, minimum height=1.5em, fill=yellow!32, text=black, align=left, font=\normalsize, inner xsep=5pt, inner ysep=4pt, text width=45em]
\tikzstyle{leaf4}=[my-box, minimum height=1.5em, fill=hidden-blue!57, text=black, align=left, font=\normalsize, inner xsep=5pt, inner ysep=4pt]
\tikzstyle{leaf5}=[my-box, minimum height=1.5em, fill=darkblue!15, text=black, align=left, font=\normalsize, inner xsep=5pt, inner ysep=4pt]
\tikzstyle{leaf6}=[my-box, minimum height=1.5em, fill=purple!30, text=black, align=left, font=\normalsize, inner xsep=5pt, inner ysep=4pt]

\newcommand{\AIboxTitleIcon}{\faIcon[regular]{lightbulb}}
\newtcolorbox{AIbox}[2][]{aibox,title={\AIboxTitleIcon~#2},#1}

\tcbset{
  takeawaybox/.style={
    width=\linewidth,
    top=8pt,
    bottom=4pt,
    colback=hidden-yellow,
    colframe=black,
    colbacktitle=black,
    enhanced,
    center,
    attach boxed title to top left={yshift=-0.1in,xshift=0.15in},
    boxed title style={boxrule=0pt,colframe=white,},
  },
  titlebox/.style={
    enhanced,
    colback=FigureBackground,
    colframe=FigureBackground,
    boxrule=0pt,
    arc=0pt,
    outer arc=0pt,
    left=0pt,
    right=0pt,
    top=0pt,
    bottom=0pt,
    before skip=0pt,
    after skip=0pt,
  },
  firstpagebluefill/.style={
    enhanced,
    frame hidden,
    colback=FigureBackground,
    colframe=FigureBackground,
    boxrule=0pt,
    arc=10pt,
    outer arc=10pt,
    sharp corners=north,
    left=0.5cm,
    right=0.5cm,
    top=0pt,
    bottom=0.5cm,
    before skip=0pt,
    after skip=0pt,
  }
}

\newtcolorbox{TakeawayBox}[2][]{takeawaybox,title=#2,#1}

\newcommand{\headingnote}[1]{
  \vspace{-0.65em}
  \par\noindent\hspace{1.8em}{\small\itshape #1}\par
  \vspace{0.25em}
}

\title{\includegraphics[height=1.5em]{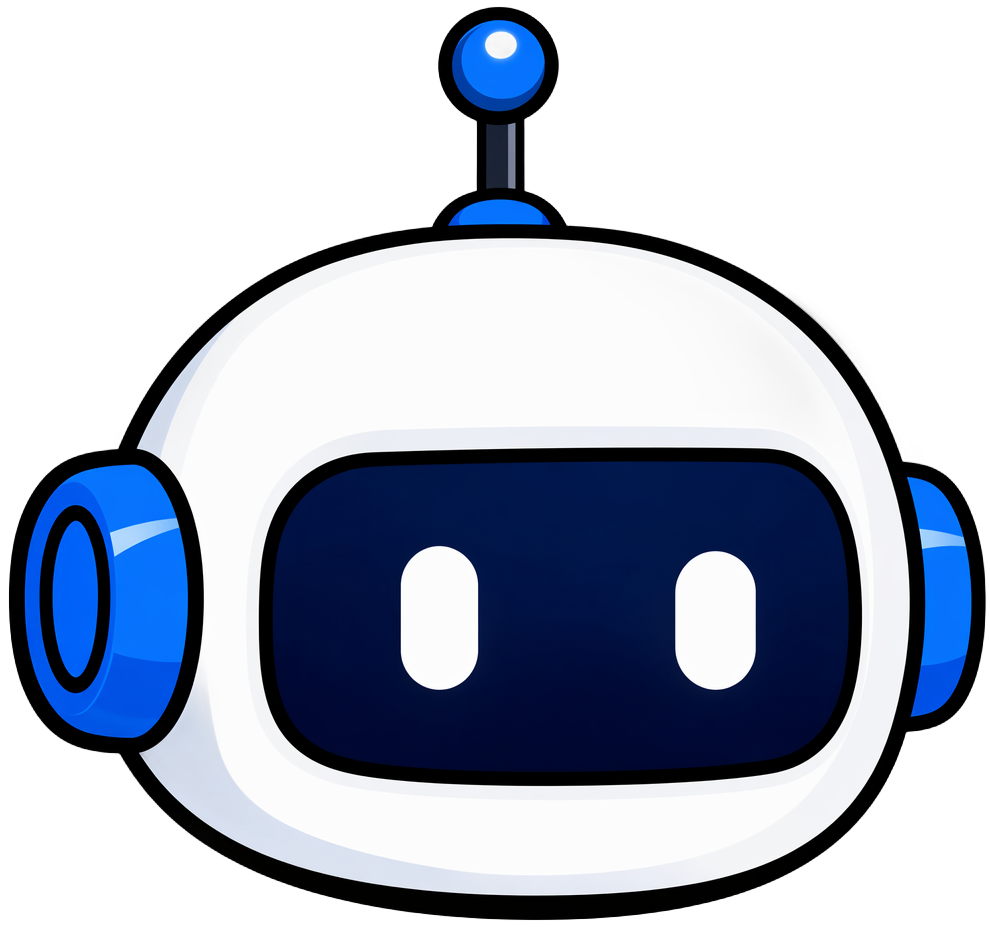}\hspace{0.4em}\textit{\papertitle}}
\renewcommand{\authorlist}{
  \begin{center}
    {\tencentsanswseven{Yongheng Zhang$^{*}$, Ziang Liu$^{*}$, Jiaxuan Zhu$^{*}$, Shuai Wang, Xiangqi Chen, Haojing Huang, Jiayi Kuang, Siyu Chen, Ao Shen, Hao Wu, Qiufeng Wang, Qian-Wen Zhang, Junnan Dong, \\ Wenhao Jiang, Ying Shen, Hai-Tao Zheng, Yinghui Li$^{\dagger}$, Di Yin, Xing Sun, Philip S. Yu}}
  \end{center}
}
\renewcommand{\affiliationlist}{
  \begin{center}
    {\small\tencentsanswseven{Tencent Youtu Lab\quad Tsinghua University\quad \\ Sun Yat-sen University\quad Central South University\quad University of Illinois at Chicago}}
  \end{center}
}

\begin{document}

\abstract{Large Language Models (LLMs) are undergoing a fundamental transformation from conversational generators into integrated AI systems capable of reasoning, action, memory, and self-improvement. We conceptualize this transition as a shift \textbf{from Chatbot to Digital Colleague}: from conversational answers to persistent work. We organize this transition along two tightly coupled dimensions. First, at the \textbf{cognitive core} level, LLMs are advancing from Chatbot-era ``fast thinking'' systems driven by next-token prediction toward Thinking LLMs that leverage inference-time computation, Chain-of-Thought reasoning, reflection, process supervision, and reinforcement learning to support more deliberate and reliable cognition. Second, at the \textbf{tool-augmented task execution} level, LLMs are progressing from tool-calling Agents that invoke external resources in an ad hoc manner toward OpenClaw-style workstation systems (\method{}) equipped with persistent Workspaces, skills, verification loops, and governance. The \textbf{``Workspace + Skill''} paradigm makes episodic tool use colleague-like via state persistence, reusable procedures, task closure, and experience reuse. We examine \textbf{data construction} shifts from instruction-response pairs to State-Action-Observation trajectories and \textbf{evaluation} from static benchmarks to sandboxed, auditable, self-evolving AI ecosystems.}
\maketitle
\thispagestyle{firstpage}
\noindent\hspace*{-0.7pt}\includegraphics[width=\dimexpr\linewidth+1.5pt\relax]{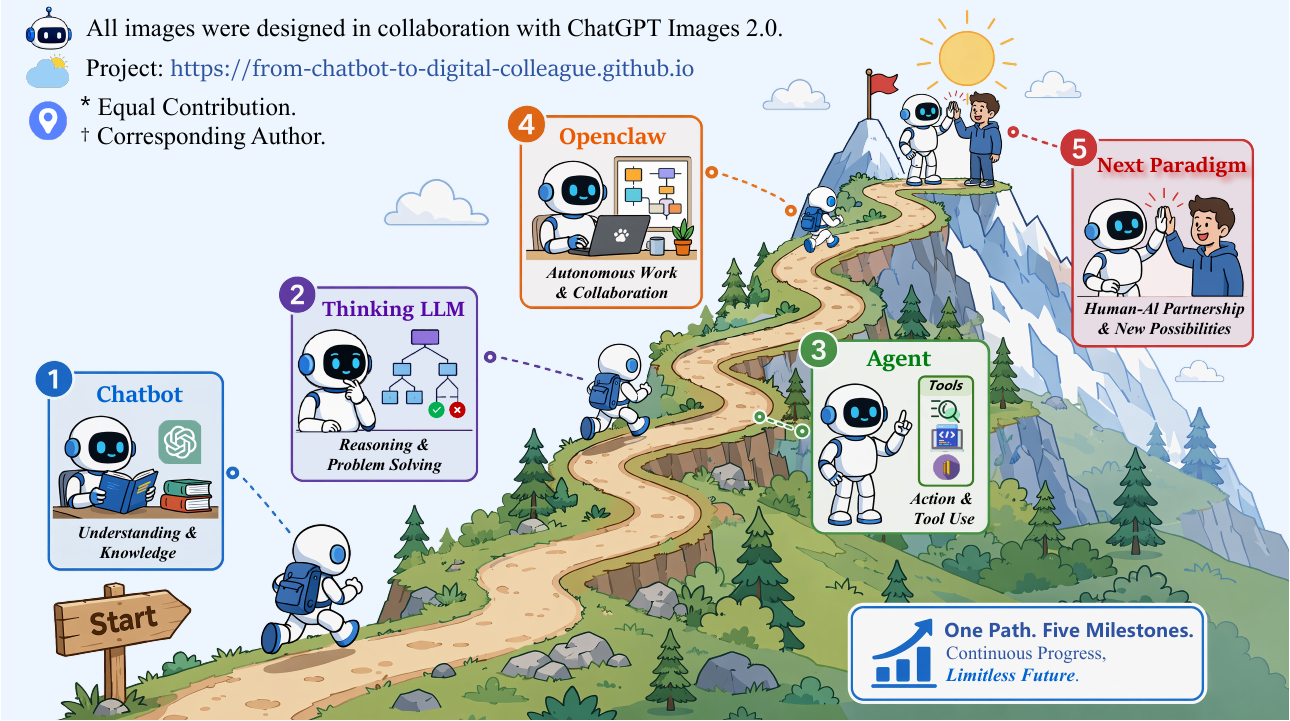}
\newpage

{\hypersetup{linkcolor=black}
\setstretch{1.9}
\tableofcontents}
\newpage

\begin{figure}[t]
    \centering
    \includegraphics[width=\textwidth]{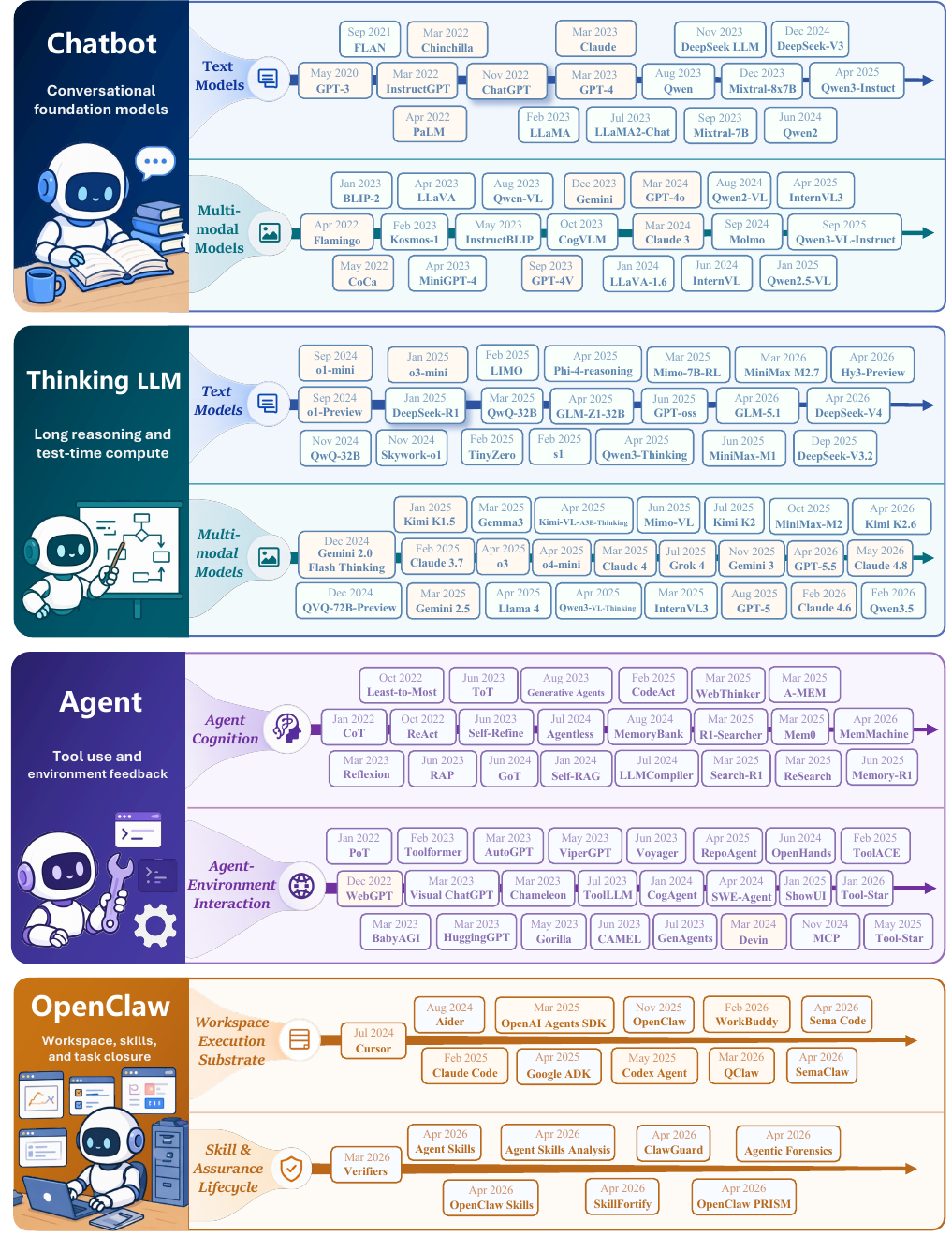}
    \vspace{-8mm}
    \caption{A roadmap and evolutionary timeline of next-generation LLM systems. The figure summarizes how these AI systems progress from simple conversational chatbots to reasoning cores, tool-using agents, and persistent workspace systems over time. Each node is labeled by its release month. \raisebox{0.7mm}{\fcolorbox{blue}{myblue}{\textcolor{myblue}{\rule{0.7mm}{0.7mm}}}} box represents open-source / open platform; \raisebox{0.7mm}{\fcolorbox{blue}{lightcream}{\textcolor{lightcream}{\rule{0.7mm}{0.7mm}}}} box represents closed / commercial system.}
    \label{fig:map}
\end{figure}

\clearpage

\section{Introduction}
\label{sec:introduction}

Large Language Models (LLMs) are undergoing a fundamental transformation~\citep{bommasani2021opportunities,min2023recent,naveed2025comprehensive,wang2024survey,xi2023rise,qin2025largelanguagemodelsmeet}. What began as statistical language generation has expanded into AI systems that can reason, act, remember, and complete tasks in open-ended digital environments~\citep{yao2022react,park2023generative,zhang2025survey}. Early progress was driven by scaling autoregressive Transformers and instruction-aligned chat interfaces, enabling systems to compress broad world knowledge into fluent responses~\citep{vaswani2017attention,brown2020language,kaplan2020scaling,hoffmann2022chinchilla,achiam2023gpt}. More recently, the frontier has shifted toward models that deliberate over difficult problems, invoke tools, interact with environments, and coordinate multi-step workflows~\citep{wei2022chain,deepseek2025r1,schick2023toolformer,qin2023toolllm,wu2023autogen,hong2023metagpt}. The central question is therefore \textit{\textbf{no longer limited to}} \textit{how can a model generate a better answer?} \textit{\textbf{Instead}}, it is how \textit{how can an AI system reliably transform user intent into completed work?} This redefines the human-AI relationship, marking the shift \textbf{from Chatbot to Digital Colleague}~\citep{openai2022chatgpt,bigscience2022bloom,google2024gemma2,jiang2023mistral7b,chen2023palix}.

This survey organizes and analyzes persistent autonomous AI along two tightly coupled dimensions. 
\ding{172} \textit{\textbf{The first dimension concerns the cognitive core}}. It explores how models generate, understand, and reason, spanning two eras, shown in Figure~\ref{fig:map}'s Chatbot and Thinking LLM panels. In the \textit{Chatbot Era}, LLMs behave like fast ``System-1'' generators: they compress parametric knowledge and produce fluent responses, but struggle with deep reasoning, verification, and long-horizon consistency~\citep{petroni2019language,huang2023towards,dziri2023faith,valmeekam2022large}. In the \textit{Thinking LLM Era}, models increasingly leverage inference-time computation, Chain-of-Thought prompting, reflection, process supervision, and reinforcement learning to support slower, more deliberate, and more reliable problem-solving processes~\citep{snell2024scaling,kojima2022large,madaan2023selfrefine,shinn2023reflexion,lightman2024lets,shao2024deepseekmath,deepseek2025r1}. PartI traces the transition to slow, reasoning-centered cognition\citep{schneider2025generative,wei2026agentic,patil2025advancing,zhao2026edge,hu2024survey}.

\ding{173} \textit{\textbf{The second dimension concerns tool-augmented task execution}}. This dimension asks how a stronger cognitive core becomes a system that can act in dynamic and complex external environments, and it also contains two eras, as illustrated in the Agent and OpenClaw panels of Figure~\ref{fig:map}. In the \textit{Agent Era}, LLMs move from passive dialogue systems into active systems that call APIs, browse websites, write code, manipulate files, and collaborate with other agents~\citep{schick2023toolformer,qin2023toolllm,yao2022react,wu2023autogen,hong2023metagpt,xi2023rise,wang2024survey}. Yet, these early agents still remain highly fragile: incorrect action formats, missing observations, failed tool calls, or unrecovered intermediate errors can derail the entire trajectory. In the \textit{OpenClaw Era}, tool use is embedded into persistent workspaces with files, terminals, browsers, logs, permissions, reusable skills, and verification procedures, enabling agents to maintain context, monitor progress, recover from failures, and verify final workspace states~\citep{xu2025toward,guo2024large,lei2025large,sun2025survey,plaat2025agentic}.

Within this two-dimensional framework, the key thesis of this survey is that \textbf{Workspace + Skill} provides the mechanism that turns chatbot-style interaction into durable digital-colleague work~\citep{wang2024openhands,yang2024sweagent,drouin2024workarena}. A \textbf{Workspace} is a persistent digital environment for AI operations, including files, terminals, browsers, editors, repositories, calendars, documents, databases, and domain-specific applications~\citep{xie2024osworld,zhou2023webarena,drouin2024workarena,wang2024openhands}. A \textbf{Skill} is a reusable, parameterizable procedure for completing tasks, including planning, tool sequencing, intermediate checks, error recovery, and validation~\citep{wang2023voyager,shinn2023reflexion,madaan2023selfrefine}. Together, they move LLMs beyond episodic responses and atomic tool calls: the workspace provides state, memory, evidence, and consequences, while the skill provides reusable operational knowledge~\citep{schick2023toolformer,qin2023toolllm,wang2023voyager,wang2024openhands}.

This perspective reframes data and evaluation paradigms in LLM development. For chatbots, data is often organized as instruction-response pairs and evaluation measures final-answer correctness or human preference. For reasoning models, data includes long-form Chain-of-Thought traces, process supervision, and verifiable rewards, while evaluation expands toward reasoning-process judgment. For agents and workspace-based systems, the fundamental unit of learning becomes the \textbf{state--action--observation trajectory}, and evaluation shifts from answer quality to task closure—whether the system reaches the intended final state under reproducible, auditable, and safe conditions~\citep{luo2025large,maestre2024beyond,du2026survey,barua2024exploring,yao2025survey}.

Despite impressive progress, current systems face major structural bottlenecks~\citep{liu2023agentbench,zhou2023webarena,jimenez2023swebench,xie2024osworld}. Reasoning can remain ungrounded or hallucinated during factual verification~\citep{huang2023hallucination,ji2023hallucination}; long-horizon execution is brittle as errors accumulate across toolchains\citep{liu2023agentbench,zhou2023webarena,jimenez2023swebench}; memory and state management often depend on transient context windows\citep{packer2023memgpt,zhang2025survey}; and safety becomes harder when outputs are executable actions with side effects~\citep{ruan2024identifying,debenedetti2024agentdojo,zhan2024injecagent}. These challenges highlight that the transition from chatbot to digital colleague requires both stronger foundation models and better execution substrates, skill abstractions, evaluation environments, and governance mechanisms~\citep{wang2024openhands,yang2024sweagent,wang2023voyager,drouin2024workarena,li2026prism,zhao2026clawguard}.

Accordingly, this survey reviews the field through four parts. Part~I examines the evolution of the LLM cognitive core, from chatbot-era language generation to thinking LLMs driven by long reasoning chains and reinforcement-learning-based cognition. Part~II studies tool-augmented task execution, from early agents to OpenClaw-style systems oriented toward workspace intelligence, skill-based execution, reliability, and governance. Part~III explains why \textbf{Workspace + Skill} is a decisive leap from ephemeral interactions to persistent stateful work and from ad-hoc prompts to composable capability packages. Part~IV analyzes the accompanying shifts in data and evaluation, from knowledge corpora and instruction data to action trajectories, process verification, and task-closure-oriented benchmarks. We then discuss open challenges and future directions toward reliable, self-evolving AI ecosystems.

The main contributions of this survey are summarized as follows:
\begin{itemize}[leftmargin=*, itemsep=2pt, topsep=3pt]
    \item \textbf{A two-dimensional view of AI:} We organize this evolution along two complementary dimensions: cognitive-core evolution (Chatbot and Thinking LLM) and tool-augmented task execution (Agent and OpenClaw), with Workspace + Skill as the mechanism for the Chatbot to Digital Colleague.
    \item \textbf{A unified account of Workspace + Skill, data, and evaluation:} We identify workspace persistence and reusable skills as mechanisms for task completion, and connect this shift with the move from instruction-response pairs to state--action--observation trajectories and task-closure evaluation.
    \item \textbf{A socio-technical roadmap for reliable autonomous AI systems:} We summarize challenges in long-horizon reliability, memory, safety, governance, skills, and system maintenance. We also discuss the broader implications of Digital Colleague systems for human–AI collaboration, including questions of ethics, skills, work pace, creativity, privacy, and asset boundaries.
\end{itemize}

Overall, this survey clarifies how LLMs are moving from conversational chatbots toward dependable digital colleagues. Importantly, the next generation of generative AI will be defined by self-evolving systems: integrated ecosystems in which models, workspaces, tools, skills, memories, evaluators, and governance mechanisms continuously convert operational experience into reusable skills, updated memories, stronger verification signals, safer policies, and more reliable work outcomes.

\section{Part I: The Evolution of LLM's Cognitive Core}
\headingnote{From ``Fast Response'' to ``Slow Thinking''}

\begin{figure}[t]
    \centering
    \includegraphics[width=\textwidth]{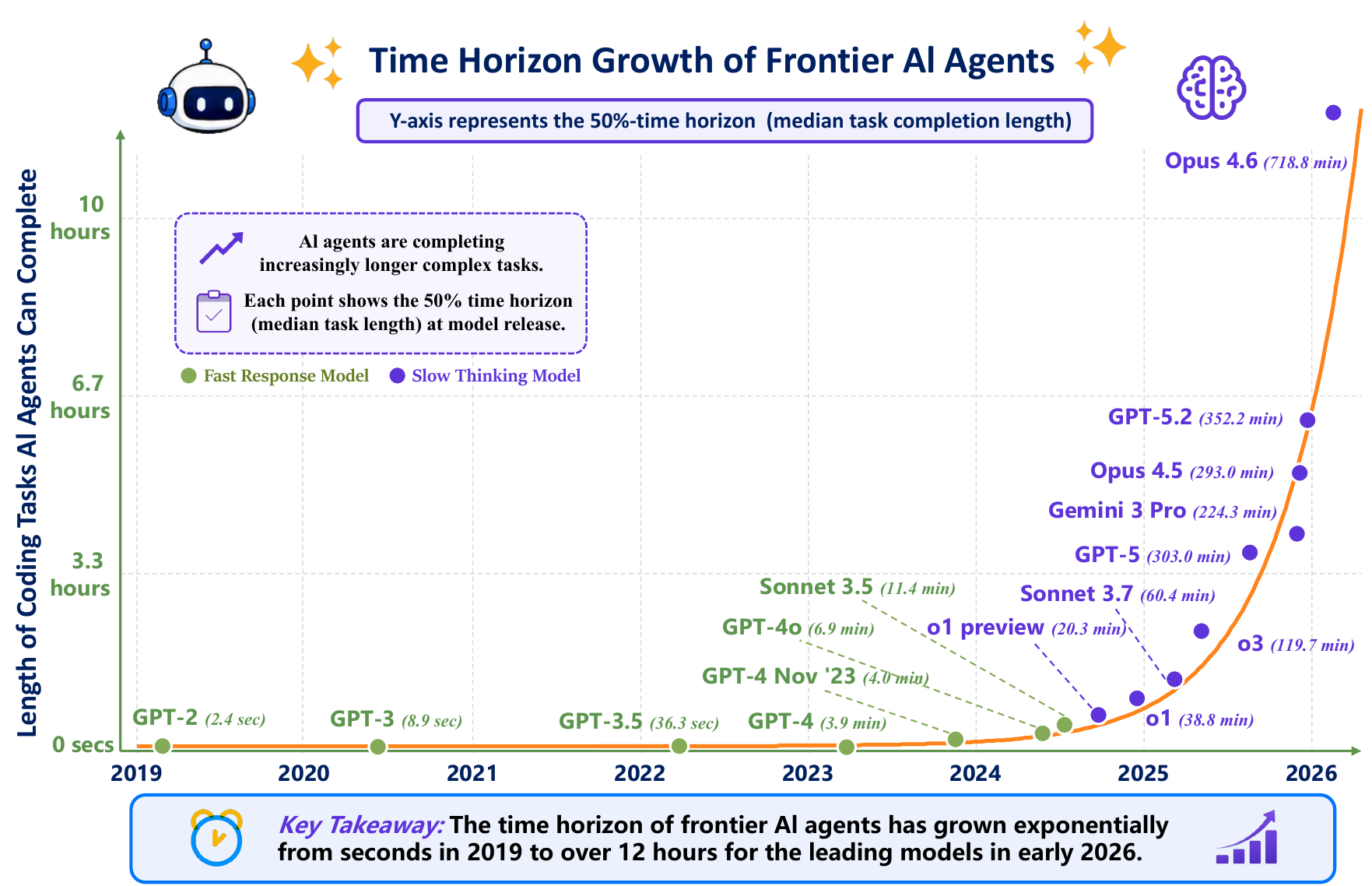}
    \caption{Time horizon growth of frontier AI agents. Each point reports the 50\%-time horizon, i.e., the median length of coding tasks that an agent can complete at release. The trend shows a transition from second-level fast-response models to slow-thinking models capable of sustaining increasingly long and complex tasks\protect\footnotemark.}
    \label{fig:horizon}
\end{figure}

This part examines the model-side cognitive core underlying next-generation generative AI systems.
As Figure~\ref{fig:horizon} illustrates this trajectory, we begin with the Chatbot era, where scaling, parametric knowledge compression, instruction alignment, and multimodal expansion turned LLMs into fluent fast-response interfaces that map prompts to plausible answers in one low-latency autoregressive pass.
We then enter the Thinking LLM era, where long Chain-of-Thought, inference-time scaling, and reinforcement learning push models toward deliberate System2 reasoning.
This progression matters because agentic systems require a reliable cognitive core: before acting in workspaces, a model must become a stronger generator, reasoner, and decision-maker\citep{lightman2024lets,madaan2023selfrefine,lu2023chameleon,qwen2026qwen3627bnonthinking,salem2026stateless,ding2024reasoning,li2025structured,ou2025survey,zhang2023multimodal,song2025progco}.

\footnotetext{Data source: \url{https://theaidigest.org/time-horizons}.}

\subsection{The Chatbot Era: Language Generation and Knowledge Compression}
\headingnote{Represented by ChatGPT}

\begin{figure}[t]
    \centering
    \includegraphics[width=\textwidth]{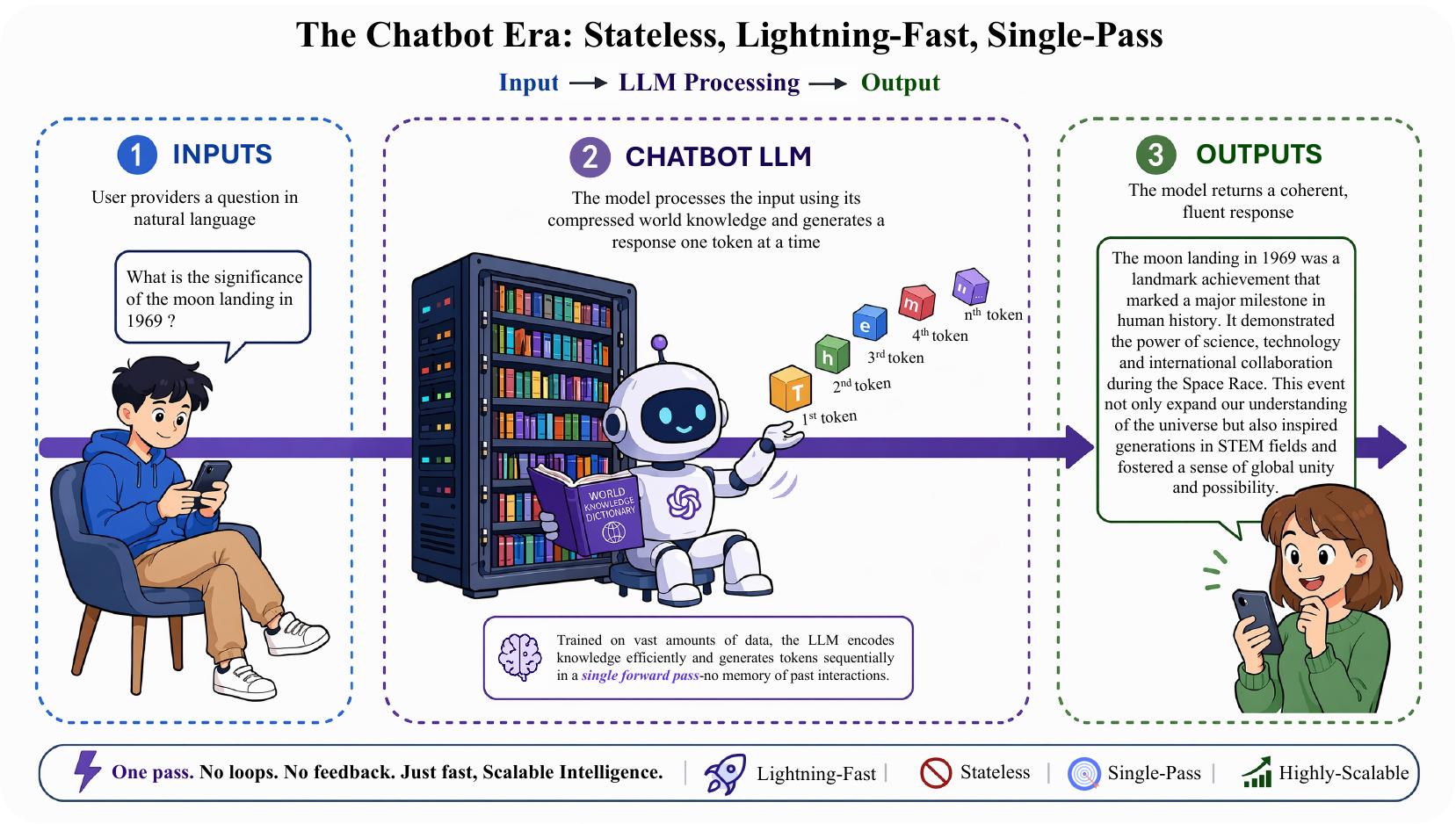}
    \caption{The Chatbot Era: a user inputs a natural-language question, the LLM performs fast, stateless, single-pass processing over compressed parametric knowledge, and immediately returns a fluent response. The figure highlights one-pass inference with no external loop, feedback-driven correction, or persistent memory.}
    \label{fig:chatbot}
\end{figure}

During this stage, as shown in Figure~\ref{fig:chatbot}, models primarily served as fast-response interfaces: they compressed linguistic and factual regularities into parameters, accepted natural-language prompts, and produced fluent answers through single-pass autoregressive generation \cite{kahneman2011thinking,zhao2023survey}. The Chatbot era was not defined by a single model release, but by the convergence of large-scale language generation, implicit knowledge compression, behavioral alignment, and later multimodal expansion. Together, these developments transformed LLMs from next-token predictors into fluent conversational systems, while simultaneously exposing the inherent ceiling of response-oriented generation when tasks require deliberate verification, search, and long-horizon reasoning~\citep{jiang2025safechain,he2025can,pang2026interactive,zhu2025chain,ranaldi2024aligning,qu2025recot,jiang2026reflexicoder,ding2026sherlock,costa2026enhancing,zhan2026can,zhang-etal-2024-autocap}.

\subsubsection{Scaling-Driven Language Generation and Parametric Knowledge Compression}

\textbf{\textit{Scaling-Driven Language Generation.}} The early development of statistical language models was constrained by the Markov assumption over local contexts. Although distributed representations and recurrent neural networks enhanced generalization, the true paradigm shift began with the Transformer architecture \cite{vaswani2017attention}, which broke the bottleneck of sequential computation and laid the foundation for large-scale parallel training. Building on this architecture, the field established the foundational ``pre-training and fine-tuning'' paradigm: GPT-1 \cite{radford2018improving} demonstrated generative pre-training followed by task-specific adaptation, while GPT-2 \cite{radford2019language} subsequently revealed that sufficiently large language models could perform zero-shot multitasking through natural-language prompts. In this sense, the first major contribution of the Chatbot era was to make language modeling a universal interface for task specification: instead of designing task-specific architectures, diverse problems could be reformulated as conditional text generation~\citep{tang2025unlocking,diao2024active,li2023effectiveness,zhang2022automatic,liu2026badthink,wang2023plan,li2025reflectevo,yan2024mirror,dhorreflexion,yang2024supercorrect,zhu2024closed,li2026cognitivemismatchmultimodallarge,xu-etal-2026-thinknote}.

The technical simplicity of Next-Token Prediction was central to this transformation. Autoregressive training converts heterogeneous web pages, books, code, dialogues, and documents into a single self-supervised objective: predicting the next token from prior context. Without manually labeled task data, this objective forces the model to absorb statistical regularities across syntax, semantics, factual associations, discourse, and basic procedures. Scaling corpus and model size therefore not only improves fluency, but also increases the density of patterns compressed into parameters. The Chatbot-era LLM is thus a high-capacity compression engine: it stores neither explicit symbolic rules nor database rows, but distributed approximations of linguistic, factual, and commonsense regularities reactivated through prompts~\citep{yao2023beyond,huang2024lateval, li2024llms,li2025rethinking, yu2024seqgpt, yu2025long,zhang2026think,jiang2025corvid,zhao2024enhancing,zhao2025evaluating,wang2026teaching,zhang2024small,pan2023automatically,ding2025multi}.

As summarized in Table \ref{tab:chatbot_llm}, this era witnessed the rapid expansion of non-reasoning LLMs, marked by parameter growth, architectural diversification, and closed-/open-source competition. Scaling Laws formed its empirical foundation: Kaplan et al. \cite{kaplan2020scaling} revealed power-law relationships between performance and scale; Gopher (280B) \cite{rae2021scaling} analyzed these effects across tasks; and Chinchilla (70B) \cite{hoffmann2022chinchilla} showed that compute-optimal training requires model size and tokens to scale together. This shift from parameter-centric to compute-optimal scaling showed that under-trained large models are inefficient, and that fast response quality depends on a balanced allocation of parameters, data, and compute~\citep{zheng2023ddcot,chen2025beyond,wen2025reinforcement,wang2026reinforcement,tang2026beyond,feng2026self,upadhyaya2024internalized,wang2024theoretical,liu2025monte,gao2024interpretable}.

OpenAI advanced this trajectory with GPT-3 (175B) \cite{brown2020language}, popularizing few-shot In-Context Learning, and later GPT-4 \cite{achiam2023gpt}, while Google introduced PaLM (540B) \cite{chowdhery2022palm}. In parallel, open-source models like OPT (175B) \cite{zhang2022opt}, BLOOM (176B) \cite{bigscience2022bloom}, and LLaMA \cite{touvron2023llama} democratized LLM research, proving smaller, extensively trained models could rival larger counterparts. This wave also changed empirical methodology: rather than treating LLMs as opaque, researchers could inspect training recipes, adaptation strategies, tokenizers, instruction data, and evaluation behavior, accelerating collective understanding of scaling-driven language generation~\citep{song2026maniplvm,liu2026trust,zhang2025surveyrl,liu2025enhancing,stojanovski2026reasoning}.

A key scale-enabled phenomenon is In-Context Learning. GPT-3 showed that a model can infer a task from a few demonstrations in the prompt without parameter updates \cite{brown2020language}. This blurs training and inference: weights provide a broad prior over tasks and knowledge, while the prompt temporarily selects and composes behavior. From the perspective of fast response, In-Context Learning acts as rapid associative retrieval from compressed experience. It requires no explicit search or long-horizon planning, but can quickly map prompt patterns to plausible continuations, making the model appear adaptive and instruction-aware even before explicit alignment training~\citep{han2026bridging,liu2026prorl,berti2025specializing,havrilla2024teaching,pan2026reward}.

As dense scaling faced bottlenecks in training cost and inference efficiency, Mixture of Experts (MoE) became a key route for sustaining scale. By activating only a few expert networks per token, MoE preserves massive capacity while reducing computation. Mixtral 8x7B \cite{jiang2024mixtral} pioneered this direction in the open-source community, while DeepSeek-V2 \cite{deepseek2024v2}, Grok-1, and DeepSeek-V3 \cite{deepseek2024v3} further pushed efficiency and matched GPT-4o \cite{hurst2024gpt} on multiple benchmarks. Even as the field shifted toward reasoning-centric paradigms, fast-response models such as Qwen3 \cite{qwen2025qwen3}, MiniMax-01 \cite{minimax2025minimax01scalingfoundationmodels}, GPT-4.5 \cite{openai2025gpt45}, and Grok 4 \cite{xai2025grok4} continued to push the efficiency frontier. This trajectory shows that Scaling Laws increasingly depend on sparse activation and efficient retrieval of compressed knowledge under low-latency constraints, not only dense parameter growth~\citep{srivastava2025technical,chen2025dast,wachi2026relative,li2026admtree,chen2026does,hong2025cooper,jiang2026verifiable}.

\textbf{\textit{Parametric Knowledge Compression.}} Unlike traditional AI systems relying on external knowledge graphs, the Chatbot era is defined by implicit knowledge compression. Through Next-Token Prediction over massive corpora, LLMs compress world knowledge, grammatical regularities, commonsense associations, and simple reasoning patterns into neural parameters. Petroni et al. \cite{petroni2019language} showed with LAMA probes that pre-trained models can function as implicit parametric knowledge bases, while later studies localized factual knowledge to Multilayer Perceptron layers \cite{meng2022locating,dai2022knowledge}. This parametric-memory view explains both the strength and weakness of Chatbot-era models. Knowledge in parameters is low-latency and flexibly recombinable, enabling open-ended answers without external databases; yet it is lossy, static after training, and difficult to audit or update precisely. Retrieval-augmented methods \cite{lewis2020retrieval} later compensated for these weaknesses, but the era's defining capability remained direct invocation of compressed knowledge from model weights~\citep{gunjal2025rubrics,zhang2025surveytts,chen2026fasttts,snell2025scaling,agarwal2025art}.

At scale, LLMs exhibited Emergent Abilities \cite{wei2022emergent}: without gradient updates, they could generalize to new tasks through a few prompt examples. Although Schaeffer et al. \cite{schaeffer2023emergent} argued that such emergence may partly reflect non-linear evaluation metrics, the practical consequence was undeniable: sufficiently scaled LLMs could act as general-purpose linguistic problem solvers across translation, summarization, question answering, coding, and simple reasoning. These abilities formed the substrate of ChatGPT-like systems. Before alignment made them helpful and conversational, scaling and parametric compression had endowed them with broad linguistic coverage, rapid associative retrieval, and flexible pattern completion. Thus, the Chatbot era represents the maturation of fast-response intelligence: fluent, knowledge-rich, broadly adaptive, and interaction-ready, yet driven by probabilistic generation rather than deliberate verification or search \citep{li2025system1system2,wang2026agenttts,ji2026survey,huang2025m1,yang2026towards,wangthink}.

\begin{table*}[htbp]
\centering
\scriptsize
\setlength{\tabcolsep}{3pt}
\renewcommand{\arraystretch}{1.1}
\caption{An overview of representative non-reasoning LLMs in the chatbot era.}
\label{tab:chatbot_llm}
\resizebox{\textwidth}{!}{

\begin{tabular}{@{} lllll | lllll @{}}
\toprule
\textbf{Model} & \textbf{Rel.} & \textbf{Para.} & \textbf{Type} & \textbf{Acc.} & \textbf{Model} & \textbf{Rel.} & \textbf{Para.} & \textbf{Type} & \textbf{Acc.} \\
\midrule

GPT-1~\cite{radford2018improving} & 2018-06 & 117M & Text & Open & Orca~\cite{mukherjee2023orca} & 2023-06 & 13B & Text & Closed \\
GPT-2~\cite{radford2019language,openai2019gpt2release} & 2019-02 & 1.5B & Text & Open & Llama 2~\cite{touvron2023llama2} & 2023-07 & 7--70B & Text & Open \\
PLATO~\cite{bao2020plato} & 2019-10 & 132M & Text & Open & InternLM~\cite{internlm2023} & 2023-07 & 7B/20B & Text & Open \\
T5~\cite{raffel2020exploring} & 2019-10 & 60M--11B & Text & Open & Claude 2~\cite{anthropic2023claude2} & 2023-07 & - & Text & Closed \\
DialoGPT~\cite{zhang2020dialogpt} & 2019-11 & 117M/345M/762M & Text & Open & WizardLM~\cite{xu2023wizardlm} & 2023-07 & 7B/13B/30B & Text & Open \\
Meena~\cite{adiwardana2020towards} & 2020-01 & 2.6B & Text & Closed & Qwen~\cite{bai2023qwen} & 2023-08 & 1.8--72B & Text & Open \\
BlenderBot~\cite{roller2021recipes} & 2020-04 & 90M/2.7B/9.4B & Text & Open & Qwen-VL~\cite{bai2023qwenvl} & 2023-08 & 9.6B & Multi & Open \\
GPT-3~\cite{brown2020language} & 2020-05 & 175B & Text & Closed & OpenFlamingo~\cite{awadalla2023openflamingo} & 2023-08 & 3--9B & Multi & Open \\
PLATO-2~\cite{bao2021plato} & 2020-06 & 93M/314M/1.6B & Text & Open & Code Llama-Instruct~\cite{roziere2023code} & 2023-08 & 7B/13B/34B & Code & Open \\
BlenderBot 2~\cite{parlai2021blenderbot2} & 2021-07 & 400M/2.7B & Text & Open & WizardMath~\cite{luo2023wizardmath} & 2023-08 & 7B/13B/70B & Text & Open \\
Jurassic-1~\cite{lieber2021jurassic} & 2021-08 & 178B & Text & Closed & WizardCoder~\cite{luo2024wizardcoder} & 2023-08 & 7B/13B/34B & Code & Open \\
Codex~\cite{chen2021evaluating} & 2021-08 & 12M-12B & Text & Closed & IDEFICS~\cite{laurenccon2023obelics} & 2023-08 & 9B/80B & Multi & Open \\
HyperCLOVA~\cite{kim2021hyperclova} & 2021-09 & 82B & Text & Closed & Phi-1.5~\cite{li2023phi15} & 2023-09 & 1.3B & Text & Open \\
PLATO-XL~\cite{bao2022plato} & 2021-09 & 11B & Text & Open & Baichuan 2~\cite{baichuan2023baichuan2} & 2023-09 & 7B/13B & Text & Open \\
Gopher~\cite{rae2021scaling} & 2021-12 & 280B & Text & Closed & GPT-4V~\cite{openai2023gpt4v} & 2023-09 & - & Multi & Closed \\
ERNIE 3.0 Titan~\cite{wang2021ernie} & 2021-12 & 260B & Text & Closed & Mistral 7B~\cite{jiang2023mistral7b} & 2023-09 & 7.3B & Text & Open \\
GLaM~\cite{du2022glam} & 2021-12 & 1.2T-A97B & Text & Closed & Mixtral~\cite{jiang2024mixtral} & 2023-09 & 7B & Text & Open \\
LaMDA~\cite{thoppilan2022lamda} & 2022-01 & 137B & Text & Closed & Kimi / Moonshot~\cite{moonshot2024kimi} & 2023-10 & - & Text & Closed \\
AlphaCode~\cite{li2022competition} & 2022-02 & 9B/41B & Text & Closed & ERNIE 4.0~\cite{baidu2023ernie4} & 2023-10 & - & Multi & Closed \\
InstructGPT~\cite{ouyang2022training} & 2022-03 & 1.3--175B & Text & Closed & Fuyu~\cite{adept2023fuyu} & 2023-10 & 8B & Multi & Open \\
Chinchilla~\cite{hoffmann2022chinchilla} & 2022-03 & 70B & Text & Closed & Zephyr-7B~\cite{tunstall2023zephyr} & 2023-10 & 7B & Text & Open \\
CodeGen~\cite{nijkamp2022conversational} & 2022-03 & 350M/2B/6B/16B & Text & Open & ChatGLM3-6B~\cite{thudm2023chatglm3-6b} & 2023-10 & 6B & Text & Open \\
PaLM~\cite{chowdhery2022palm} & 2022-04 & 540B & Text & Closed & Skywork-13B~\cite{wei2023skywork} & 2023-10 & 13B & Text & Open \\
Flamingo~\cite{alayrac2022flamingo} & 2022-04 & 80B & Multi & Closed & GPT-4 Turbo~\cite{openai2023devday,openai2024gpt4turbo} & 2023-11 & - & Text & Closed \\
OPT~\cite{zhang2022opt} & 2022-05 & 125M--175B & Text & Open & Grok-1~\cite{xai2024grok} & 2023-11 & 314B-A78.5B & Text & Open \\
GODEL~\cite{peng2022godel} & 2022-06 & 220M/770M & Text & Open & Yi~\cite{young2024yi} & 2023-11 & 6B/34B & Text & Open \\
BLOOM~\cite{bigscience2022bloom} & 2022-07 & 176B & Text & Open & CogVLM~\cite{wang2023cogvlm} & 2023-11 & 17B & Multi & Open \\
BlenderBot 3~\cite{shuster2022blenderbot} & 2022-08 & 3B/30B/175B & Text & Open & Claude 2.1~\cite{anthropic2023claude21} & 2023-11 & - & Text & Closed \\
PaLI / PaLI-X~\cite{chen2022pali,chen2023palix} & 2022-09 & 17B/55B & Multi & Closed & Inflection-2~\cite{inflection2023inflection2} & 2023-11 & - & Text & Closed \\
Sparrow~\cite{glaese2022improving} & 2022-09 & 70B & Text & Closed & DeepSeek Coder Instruct~\cite{deepseekai2023deepseekcoder} & 2023-11 & 1B--33B & Code & Open \\
CodeGeeX~\cite{zheng2023codegeex} & 2022-09 & 13B & Text & Open & OpenChat 3.5~\cite{wang2024openchat} & 2023-11 & 7B & Text & Open \\
GLM-130B~\cite{zeng2022glm130b} & 2022-10 & 130B & Text & Open & DeepSeek LLM~\cite{bi2024deepseek} & 2023-11 & 7B/67B & Text & Open \\
Galactica~\cite{taylor2022galactica} & 2022-11 & 120B & Text & Open & Orca 2~\cite{mitra2023orca} & 2023-11 & 7B/13B & Text & Open \\
BLIP-2~\cite{li2023blip} & 2023-01 & 4B-12B & Multi & Open & Mixtral 8x7B~\cite{jiang2024mixtral} & 2023-12 & 47B-A13B & Text & Open \\
Llama~\cite{touvron2023llama} & 2023-02 & 7--65B & Text & Open & Phi-2~\cite{microsoft2023phi2} & 2023-12 & 2.7B & Text & Open \\
Alpaca~\cite{alpaca2023stanford} & 2023-03 & 7B & Text & Open & Gemini 1.0~\cite{gemini2023} & 2023-12 & - & Multi & Closed \\
Claude 1~\cite{anthropic2023claude,bai2022constitutional} & 2023-03 & - & Text & Closed & InternVL 1.0~\cite{chen2023internvl} & 2023-12 & 6B+ & Multi & Open \\
PanGu-$\Sigma$~\cite{ren2023pangu} & 2023-03 & 1.085T & Text & Closed & SOLAR-10.7B~\cite{kim2024solar} & 2023-12 & 10.7B & Text & Open \\
BloombergGPT~\cite{wu2023bloomberggpt} & 2023-03 & 50B & Text & Closed & GLM-4~\cite{zhipu2024glm4} & 2024-01 & - & Text & Closed \\
ChatGLM-6B~\cite{thudm2023chatglm6b} & 2023-03 & $\sim$6.2B & Text & Open & GLM-4V~\cite{zhipu2024glm4} & 2024-01 & 9B & Multi & Closed \\
GPT-4~\cite{achiam2023gpt} & 2023-03 & - & Multi & Closed & LLaVA-NeXT~\cite{liu2024llavanext} & 2024-01 & 7--34B & Multi & Open \\
PaLM-E~\cite{driess2023palme} & 2023-03 & 562B & Multi & Closed & Stable LM 2~\cite{bellagente2024stable} & 2024-01 & 1.6B & Text & Open \\
Vicuna~\cite{vicuna2023lmsys} & 2023-03 & 7B/13B & Text & Open & Yi-VL~\cite{young2024yi} & 2024-01 & 6B/34B & Multi & Open \\
GPT-3.5 Turbo~\cite{openai2023gpt35turboapi} & 2023-03 & - & Text & Closed & Mistral Large~\cite{mistral2024large} & 2024-02 & - & Text & Closed \\
Pythia~\cite{biderman2023pythia} & 2023-04 & 70M--12B & Text & Open & Qwen 1.5~\cite{qwen2024qwen15moe} & 2024-02 & 0.5--72B & Text & Open \\
LLaVA~\cite{liu2023llava} & 2023-04 & 7B/13B & Multi & Open & Gemini 1.5~\cite{gemini2024gemini15} & 2024-02 & - & Multi & Closed \\
MiniGPT-4~\cite{zhu2023minigpt4} & 2023-04 & 7B/13B & Multi & Open & OLMo~\cite{groeneveld2024olmo} & 2024-02 & 1B/7B & Text & Open \\
Dolly 2.0~\cite{databricks2023dolly15k} & 2023-04 & 12B & Text & Open & StarCoder2~\cite{lozhkov2024starcoder} & 2024-02 & 3B/7B/15B & Code & Open \\
Stable LM~\cite{stabilityai2023stablelm} & 2023-04 & 3B/7B & Text & Open & Reka Flash~\cite{rekaai2024rekaflash} & 2024-02 & 21B & Multi & Closed \\
Falcon~\cite{almazrouei2023falcon,penedo2023refinedweb} & 2023-05 & 7--180B & Text & Open & Gemma~\cite{gemma2024} & 2024-03 & 2B/7B & Text & Open \\
MPT~\cite{mosaicml2023mpt7b,mosaicml2023mpt30b} & 2023-05 & 7B/30B & Text & Open & Qwen1.5-MoE~\cite{qwen2024qwen15moe} & 2024-03 & 14B-A2.7B & Text & Open \\
StarCoder~\cite{li2023starcoder} & 2023-05 & 15.5B & Text & Open & DBRX~\cite{databricks2024dbrx} & 2024-03 & 132B-A36B & Text & Open \\
RedPajama~\cite{together2023redpajama} & 2023-05 & 3B/7B & Text & Open & Jamba~\cite{lieber2024jamba} & 2024-03 & 52B-A12B & Text & Open \\
InstructBLIP~\cite{dai2023instructblip} & 2023-05 & - & Multi & Open & Claude 3~\cite{anthropic2024claude3} & 2024-03 & - & Multi & Closed \\
PaLM 2~\cite{anil2023palm} & 2023-05 & - & Text & Closed & Command R~\cite{cohere2024commandr} & 2024-03 & 35B & Text & Open \\
CodeT5+~\cite{wang2023codet5+} & 2023-05 & 220M-16B & Code & Open & Inflection-2.5~\cite{inflection2024inflection25} & 2024-03 & - & Text & Closed \\
Inflection-1~\cite{inflection2023inflection1} & 2023-06 & - & Text & Closed & DeepSeek-VL~\cite{deepseekai2024deepseekvl} & 2024-03 & 1.3B/7B & Multi & Open \\
Phi-1~\cite{gunasekar2023phi1} & 2023-06 & 1.3B & Text & Open & Grok-1.5~\cite{xai2024grok15} & 2024-03 & - & Text & Closed \\
Aquila~\cite{baai2023aquila} & 2023-06 & 7B/33B & Text & Open & MM1~\cite{mckinzie2024mm1} & 2024-03 & 3B/7B/30B & Multi & Closed \\
ChatGLM2-6B~\cite{thudm2023chatglm2-6b} & 2023-06 & 6B & Text & Open & Yi-9B~\cite{young2024yi} & 2024-03 & 9B & Text & Open \\
Baichuan-Chat~\cite{baichuaninc2023baichuan7b,baichuaninc2023baichuan13b} & 2023-06 & 7B/13B & Text & Open & Phi-3~\cite{abdin2024phi3} & 2024-04 & 3.8--14B & Text & Open \\
XGen-7B~\cite{nijkamp2023xgen} & 2023-06 & 7B & Text & Open & Mixtral 8x22B~\cite{mistral2024mixtral8x22b} & 2024-04 & 141B-A39B & Text & Open \\
\bottomrule
\end{tabular}
}

\end{table*}

\clearpage

\begin{table*}[!p]
\ContinuedFloat
\centering
\scriptsize
\setlength{\tabcolsep}{3pt}
\renewcommand{\arraystretch}{1.1}
\caption[]{An overview of representative non-reasoning LLMs in the chatbot era (continued).}
\resizebox{\textwidth}{!}{
\begin{tabular}{@{} lllll | lllll @{}}
\toprule
\textbf{Model} & \textbf{Rel.} & \textbf{Para.} & \textbf{Type} & \textbf{Acc.} & \textbf{Model} & \textbf{Rel.} & \textbf{Para.} & \textbf{Type} & \textbf{Acc.} \\
\midrule

Llama 3~\cite{meta2024llama3} & 2024-04 & 8B-A70B & Text & Open & Hunyuan-Large~\cite{sun2024hunyuan} & 2024-11 & 389B-A52B & Text & Open \\
Command R+~\cite{cohere2024commandrplus} & 2024-04 & 104B & Text & Open & OLMo 2~\cite{olmo20242} & 2024-11 & 7B/13B & Text & Open \\
InternVL 1.5~\cite{chen2024far} & 2024-04 & 26B & Multi & Open & Pixtral Large~\cite{mistral2024pixtrallarge} & 2024-11 & 124B & Multi & Mixed \\
Reka Core~\cite{rekaai2024rekacore} & 2024-04 & - & Multi & Closed & SmolVLM~\cite{marafioti2025smolvlm} & 2024-11 & 2B & Multi & Open \\
CodeQwen1.5~\cite{qwen2024codeqwen15} & 2024-04 & 7B & Code & Open & DeepSeek-V3~\cite{deepseek2024v3} & 2024-12 & 671B-A37B & Text & Open \\
IDEFICS2~\cite{laurenccon2024matters} & 2024-04 & 8B & Multi & Open & Llama 3.3~\cite{meta2024llama33} & 2024-12 & 70B & Text & Open \\
OpenELM~\cite{mehta2024openelm} & 2024-04 & 270M-3B & Text & Open & PaliGemma 2~\cite{google2024paligemma2} & 2024-12 & 3B/10B/28B & Multi & Open \\
Snowflake Arctic~\cite{snowflake2024snowflakearctic} & 2024-04 & 480B-A17B & Text & Open & DeepSeek-VL2~\cite{wu2024deepseek} & 2024-12 & 27B-A4.5B & Multi & Open \\
Doubao~\cite{bytedance2024doubao} & 2024-05 & - & Text & Closed & Falcon 3~\cite{Falcon3} & 2024-12 & 1B-10B & Text & Open \\
DeepSeek-V2~\cite{deepseek2024v2} & 2024-05 & 236B-A21B & Text & Open & Granite 3.1~\cite{ibm2024granite31} & 2024-12 & 1B-8B & Text & Open \\
GPT-4o~\cite{hurst2024gpt} & 2024-05 & - & Multi & Closed & InternVL2.5~\cite{chen2024expanding} & 2024-12 & 1B--78B & Multi & Open \\
CogVLM2~\cite{hong2024cogvlm2} & 2024-05 & 19B & Multi & Open & MiniMax-Text-01~\cite{minimax2025minimax01scalingfoundationmodels} & 2025-01 & 456B-A45.9B & Text & Open \\
MiniCPM-V~\cite{yao2024minicpmv} & 2024-05 & 2--8B & Multi & Open & MiniMax-VL-01~\cite{minimax2025minimax01scalingfoundationmodels} & 2025-01 & 456B-A45.9B & Multi & Open \\
Codestral~\cite{mistral2024codestral} & 2024-05 & 22B & Code & Open & Qwen2.5-Max~\cite{qwen2025qwen25max} & 2025-01 & - & Text & Closed \\
Falcon 2~\cite{malartic2024falcon2} & 2024-05 & 11B & Multi & Open & MiniCPM-o 2.6~\cite{openbmb2025minicpmo26} & 2025-01 & 8B & Multi & Open \\
PaliGemma~\cite{beyer2024paligemma} & 2024-05 & 3B & Multi & Open & Qwen2.5-VL~\cite{bai2025qwen2} & 2025-01 & 3B/7B/72B & Multi & Open \\
Aya 23~\cite{aryabumi2024aya} & 2024-05 & 8B/35B & Text & Open & Janus-Pro~\cite{chen2025janus} & 2025-01 & 1B/7B & Multi & Open \\
Granite Code~\cite{mishra2024granite} & 2024-05 & 3B-34B & Code & Open & Mistral Small 3~\cite{mistral2025mistralsmall3} & 2025-01 & 24B & Text & Open \\
Qwen 2~\cite{qwen2024qwen2} & 2024-06 & 0.5--72B & Text & Open & GPT-4.5~\cite{openai2025gpt45} & 2025-02 & - & Text & Closed \\
GLM-4-9B~\cite{glm2024glm4} & 2024-06 & 9B & Text & Open & Phi-4-mini~\cite{abouelenin2025phi} & 2025-02 & 4B & Text & Open \\
Claude 3.5 Sonnet~\cite{anthropic2024claude35} & 2024-06 & - & Multi & Closed & Phi-4-multimodal~\cite{abouelenin2025phi} & 2025-02 & 6B & Multi & Open \\
Cambrian-1~\cite{tong2024cambrian} & 2024-06 & 3--34B & Multi & Open & Command A~\cite{cohere2025commanda} & 2025-03 & 111B & Text & Closed \\
DeepSeek-Coder-V2~\cite{deepseek2024coderv2} & 2024-06 & 236B-A21B & Code & Open & Mistral Small 3.1~\cite{mistral2025small31} & 2025-03 & 24B & Multi & Open \\
Nemotron-4~\cite{nvidia2024nemotron} & 2024-06 & 340B & Text & Open & Aya Vision~\cite{dash2025aya} & 2025-03 & 8B/32B & Multi & Open \\
Gemma 2~\cite{google2024gemma2} & 2024-06 & 2B/9B/27B & Text & Open & Qwen2.5-VL-32B~\cite{qwen2025qwen25vl32b} & 2025-03 & 32B & Multi & Open \\
Skywork-MoE~\cite{wei2024skywork} & 2024-06 & 146B-A22B & Text & Open & OLMo 2 32B~\cite{allenai2025olmo232b} & 2025-03 & 32B & Text & Open \\
InternVL 2.0~\cite{internvl2024internvl2} & 2024-07 & 1--76B & Multi & Open & GPT-4.1~\cite{openai2025gpt41} & 2025-04 & - & Multi & Closed \\
Llama 3.1~\cite{dubey2024llama3} & 2024-07 & 8--405B & Text & Open & GPT-4.1 mini~\cite{openai2025gpt41} & 2025-04 & - & Multi & Closed \\
InternLM 2.5~\cite{internlm2024internlm25} & 2024-07 & 1.8--20B & Text & Open & GPT-4.1 nano~\cite{openai2025gpt41} & 2025-04 & - & Multi & Closed \\
GPT-4o mini~\cite{openai2024gpt4omini} & 2024-07 & - & Multi & Closed & Granite 3.3~\cite{ibm2025granite33} & 2025-04 & 2B/8B & Text & Open \\
Codestral Mamba~\cite{mistral2024codestralmamba} & 2024-07 & 7B & Code & Open & Kimi-VL-A3B-Instruct~\cite{kimiteam2025kimivl} & 2025-04 & 16B-A2.8B & Multi & Open \\
Mistral NeMo~\cite{mistral2024nemo} & 2024-07 & 12B & Text & Open & Amazon Nova Premier~\cite{Intelligence2025} & 2025-04 & - & Multi & Closed \\
SmolLM~\cite{allal2024smollm} & 2024-07 & 135M/360M/1.7B & Text & Open & Mistral Medium 3~\cite{mistral2025mistralmedium3} & 2025-05 & - & Multi & Closed \\
Mistral Large 2~\cite{mistral2024mistrallarge2} & 2024-07 & 123B & Text & Open & Devstral~\cite{mistral2025devstral} & 2025-05 & 24B & Code & Open \\
LLaVA-OV~\cite{llava2024onevision} & 2024-08 & 0.5--72B & Multi & Open & ERNIE-4.5-300B-A47B~\cite{baidu2025ernie45technicalreport} & 2025-06 & 300B-A47B & Multi & Open \\
Grok-2~\cite{xai2024grok2} & 2024-08 & - & Text & Closed & Qwen3-4B-Instruct~\cite{qwen2025qwen3} & 2025-07 & 4B & Text & Open \\
Grok-1.5V~\cite{xai2024grok15v} & 2024-08 & - & Multi & Closed & Kimi K2 Instruct~\cite{kimiteam2025kimik2} & 2025-07 & 1T-A32B & Text & Open \\
Phi-3.5-mini-instruct~\cite{abdin2024phi3} & 2024-08 & 3.8B & Text & Open & Qwen3-Coder~\cite{qwen2025qwen3coder} & 2025-07 & 480B-A35B & Text & Open \\
Phi-3.5-MoE-instruct~\cite{abdin2024phi3} & 2024-08 & 42B-A6.6B & Multi & Open & FastVLM~\cite{vasu2025fastvlm} & 2025-07 & 0.5B/1.5B/7B & Multi & Open \\
Jamba 1.5~\cite{team2024jamba} & 2024-08 & 398B-A94B & Text & Open & LFM2-VL~\cite{amini2025lfm2} & 2025-08 & 450M/1.6B/3B & Multi & Open \\
Qwen2-VL~\cite{wang2024qwen2vl} & 2024-09 & 2--72B & Multi & Open & LongCat-Flash-Chat~\cite{team2025longcat} & 2025-08 & 560B-A27B & Multi & Open \\
Llama 3.2 Text~\cite{meta2024llama32} & 2024-09 & 1B/3B & Text & Open & Qwen3-Next~\cite{qwen2025qwen3next80ba3binstruct} & 2025-09 & 81B-A3B & Text & Open \\
Llama 3.2 Vision~\cite{meta2024llama32} & 2024-09 & 11B/90B & Multi & Open & Qwen3-VL~\cite{bai2025qwen3} & 2025-09 & 235B-A22B & Multi & Open \\
Qwen2.5~\cite{qwen2024qwen25} & 2024-09 & 0.5--72B & Text & Open & Mistral Large 3~\cite{mistral2025mistrallarge3} & 2025-12 & 675B-A41B & Multi & Open \\
Pixtral~\cite{agrawal2024pixtral} & 2024-09 & 12B & Multi & Open & Ministral 3 Instruct~\cite{mistral2025mistrallarge3} & 2025-12 & 3B/8B/14B & Multi & Open \\
OLMoE~\cite{muennighoff2025olmoe} & 2024-09 & 7B-A1B & Text & Open & Devstral 2~\cite{mistral2025devstral2} & 2025-12 & 123B & Text & Open \\
Molmo~\cite{muennighoff2025olmoe} & 2024-09 & 7B/72B & Multi & Open & Devstral Small 2~\cite{mistral2025devstral2} & 2025-12 & 24B & Text & Open \\
Claude 3.5 Haiku~\cite{anthropic2024claude35haiku} & 2024-10 & - & Text & Closed & Youtu-LLM~\cite{lu2026youtullmunlockingnativeagentic} & 2026-01 & 1.96B & Text & Open \\
Aya Expanse~\cite{dang2024aya} & 2024-10 & 8B/32B & Text & Open & Youtu-VL~\cite{wei2026youtuvlunleashingvisualpotential} & 2026-01 & 4B & Multi & Open \\
Granite 3.0~\cite{granite2024granite} & 2024-10 & 1B-8B & Text & Open & Qwen3-Coder-Next~\cite{cao2026qwen3} & 2026-02 & 80B-A3B & Text & Open \\
Yi-Lightning~\cite{wake2024yi} & 2024-10 & - & Text & Closed & LongCat-Flash-Lite~\cite{liu2026scaling} & 2026-02 & 68.5B-A3B & Text & Open \\
Qwen2.5-Coder~\cite{qwen2024qwen25coder} & 2024-11 & 0.5--32B & Text & Open & Mistral Small 4-instruct~\cite{mistral2026mistralsmall4instruct} & 2026-03 & 119B-A6B & Multi & Open \\
Llama-3.1-Nemotron-70B~\cite{wang2024helpsteer2} & 2024-11 & 70B & Text & Open & LongCat-Next~\cite{team2026longcat} & 2026-03 & 68.5B-A3B & Multi & Open \\
\bottomrule
\end{tabular}
}

\end{table*}
\clearpage

\subsubsection{Alignment, Multimodal Expansion, and the Limits of Fast-Response Cognition}

\textbf{\textit{Behavioral Alignment.}} However, possessing vast parametric knowledge is insufficient for building a competent conversational system. The critical leap from GPT-3 to ChatGPT was behavioral alignment, which transformed a continuation model into an instruction-following assistant. Works such as FLAN \cite{wei2021finetuned} employed Supervised Fine-Tuning across diverse instruction sets, while OpenAI's InstructGPT \cite{ouyang2022training} established the Reinforcement Learning from Human Feedback (RLHF) pipeline to align outputs with human preferences. The finding that a 1.3B InstructGPT model outperformed the 175B GPT-3 in human evaluations highlighted a crucial insight: alignment and capability are orthogonal dimensions. Alignment improves instruction following and interaction quality, whereas larger-scale pretraining remains central to general knowledge and broad task competence~\citep{kim2026cost,wu2024inference,xu2025let,lin2025plan,pan2025survey,jin2025energy}.

Different institutions further refined conversational alignment along different dimensions. Google's LaMDA (137B) \cite{thoppilan2022lamda} emphasized dialog safety, factuality, and quality. Anthropic's Claude adopted the Helpful, Honest, and Harmless (HHH) framework \cite{bai2022training} and Constitutional AI \cite{bai2022constitutional}, enabling models to critique and revise their own outputs according to constitutional principles while reducing reliance on direct human annotation. Later, Direct Preference Optimization \cite{rafailov2023direct} simplified preference alignment by unifying reward modeling and policy optimization. These collective efforts allowed LLMs to move beyond raw text continuation and interact with users through human-like fluency, helpfulness, and emotional awareness~\citep{wang2025think,khalifa2026reasoning,lin2025plan,li2026test,liu2025efficient,qin-etal-2023-cross,du2024llms}.

\textbf{\textit{Multimodal Expansion.}} Beyond behavioral alignment, the late Chatbot era also expanded the sensory boundary of LLMs through multimodal fusion, gradually transitioning them from purely language intelligence toward a more comprehensive perceptual intelligence \cite{kuang2025natural,li2024towards,kuang2025express,liu2026tangrampuzzle,wan2025wanopenadvancedlargescale,dong2024modality}. Early visual-language models such as LLaVA \cite{liu2023llava} relied on stitched architectures that connected external visual encoders to LLMs through projection layers, but such designs suffered from cross-modal alignment bottlenecks. By 2024, the field shifted toward native end-to-end multimodality. Models such as GPT-4o \cite{hurst2024gpt} and Google's Gemini series \cite{gemini2023} processed text, image, and audio modalities within increasingly unified neural network architectures, while also extending context windows to support long-video and long-context understanding. Open-source models such as InternVL 1.5 \cite{chen2024far} and Qwen2-VL \cite{wang2024qwen2vl} rapidly narrowed the gap with closed-source frontiers. In parallel, domain-specific continual pre-training produced strong coding specialists, including DeepSeek-Coder-V2 \cite{deepseek2024coderv2} and Qwen2.5-Coder \cite{qwen2024qwen25coder}, extending fast-response generation into formal language domains to handle complex software engineering tasks~\citep{liu2025safe,luo2024improve,wang2026survey,zhang2025lessons,luo2026unlocking,zeng2025janusvln,gemmateam2025gemma3technicalreport}.

\textbf{\textit{Limits of Fast-Response Cognition.}} Despite the unprecedented success of conversational and multimodal models, their autoregressive generation paradigm dictates inherent limitations as fast-response systems. These shortcomings are especially visible in deterministic domains requiring rigorous logic. First, standard LLMs tend to primarily rely on surface-level pattern matching rather than strict multi-step deduction. In formal mathematical proofs and long-horizon code generation, local errors in intermediate steps can cascade into global failures \cite{chen2021evaluating,hendrycks2021math}. Because mathematical reasoning requires precise intermediate states and combinatorial planning, the inability of standard LLMs to dynamically allocate additional computation for deeper exploration causes accuracy to drop sharply as problem complexity increases~\citep{zeng2026priordrive,zheng2025survey,ying2024internlm,zhang2025process,setlur2025rewarding,yang2025beyond,dong2024cost}.

Second, because knowledge is implicitly compressed without persistent external grounding, LLMs are prone to hallucinations, often generating fluent but fabricated statements with high confidence \cite{ji2023hallucination,zhang2025siren}. Evaluations further suggest that scaling model size alone may even exacerbate deceptive fluency \cite{lin2022truthfulqa}. Most critically, the autoregressive paradigm lacks intrinsic verification and self-correction mechanisms. As pointed out by Yao et al. \cite{yao2023tree} and LeCun \cite{lecun2022path}, greedy left-to-right probabilistic decoding does not naturally support lookahead, backtracking, or global search---the cognitive operations required for hard mathematics, coding, and planning. Empirical studies \cite{cobbe2021gsm8k,huang2024large} further show that, without environmental feedback or independent verifiers, LLMs cannot reliably achieve genuine self-correction using only their probability distributions~\citep{pronesti2026beyond,xiong2025self,zhu2025retrieval,sun2025efficient,ospanov2025hermes}.

This fast-response mechanism, therefore, constitutes an insurmountable performance ceiling for tasks that require deliberate reasoning. Prompt engineering techniques such as Chain-of-Thought \cite{wei2022chain} and Self-Consistency \cite{wang2022self} attempted to trade longer token generation for deeper reasoning, but they did not fundamentally alter the training paradigm of unidirectional, response-oriented output. For LLMs to master complex mathematical derivations and algorithmic programming, their cognitive core must transition from immediate answer generation to deliberate, trial-and-error decision-making. This need to inject search, verification, and test-time computation into model reasoning catalyzed the emergence of the Thinking LLM era, driven by Reinforcement Learning and inference-time scaling~\citep{yu2025gcot,besta2025demystifying,chowdhury2025zero,joshi2025review,li2025automatic,wu2025hunyuanvideo15technicalreport}.

\begin{AIbox}{Key Difference: Fast Response vs. Deliberate Reasoning}
    \begin{itemize}[left=2pt,topsep=1pt,itemsep=2pt, parsep=1pt]
        \item Chatbot-era LLMs turned language generation into a universal interface by compressing massive linguistic, factual, and commonsense patterns into parameters, enabling fluent, low-latency responses and rapid in-context adaptation.

        \item Their core limitation is that fast autoregressive response lacks intrinsic verification, lookahead, and backtracking, making complex mathematics, coding, and long-horizon planning fragile without additional reasoning mechanisms.
    \end{itemize}
\end{AIbox}

\subsection{The Thinking LLM Era: Reasoning and Reinforcement-Learning-Driven Cognition}
\headingnote{Represented by OpenAI o1 and DeepSeek R1}

\begin{figure}[t]
    \centering
    \includegraphics[width=\textwidth]{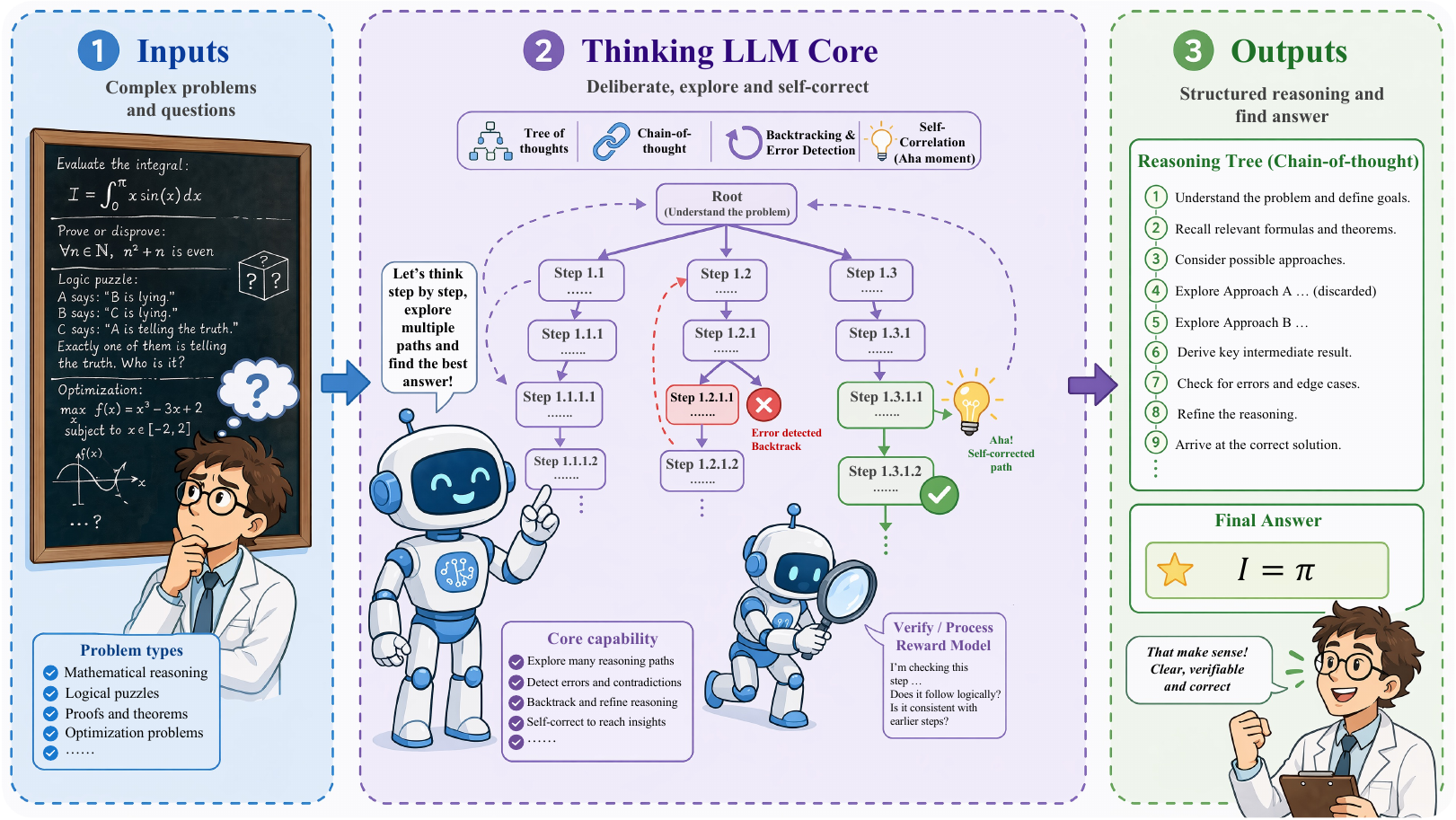}
    \caption{The Thinking LLM Era: the model allocates additional inference-time computation, generates long reasoning traces, explores alternatives, verifies intermediate steps, and then returns a more deliberate answer. The figure contrasts slow, reflective System-2-style reasoning with the chatbot's fast single-pass response.}
    \label{fig:thinkingllm}
\end{figure}

As shown in Figure~\ref{fig:thinkingllm}, the Thinking LLM era begins when large language models are no longer treated only as fast generators of plausible text, but as systems that can allocate computation to deliberate reasoning before answering~\citep{fricke2026framework,tan2024thought,praas2023self,yao2023tree,vacareanu2024general}.
Instead of relying solely on compressed parametric knowledge and shallow pattern completion, reasoning-oriented models generate extended intermediate traces, explore alternatives, verify partial results, and learn from outcome signals~\citep{chiu2024r,liao2026research,hu2024unveiling,chen2026cap,zhang2024chain}.
This era therefore upgrades the LLM from a conversational interface into a more reliable cognitive engine for mathematical reasoning, coding, planning, and later agentic decision making.
We summarize this transition through two intertwined developments: long Chain-of-Thought with inference-time scaling, and reinforcement-learning-driven reasoning that internalizes search, reflection, and self-correction~\citep{peng2025revisiting,chen2024not,li2025think,sui2025stop,wangefficient,zhang-etal-2024-wrong,11209984,liu2026letsthinkimagesefficiently}.

\subsubsection{Long Chain-of-Thought and Inference-Time Scaling}

The Chatbot era demonstrated that LLMs, through Next-Token Prediction on massive corpora, could compress world knowledge into neural parameters and serve as effective human-machine interfaces \cite{zhao2023survey}. Yet this capability is fundamentally limited by the fast-response nature of autoregressive generation: it is fluent, associative, and pattern-driven, but does not by itself support deliberate verification or search. The model is not genuinely thinking but performing probabilistic prediction, leading to hallucination, fragile reasoning, and systematic failure on tasks requiring deep multi-step inference \cite{huang2023hallucination,mirzadeh2024gsmsymbolic}. The arrival of OpenAI o1 \cite{openai2024o1} and DeepSeek R1 \cite{deepseek2025r1} provided a decisive shift. Before emitting a final response, these Reasoning LLMs (RLLMs) generate an extended internal deliberation, potentially thousands of tokens long, in which they decompose problems, explore alternative strategies, and detect and correct errors. This constitutes System 2 ``slow thinking'' \cite{stanovich2000individual}, deliberate, effortful, and self-monitored, marking a qualitative leap beyond the immediate response generation of the Chatbot era~\citep{yue2025don,hankal2025adaptive,liu2025think,an2026don,wang2026thoughts}.

The technical core of this transition is Long Chain-of-Thought (Long CoT) \cite{chen2025towards}, a qualitatively different mode of reasoning from the traditional Short CoT of the Chatbot era. Short CoT, exemplified by "Let's think step by step" \cite{kojima2022large} and few-shot demonstrations \cite{wei2022chain}, introduced intermediate reasoning steps but remained fundamentally shallow: linear, single-path traces with limited logical depth, no branching, and no self-revision. Empirical analysis confirmed that its benefits are largely confined to mathematical and symbolic tasks \citep{sprague2025cot,li2025one, li2025refine, han2025your,hu2026smartthinker,sun2025stop,shen2026alleviating,wan2026mitigating,vitcot}.

Long CoT transcends these limitations along three dimensions: deep reasoning, where chains sustain coherent derivations across tens or hundreds of steps, provably expanding Transformer expressiveness beyond fixed-depth computation \cite{feng2023towards}; extensive exploration, where models branch into alternative approaches within a single generation, internalizing the multi-path search previously requiring external structures like Tree-of-Thought \cite{yao2023tree} and Graph-of-Thought \cite{besta2024graph}; and feasible reflection, where models revisit earlier steps to detect and correct errors, a capability that Self-Refine \cite{madaan2023selfrefine} achieved only through iterative multi-call pipelines. These three capabilities were developed independently during the Chatbot era through external frameworks, but reinforcement learning in the o1/R1 generation unified them inside a single model, enabling spontaneous interleaving of derivation, branching, and self-correction within one output sequence \cite{deepseek2025r1,yeo2025demystifying}. This internalization is the essential breakthrough of the Thinking LLM paradigm~\citep{zheng2025fast,ge2025innate,guo2025deepseek,cai2025unilaw,zhang2025100}.

The Thinking LLM paradigm has given rise to several distinctive phenomena. The first is \textit{inference-time scaling}: rather than improving performance solely through larger models and more training data, Thinking LLMs invest additional computation at inference time \cite{snell2024scaling}, either sequentially by generating longer reasoning chains or in parallel by sampling multiple paths and selecting the best. Strikingly, with sufficient inference-time compute a 1B model can surpass a 405B model on mathematical benchmarks \cite{liu2025can}, shifting the paradigm from "scaling up models" to "scaling up reasoning." The second phenomenon concerns \textit{reasoning boundaries} and \textit{overthinking}: longer chains do not monotonically improve performance, as each model has a capability upper bound beyond which errors accumulate and accuracy degrades \cite{chen2024unlocking,chen2025donot}, motivating efficiency techniques such as Long-to-Short distillation. The third is the \textit{Aha moment} reported during DeepSeek-R1's pure RL training, where the model spontaneously produced self-reflective utterances within its reasoning traces \cite{deepseek2025r1}. This finding remains contested, as subsequent analysis suggested such behavior may reflect pretraining biases amplified by GRPO's length optimization rather than genuine emergent reflection \cite{liu2025understanding}. These phenomena collectively define the empirical landscape of Thinking LLMs and motivate the technical routes examined below~\citep{parmar2025challenges,dang2025reinforcement,liu2025logical,chen2026learning,so2025large}.

\subsubsection{Reinforcement-Learning-Driven Reasoning and Unified Cognitive Systems}

The technical evolution of Thinking LLMs can be traced along two intertwined threads: how reasoning capabilities are elicited, and how models are trained to acquire them.

\textbf{Elicitation of Reasoning: From External Scaffolding to Internal Autonomy.} The earliest attempts to elicit reasoning relied on prompt engineering. Few-shot CoT \cite{wei2022chain} and zero-shot CoT \cite{kojima2022large} demonstrated that inserting intermediate reasoning steps could improve performance, but the resulting traces remained shallow and single-path. Subsequent work introduced external search structures: Tree-of-Thought \cite{yao2023tree} enabled multi-candidate generation with evaluation and backtracking, Graph-of-Thought \cite{besta2024graph} generalized reasoning to arbitrary graph structures, and Self-Consistency \cite{wang2022self} improved robustness by sampling multiple reasoning paths and selecting the most frequent answer through majority voting. While effective, all of these methods depended on external orchestration rather than learned reasoning behavior~\citep{liu2025fin,hayder2025highlighting,sun2025reinforcement,dong2025enhancing,ren2025deepseek,lin2023solving}.

The decisive shift came with RL-driven internalization in OpenAI o1 \cite{openai2024o1} and DeepSeek R1 \cite{deepseek2025r1}, where models trained with reinforcement learning spontaneously generated deep, branching, self-correcting reasoning within a single output, eliminating the need for any external scaffold. Most recently, reasoning has evolved from a standalone capability into an adjustable mode within unified systems. Early theoretical groundwork by Dualformer \cite{su2024dualformer} demonstrated that a single Transformer could be trained to flexibly switch between fast and slow reasoning modes. This idea was realized at scale when Qwen3 \cite{qwen2025qwen3} introduced seamless thinking/non-thinking mode switching, allowing dynamic allocation of reasoning effort based on task complexity. This trajectory, from prompt-elicited to externally structured to RL-internalized to hybrid-mode reasoning, represents a progressive movement toward fully autonomous deliberation~\citep{neha2025survey,du2025challenge,hou2025thinkprune,xu2026rlkd,yan2026distribution}.

\textbf{Training Paradigm: From Imitation to Self-Evolution.} Early training for reasoning relied on supervised fine-tuning (SFT) over human-annotated or model-generated CoT data \cite{cobbe2021gsm8k}, which taught models to imitate reasoning formats but could not push them beyond the patterns present in the training set. The distillation of Long CoT from stronger models offered a more efficient path: DeepSeek itself released the R1-Distill series by distilling R1's reasoning traces into smaller models \cite{deepseek2025r1}, while LIMO \cite{ye2025limo} demonstrated that merely 817 curated examples could elicit strong mathematical reasoning, and s1 \cite{muennighoff2025s1} achieved comparable results with 1,000 samples. These findings suggest that reasoning capabilities are already latent in pretrained models and require activation rather than creation from scratch, a hypothesis with significant, profound, and practical implications for the efficiency of reasoning training~\citep{dai2024beyond,luo2025deconstructing,gao2025towards,feng2025efficient,zhang2026reinforcement}.

The true leap came with RL-based self-learning, especially through reinforcement learning with verifiable rewards (RLVR), where rule-based answer matching and format constraints provide scalable outcome-level supervision for mathematics, coding, and other self-verifiable domains. Algorithmically, this line evolved from PPO~\cite{schulman2017proximal}, which relies on a clipped surrogate objective and, in RLHF-style implementations, typically requires separate policy, reference, reward, and value models, to GRPO ~\cite{shao2024deepseekmath}, which removes the value model and estimates advantages by comparing multiple responses sampled for the same prompt. This critic-free design substantially reduced the memory burden of PPO and became the foundation of R1-style reasoning RL~\citep{wang2025efficient,yeo2025demystifying,li2023mixed,cetin2026reinforcement,shridhar2023distilling}.

After GRPO, policy optimization for long-CoT reasoning developed rapidly along an outcome-reward trajectory. DAPO~\cite{yu2026dapo} turned the reproduction of R1-like RL into a practical large-scale recipe by introducing decoupled clipping, dynamic sampling, token-level policy-gradient loss, and overlong reward shaping. Shortly afterward, Dr. GRPO ~\cite{liu2025understanding} revisited GRPO from an optimization-bias perspective, showing that response-length normalization and group-level standard-deviation normalization can introduce length and difficulty biases, and proposing a simplified objective that improves token efficiency while maintaining reasoning performance. CISPO~\cite{chen2025minimax} further modified the clipping mechanism by clipping importance-sampling weights rather than token-level policy updates, thereby preserving gradient contributions from rare but reasoning-critical tokens. GSPO~\cite{zheng2025group} moved the trust-region mechanism from token level to sequence level, defining importance ratios by sequence likelihood and aligning clipping, reward, and optimization with the sequence-level nature of outcome rewards. SAPO ~\cite{gao2025soft} continued this trend by replacing hard clipping with a smooth, temperature-controlled adaptive gate, preserving sequence coherence while selectively down-weighting highly off-policy tokens~\citep{wan2025qwenlong,kumar2025llm,kim2026correct,lu2024autopsv,chen2024alphamath}.
 A key empirical finding is that the simplest reward designs, rule-based answer matching plus format checking, proved most effective, as demonstrated by DeepSeek-R1's success. While process reward models~\cite{lightman2024lets} theoretically offer finer-grained supervision than outcome-based rewards, they face challenges of annotation cost and reward hacking, and outcome supervision has proven sufficient in practice~\citep{guo2025right,weng2023large,ahn2024large,amjad2026mathematical,poola2023tuning}.

\begin{table}[htbp]
\centering
\scriptsize
\setlength{\tabcolsep}{3pt}
\renewcommand{\arraystretch}{1.1}
\caption{An overview of representative reasoning LLMs in the Thinking LLM era.}
\label{tab:thinking_llm}
\resizebox{\textwidth}{!}{
\begin{tabular}{@{} lllll | lllll @{}}
\toprule
\textbf{Model} & \textbf{Rel.} & \textbf{Para.} & \textbf{Type} & \textbf{Acc.} & \textbf{Model} & \textbf{Rel.} & \textbf{Para.} & \textbf{Type} & \textbf{Acc.} \\
\midrule
o1-preview~\cite{openai2024o1systemcard} & 2024-09 & -- & Text & Closed & Claude 4 Opus~\cite{anthropic2025claude4systemcard} & 2025-05 & -- & Multi & Closed \\
o1-mini~\cite{openai2024o1systemcard} & 2024-09 & -- & Text & Closed & MiniMax-M1~\cite{chen2025minimax} & 2025-06 & 456B/46B & Text & Open \\
Marco-o1~\cite{zhao2024marcoo1} & 2024-11 & 7B & Text & Open & Kimi-Dev-72B~\cite{moonshotai2025kimidev} & 2025-06 & 72B & Code & Open \\
QwQ-32B-Preview~\cite{qwen2024qwqpreview} & 2024-11 & 32B & Text & Open & MiMo-VL-7B~\cite{xiaomimimo2025mimovl} & 2025-06 & 7B & Multi & Open \\
Skywork-o1 Open~\cite{skywork2024skyworko1open} & 2024-11 & 8B & Text & Open & Hunyuan-A13B-Instruct~\cite{tencent2025hunyuana13binstruct} & 2025-06 & 80B-A13B & Text & Open \\
o1~\cite{openai2024o1systemcard} & 2024-12 & -- & Text & Closed & Kimi K2~\cite{kimiteam2025kimik2} & 2025-07 & 1T/32B & Multi & Open \\
o1-pro~\cite{openai2024chatgptpro} & 2024-12 & -- & Text & Closed & Qwen3-Coder~\cite{qwen2025qwen3coder} & 2025-07 & 480B/35B & Code & Open \\
Gemini 2.0 Flash Thinking~\cite{google2025gemini20flashthinking} & 2024-12 & -- & Multi & Closed & Qwen3-235B-Thinking-2507~\cite{qwen2025qwen3235bthinking2507} & 2025-07 & 235B/22B & Text & Open \\
QVQ-72B-Preview~\cite{qwen2024qvqpreview} & 2024-12 & 72B & Multi & Open & Grok 4~\cite{xai2025grok4modelcard} & 2025-07 & -- & Multi & Closed \\
DeepSeek-R1-Zero~\cite{deepseek2025r1} & 2025-01 & 671B/37B & Text & Open & SmolLM3~\cite{bakouch2025smollm3} & 2025-07 & 3B & Text & Open \\
DeepSeek-R1~\cite{deepseek2025r1} & 2025-01 & 671B/37B & Text & Open & GPT-5~\cite{openai2025gpt5systemcard} & 2025-08 & ${\sim}$300B & Multi & Closed \\
R1-Distill-Qwen~\cite{deepseek2025r1} & 2025-01 & 1.5B--32B & Text & Open & DeepSeek-V3.1~\cite{deepseekai2025deepseekv31} & 2025-08 & 685B/37B & Text & Open \\
R1-Distill-Llama~\cite{deepseek2025r1} & 2025-01 & 8B/70B & Text & Open & GPT-oss-120B~\cite{agarwal2025gptoss} & 2025-08 & 117B/5.1B & Text & Open \\
Kimi k1.5~\cite{kimiteam2025kimiK15} & 2025-01 & -- & Multi & Closed & GPT-oss-20B~\cite{agarwal2025gptoss} & 2025-08 & 20B & Text & Open \\
Sky-T1-32B~\cite{novasky2025skyt1} & 2025-01 & 32B & Text & Open & Claude Opus 4.1~\cite{anthropic2025claudeopus41} & 2025-08 & -- & Multi & Closed \\
o3-mini~\cite{openai2025o3minisystemcard} & 2025-01 & -- & Text & Closed & ERNIE 4.5-Thinking~\cite{baidu2025ernie45thinking} & 2025-09 & 21B/3B & Text & Open \\
s1~\cite{muennighoff2025s1} & 2025-02 & 32B & Text & Open & Claude Sonnet 4.5~\cite{anthropic2025claudesonnet45} & 2025-09 & -- & Multi & Closed \\
LIMO~\cite{ye2025limo} & 2025-02 & 32B & Text & Open & Grok 4 Fast~\cite{xai2025grok4} & 2025-09 & -- & Multi & Closed \\
Grok 3~\cite{xai2025grok3} & 2025-02 & -- & Multi & Closed & MiniMax-M2~\cite{minimax2025minimaxm2} & 2025-10 & 230B/10B & Multi & Open \\
Grok 3 mini~\cite{xai2025grok3} & 2025-02 & -- & Text & Closed & Claude Haiku 4.5~\cite{anthropic2025claudehaiku45} & 2025-10 & -- & Multi & Closed \\
Claude 3.7 Sonnet~\cite{anthropic2025claude37sonnet} & 2025-02 & -- & Multi & Closed & Grok 4.1 Fast~\cite{xai2026grok41} & 2025-10 & -- & Multi & Closed \\
Hunyuan-T1-Preview~\cite{tencent2025hunyuanT1} & 2025-02 & -- & Text & Closed & Ring-1T~\cite{team2025every} & 2025-10 & 1T-A50B & Text & Open \\
Open-Reasoner-Zero~\cite{hu2025openreasonerzero} & 2025-02 & 7B/32B & Text & Open & GPT-5.1~\cite{openai2025gpt51} & 2025-11 & -- & Multi & Closed \\
TinyZero~\cite{pan2025tinyzero} & 2025-02 & 3B & Text & Open & Gemini 3 Pro~\cite{google2025gemini3} & 2025-11 & -- & Multi & Closed \\
Eurus-2-PRIME~\cite{cui2025prime} & 2025-02 & 7B & Text & Open & Grok 4.1~\cite{xai2025grok41} & 2025-11 & -- & Multi & Closed \\
Bespoke-Stratos~\cite{bespokelabs2025stratos7b} & 2025-02 & 7B & Text & Open & Claude Opus 4.5~\cite{anthropic2025claudeopus45} & 2025-11 & -- & Multi & Closed \\
Light-R1~\cite{wen2025lightr1} & 2025-02 & 7B/14B & Text & Open & DeepSeek-V3.2~\cite{deepseekai2025deepseekv32} & 2025-12 & 671B/37B & Text & Open \\
Hunyuan-TurboS~\cite{team2025hunyuan} & 2025-02 & 560B-A56B & Text & Closed & DeepSeek-V3.2-Speciale~\cite{deepseekai2025deepseekv32} & 2025-12 & 671B/37B & Text & Open \\
Gemma 3~\cite{gemmateam2025gemma3technicalreport} & 2025-03 & 4B/12B/27B & Multi & Open & Gemini 3 Flash~\cite{google2025gemini3flash} & 2025-12 & -- & Multi & Closed \\
QwQ-32B~\cite{qwen2025qwq32b} & 2025-03 & 32B & Text & Open & MiMo-V2-Flash~\cite{xiaomi2026mimov2flash} & 2025-12 & 309B/15B & Text & Open \\
Hunyuan-T1~\cite{tencent2025hunyuanT1} & 2025-03 & -- & Text & Closed & GLM-4.7~\cite{zai2025glm47} & 2025-12 & 358B & Text & Open \\
Gemini 2.5 Pro~\cite{google2025gemini25pro} & 2025-03 & -- & Multi & Closed & Devstral 2~\cite{mistral2025devstral2} & 2025-12 & 123B & Code & Open \\
DeepSeek-V3-0324~\cite{deepseek2025deepseekv30324} & 2025-03 & 671B/37B & Text & Open & GPT-5.2~\cite{openai2025gpt52} & 2025-12 & -- & Multi & Closed \\
Phi-4-reasoning~\cite{abdin2025phi4reasoning} & 2025-04 & 14B & Text & Open & LongCat-Flash-Thinking-2601~\cite{team2026longcatflashthinking2601} & 2026-01 & 560B-A27B & Text & Open \\
Phi-4-reasoning-plus~\cite{abdin2025phi4reasoning} & 2025-04 & 14B & Text & Open & Step 3.5 Flash~\cite{stepfun2026step35flash} & 2026-02 & -- & Text & Open \\
Qwen3~\cite{qwen2025qwen3} & 2025-04 & 0.6B--235B & Text & Open & Kimi K2.5~\cite{kimiteam2026kimik25} & 2026-02 & 1T/32B & Multi & Open \\
o3~\cite{openai2025o3o4mini} & 2025-04 & -- & Multi & Closed & Qwen3.5~\cite{qwen2026qwen35} & 2026-02 & 397B/17B & Multi & Open \\
o4-mini~\cite{openai2025o3o4mini} & 2025-04 & -- & Multi & Closed & Gemini 3.1 Pro~\cite{google2026gemini31pro} & 2026-02 & -- & Multi & Closed \\
Kimi-VL-A3B-Thinking~\cite{kimiteam2025kimivl} & 2025-04 & 2.8B act. & Multi & Open & GPT-5.3-Codex~\cite{openai2026gpt53codex} & 2026-02 & -- & Code & Closed \\
GLM-Z1-32B~\cite{zai2025glm40414} & 2025-04 & 32B & Text & Open & Claude Opus 4.6~\cite{anthropic2026claudeopus46} & 2026-02 & -- & Multi & Closed \\
Z1-Rumination-32B~\cite{zai2025glm40414} & 2025-04 & 32B & Text & Open & MiniMax-M2.5~\cite{minimax2026minimaxm25} & 2026-02 & 230B-A10B & Text & Open \\
GLM-Z1-9B~\cite{zai2025glm40414} & 2025-04 & 9B & Text & Open & GPT-5.4~\cite{openai2026gpt54} & 2026-03 & -- & Multi & Closed \\
Llama 4 Maverick~\cite{meta2025llama4} & 2025-04 & 400B/17B & Multi & Open & Nemotron-Cascade-2~\cite{nvidia2026nemotroncascade2} & 2026-03 & 30B/3B & Code & Open \\
Llama 4 Scout~\cite{meta2025llama4} & 2025-04 & 109B/17B & Multi & Open & GPT-5.3~\cite{openai2026gpt53instant} & 2026-03 & -- & Multi & Closed \\
Seed-Thinking-v1.5~\cite{bytedanceseed2025seedthinking} & 2025-04 & -- & Text & Open & MiniMax-M2.7~\cite{minimax2026minimaxm27} & 2026-03 & 230B-A10B & Text & Open \\
Nemotron-Ultra-253B~\cite{bercovich2025llamanemotron} & 2025-04 & 253B/17B & Text & Open & MiMo-V2.5-Pro~\cite{xiaomi2026mimov25pro} & 2026-04 & -- & Multi & Open \\
ERNIE-4.5-VL~\cite{baidu2025ernie45technicalreport} & 2025-04 & 424B-A47B & Multi & Open & Kimi K2.6~\cite{moonshotai2026kimik26} & 2026-04 & 1T/32B & Multi & Open \\
Codex-1~\cite{openai2025codex} & 2025-05 & -- & Code & Closed & GLM-5.1~\cite{zai2026glm51} & 2026-04 & 754B & Text & Open \\
DeepSeek-R1-0528~\cite{deepseekai2025r10528} & 2025-05 & 671B/37B & Text & Open & DeepSeek-V4~\cite{deepseekai2026deepseekv4} & 2026-04 & 1.6T & Text & Open \\
R1-Distill-Qwen3-8B~\cite{deepseekai2025r10528} & 2025-05 & 8B & Text & Open & Qwen3.6~\cite{qwen2026qwen36} & 2026-04 & 35B/3B+ & Multi & Open \\
R1-Distill-Qwen3-32B~\cite{deepseekai2025r10528} & 2025-05 & 32B & Text & Open & Gemma 4~\cite{google2026gemma4} & 2026-04 & 2B--26B & Multi & Open \\
MiMo-7B-RL~\cite{xiaomimimo2025mimo} & 2025-05 & 7B & Text & Open & GPT-5.5~\cite{openai2026gpt55systemcard} & 2026-04 & -- & Multi & Closed \\
MiMo-7B-RL-0530~\cite{xiaomimimo2025mimo} & 2025-05 & 7B & Text & Open & Claude Opus 4.7~\cite{anthropic2026claudeopus47} & 2026-04 & -- & Multi & Closed \\
Doubao 1.5 Pro Thinking~\cite{bytedance2025doubao15thinking} & 2025-05 & -- & Text & Closed & Claude Mythos Preview~\cite{anthropic2026claudemythospreview} & 2026-04 & -- & Multi & Closed \\
Gemini 2.5 Flash~\cite{google2025gemini25io} & 2025-05 & -- & Multi & Closed & Grok 4.3~\cite{xai2026grok43nonreasoning} & 2026-05 & -- & Multi & Closed \\
InternVL3~\cite{zhu2025internvl3} & 2025-05 & 2B--78B & Multi & Open & Ring-2.6-1T~\cite{inclusionai2026ring26} & 2026-05 & 1T-A63B & Text & Open \\
Devstral~\cite{mistral2025devstral} & 2025-05 & 24B & Code & Open & ERNIE 5.1~\cite{baidu2026ernie51} & 2026-05 & -- & Multi & Closed \\
Claude 4 Sonnet~\cite{anthropic2025claude4systemcard} & 2025-05 & -- & Multi & Closed & Claude Opus 4.8~\cite{anthropic2026claudeopus48} & 2026-05 & -- & Multi & Closed \\
\bottomrule
\end{tabular}
}
\end{table}

Beyond algorithms, the complex interplay between SFT and RL has proven critical: SFT provides stable formatting and cold-start initialization, while RL expands the capability boundary through exploration. DeepSeek-R1 instantiated this synergy through a four-stage pipeline of cold-start SFT, reasoning RL, rejection sampling SFT, and general RL, and Qwen3 further refined it into Long CoT cold-start, reasoning RL, thinking-mode fusion, and general RL, representing the most complete publicly documented hybrid training workflows to date. The overarching trend is clear: training has progressed from single-stage imitation to multi-stage synergy, with open-source frameworks such as OpenRLHF~\cite{hu2024openrlhf} and verl~\cite{sheng2025hybridflow} alongside reproduction initiatives like TinyZero~\cite{pan2025tinyzero} and open-r1~\cite{openr1} enabling the broader academic and developer community to successfully reproduce R1-level reasoning training at relatively modest compute cost \citep{li2025system1system2,stechly2025self,duc2025mathematics,rugaba2026v,gull2025engtrace,yu2024ovm}.

The progression from independent reasoning models toward unified systems has unfolded rapidly across the industry. Models such as o1 \cite{openai2024o1}, R1 \cite{deepseek2025r1}, and QwQ \cite{qwen2025qwq32b} were initially released as dedicated reasoning products separate from general-purpose dialogue systems. This separation dissolved as o3/o4-mini enabled tool invocation during reasoning \cite{openai2025o3o4mini}, Qwen3 introduced hybrid thinking/non-thinking modes \cite{qwen2025qwen3}, and GPT-5 unified reasoning and dialogue through internal routing \cite{openai2025gpt5systemcard}. Open-source frameworks further democratized R1-level training at modest cost \cite{li2025system1system2}. Reasoning has thus transformed from an isolated model family into a tunable capability dimension within general-purpose systems. Table~\ref{tab:thinking_llm} summarizes representative reasoning LLMs that mark this shift from isolated reasoning products to unified cognitive systems. This fusion with tool use and environmental interaction means modern Thinking LLMs already exhibit rudiments of agentic behavior, naturally raising the question of what further architectural support is needed to move from a powerful "brain" to a reliable autonomous agent~\citep{yu2025formalmath,li2023chain,lin2026scaling,alon2025integrating,liu2025revisiting}.

\begin{AIbox}{Trend: From Long CoT to Agentic Decision Cores}
    \begin{itemize}[left=2pt,topsep=1pt,itemsep=2pt, parsep=1pt]
        \item Thinking LLMs shift scaling from parameter growth alone to inference-time computation, using long reasoning traces, alternative exploration, and self-correction to support harder mathematical, coding, and planning tasks.

        \item Reinforcement learning internalizes search and reflection into the model, but real-world autonomy still requires persistent state, environmental feedback, and multi-step execution beyond single-turn reasoning.
    \end{itemize}
\end{AIbox}

\section{Part II: The Evolution of Tool-Augmented Task Execution}
\headingnote{From ``Experimental Tool User'' to ``Workstation Expert''}

This part shifts the focus from the model's internal cognition to its external ability to act.
Once LLMs acquire stronger reasoning capabilities, the next question is whether they can use that cognition to operate tools, interact with environments, and complete external tasks~\citep{autogpt2023,babyagi2023,anthropic2024claude35haiku,mialon2023gaia,mcp2026servers}.
We first revisit the Agent era, where perception, planning, memory, and tool invocation are organized into an environment--action--feedback loop.
We then examine the OpenClaw era, where this loop is embedded into persistent workspaces with reusable skills, stateful execution, and stronger requirements for task closure, reliability, and governance~\citep{singh2025agentic,huang2024understanding,dong2024survey,yehudai2025survey,mamun2026anatomical,liu2026knowledge,dong2025youtu}.

\subsection{The Agent Era: Environment--Action--Feedback Loops}
\headingnote{General Intelligent Body: tool invocation and initial autonomy}

\begin{figure}[t]
    \centering
    \includegraphics[width=\textwidth]{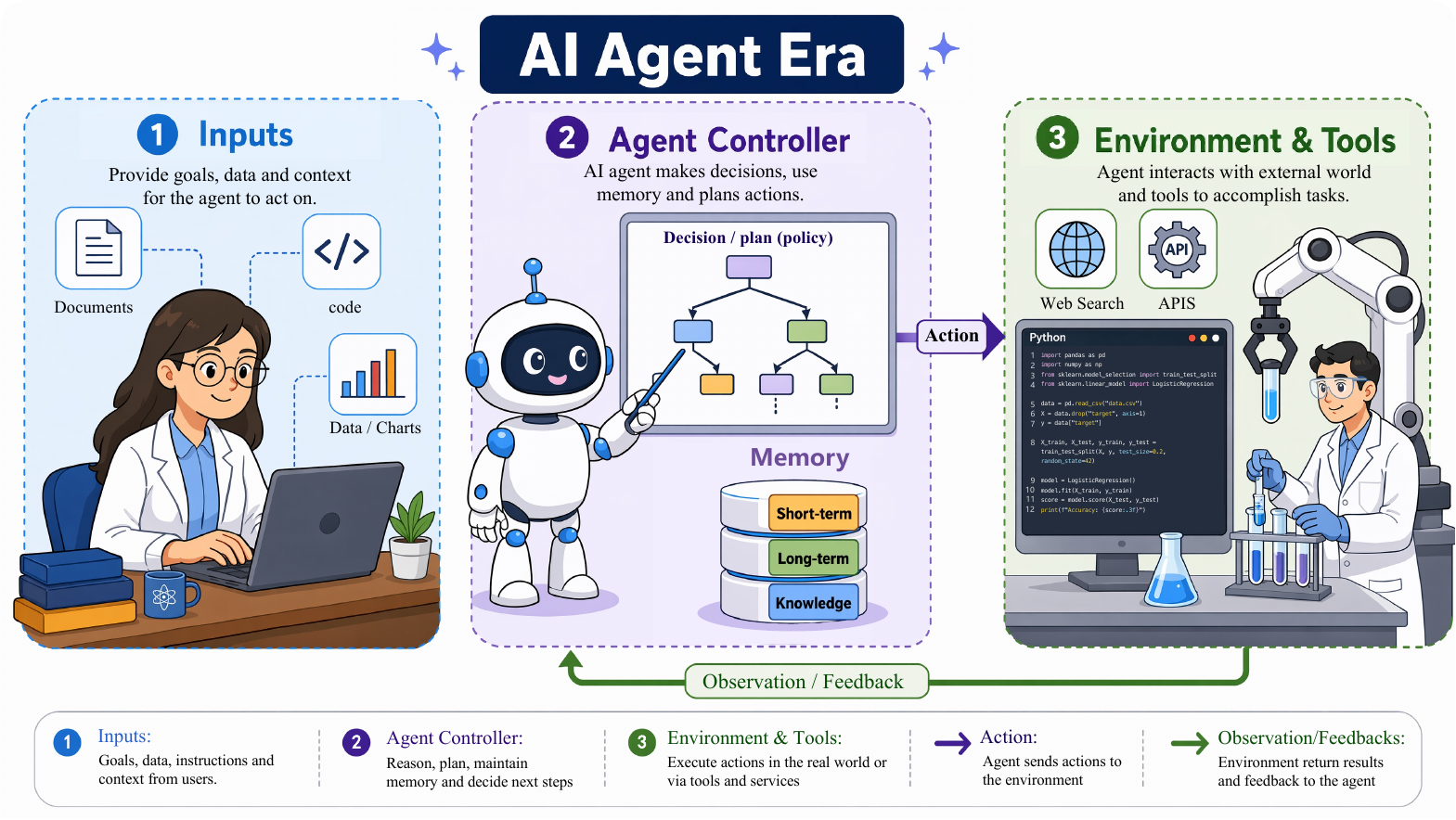}
    \caption{The Agent Era: the model observes an external environment, plans the next step, invokes tools or actions, receives feedback, and iterates toward the task goal. The figure illustrates the observe--think--act--observe loop that gives LLMs an initial form of autonomy beyond single-turn answering.}
    \label{fig:agent}
\end{figure}

As shown in Figure~\ref{fig:agent}, the Agent era marks the first major attempt to turn LLMs from passive conversational systems into active problem solvers.
Instead of producing a single answer and stopping, an agent observes an environment, reasons about the next step, invokes a tool or action, receives feedback, and iterates.
This loop gives LLMs an initial form of autonomy: they can search, call APIs, write code, browse pages, remember intermediate information, and adjust plans from external results~\citep{zhao2023depth,xu2025llm,yang2026toward,xie2024large,cao2025large}.
However, this autonomy remains fragile because the environment is still treated as a sequence of disconnected tool responses rather than as a persistent workspace.
We therefore review the core capabilities that define this era, followed by the evaluation evidence and structural bottlenecks that motivate the transition to OpenClaw-style workstation agents.

\subsubsection{Tool Invocation and Core Agent Capabilities}

\textbf{Agentic Interaction Loop.} The Agent era marks the first organized and systematic effort to extend pretrained LLMs from single-turn question answering into sustained, multi-step interaction with complex external environments. Different from traditional LLMs, the agent operates within an environment loop, observing the world, selecting actions that may invoke external tools, receiving feedback, and iterating. This closed loop of observing, thinking, acting, and observing again is what fundamentally distinguishes an agent from a chatbot. ReAct \cite{yao2022react} established the canonical form of this loop by interleaving \emph{Thought}, \emph{Action}, and \emph{Observation} in an alternating chain, demonstrating that synergizing reasoning and acting outperforms either in isolation on knowledge-intensive and decision-making tasks. The pattern became so influential that nearly every later agent system can be read as an extension, refinement, or specialization of the ReAct loop~\citep{ruan2023tptu,lee2026survey,shen2024llm,li2024survey,NEURIPS2025_48dcc43a,zhang2026latentvisualcachevideo}.

\textbf{Agent Architecture.} Several influential frameworks have sought to formalize what makes a system an agent rather than merely an LLM with tools. Wang et al. \cite{wang2024survey} proposed a four-module architecture consisting of \emph{Profile}, \emph{Memory}, \emph{Planning}, and \emph{Action}, establishing the vocabulary that subsequent work has largely adopted. Xi et al. \cite{xi2023rise} approached the same question from cognitive science, drawing parallels between LLM agents and theories of human cognition. The CoALA framework \cite{sumers2023cognitive} further refined this direction by mapping agent capabilities onto constructs from cognitive psychology, including working memory and long-term memory such as episodic, semantic, and procedural memory, as well as a decision-making cycle. More recently, Buyya et al. \cite{buyya2026agentic} identified six modular dimensions of LLM agents, including \emph{Perception}, \emph{Memory}, \emph{Action}, \emph{Planning}, \emph{Reflection}, and \emph{Learning}, offering a control-theoretic lens through a POMDP formulation that complements the cognitive perspective. Despite terminological differences, these frameworks converge on four essential capabilities. \emph{Perception} refers to observing and interpreting the environment. \emph{Planning} involves decomposing goals and reasoning about action sequences. \emph{Memory} concerns maintaining context and accumulating experience. \emph{Tool use} means invoking external APIs to effect change. We trace these developments below~\citep{li2025review,xu2026evolution,xi2026toolgym,chezelles2024browsergym,kruppweb,wu2026omniflowphysicsgroundedmultimodalagent,11356944,ICLR2024_d53538ba,dong2026deep}.

\textbf{Perception.} An agent's effectiveness is fundamentally bounded by what it can observe. Early agents operated in purely textual environments, parsing structured outputs from APIs or web scrapers. A first generation of multi-modal perception delegated sensory processing to external specialist models orchestrated by the LLM. For example, HuggingGPT \cite{shen2023hugginggpt} treats the LLM as a controller that selects models from the Hugging Face ecosystem for sub-tasks such as image captioning and object detection. Along similar lines, Visual ChatGPT \cite{wu2023visual} chains visual foundation models via a prompt manager to handle diverse visual interactions. Taking a program-synthesis approach, ViperGPT \cite{suris2023vipergpt} generates Python programs that compose vision-API calls for compositional visual reasoning. While effective as proof-of-concept demonstrations, these indirect pipelines suffer from error compounding, since each specialist model introduces its own failure modes and the LLM has no direct access to raw sensory signals. The emergence of powerful vision language models has since enabled agents to perceive environments directly through screenshots. Set-of-Mark (SoM) prompting \cite{yang2023set} overlays numbered markers onto UI elements, enabling GPT-4V to refer to specific interface components by index. A dedicated line of GUI agents has pushed this further: CogAgent \cite{hong2024cogagent} employs a dual-resolution visual encoder for fine-grained UI recognition, ShowUI \cite{lin2025showui} unifies vision, language, and action in a single model that directly outputs UI operations from screenshots, and UI-TARS \cite{qin2025ui} achieves context-aware understanding of both desktop and mobile interfaces through large-scale GUI training. Despite these advances, agent perception remains fragmented: most agents observe one snapshot at a time, without persistent visual working memory across steps and tasks~\citep{guo2026mcp,ahn2026orchestrationbench,ding2026wildclawbench,hu2026agentic,li2026benchmark,koh2024visualwebarena,rawles2025androidworld,wang2024mobile,zhang2025appagent,cheng2024seeclick,10.1145/3690642,dong2026deep, zhang2026ssrb, huang2025deepresearchguard,zhang2026thinkingvideovideogenerators}.

\textbf{Planning.} Planning, broadly defined as the ability to break a complex goal into achievable sub-steps and recover when things go wrong, has progressed through several generations \cite{li2025towards,li2025benchmarking, kuang2026atomic, guo2026evoconfig}. The foundational insight came from Chain-of-Thought (CoT) prompting \cite{wei2022chain}, which demonstrated that generating intermediate reasoning steps dramatically improves multi-step performance and now forms the backbone of most agent planning systems. Building on this foundation, Tree of Thoughts (ToT) \cite{yao2023tree} extended the paradigm by allowing multiple reasoning branches and backtracking, effectively casting planning as a search problem. Graph of Thoughts (GoT) \cite{besta2024graph} generalized this idea further by supporting arbitrary graph topologies over reasoning steps. Complementary to search-based methods is the strategy of task decomposition. Decomposed Prompting \cite{khot2022decomposed} breaks problems into modular sub-problems handled by specialized prompts, while Least-to-Most Prompting \cite{zhou2022least} takes an incremental approach by solving progressively harder sub-problems, using earlier solutions as building blocks. Beyond forward planning, a distinguishing feature of effective agents is their capacity for in-episode learning from failure. Reflexion \cite{shinn2023reflexion} addresses this by generating natural-language reflections on failed attempts and incorporating them into subsequent tries. Similarly, Self-Refine \cite{madaan2023selfrefine} uses iterative self-feedback to improve outputs without external supervision. Taking a more formal approach, Reasoning via Planning (RAP) \cite{hao2023reasoning} brings classical search into the picture by combining Monte Carlo Tree Search with the LLM serving as both world model and value function. More recently, a paradigm shift has emerged through the use of reinforcement learning to train models that autonomously interleave reasoning with tool use at inference time.

\textbf{Memory.} LLMs are fundamentally stateless, as each inference call starts with a blank slate and the only memory available is whatever fits within the context window~\cite{packer2023memgpt,wang2023augmenting,shan2025cognitive,fang2025lightmem,du2026memory}. For agents operating over extended horizons, this poses a severe limitation. A useful cognitive framing distinguishes semantic, episodic, and procedural memory: agents must access external knowledge, preserve task experiences, and reuse learned procedures over time \cite{zhang2025survey,pink2025position}. In practice, agent memory has moved from retrieval-based knowledge access to persistent experience accumulation, multimodal environmental memory, and scalable learned management~\citep{li2026world,zhu2025evolutionary,qin2025aptbench,jiang2026sok,adamenko2025swe}.

Moving beyond retrieval, Generative Agents \cite{park2023generative} introduced the influential memory stream architecture, where agents record observations as timestamped entries and periodically synthesize high-level reflections from low-level experiences. Subsequent work explored more structured forms of long-term memory: MemoryBank \cite{zhong2024memorybank} proposed persistent memory with forgetting mechanisms, while ChatDB \cite{hu2023chatdb} used databases as symbolic external memory accessible via SQL. These systems shift memory from external knowledge access to persistent experience accumulation~\citep{deng2025swe,zhang2026swe,tian2026swe,rashid2025swe,badertdinov2026swe}.

Drawing on cognitive science, the Memory Mechanism Survey \cite{zhang2025survey} provides a systematic taxonomy distinguishing between semantic, episodic, and procedural memory for agents, and the Episodic Memory position paper \cite{pink2025position} further argues that episodic memory is the missing piece for maintaining logical consistency across extended interactions. This framing clarifies that agent memory is not a monolithic storage buffer, but a set of distinct mechanisms for preserving facts, experiences, procedures, and self-consistency~\citep{aleithan2024swe,yang2026swe,wang2025swe,sonwane2026omnicode,garg2025saving}.

As autonomous agents become increasingly multimodal, their memory must also preserve perceptual experience rather than only textual interaction histories. MEIA \cite{liu2024meia} introduces a multimodal environmental memory that stores object-level, spatial, and temporal information for embodied agents. MIRIX \cite{wang2025mirix} proposes a modular multi-agent memory system covering semantic, episodic, procedural, and resource memories across heterogeneous text and visual inputs. M3-Agent \cite{long2025seeing} further studies long-term multimodal memory for agents that perceive visual and auditory streams, showing how such memories can support future reasoning and downstream interaction~\citep{ma2025swe,he2025swe,peng2025swe,han2026swe,prathifkumar2025does}.

More recently, a new generation of production-scale systems has advanced agent memory toward practical deployment. Mem0 \cite{chhikara2025mem0} implements a scalable memory-centric architecture achieving 91\% lower p95 latency and over 90\% token cost savings compared to full-context baselines, and its enhanced variant Mem0g introduces graph-based representations to capture relational structure. Taking a different perspective, A-MEM \cite{xu2026mem} draws inspiration from the Zettelkasten method, enabling agents to build interconnected knowledge networks through dynamic indexing and linking. Perhaps most significantly, reinforcement learning is now being applied to teach agents how to manage their own memory autonomously. MEM1 \cite{zhou2025mem1} enables agents to operate with constant memory across long multi-turn tasks through end-to-end RL, achieving 3.5$\times$ performance improvement while reducing memory usage by 3.7$\times$. Along a complementary line, Memory-R1 \cite{yan2025memory} trains a Memory Manager with structured operations (ADD, UPDATE, DELETE, NOOP) using only 152 training pairs, yet outperforms strong baselines across three benchmarks. Mem-$\alpha$ \cite{wang2025mem} and MemMachine \cite{wang2026memmachine} represent further points in this design space, with the former learning memory construction via RL and the latter preserving factual integrity through sentence-level episode storage. The trend is clear: memory is evolving from static retrieval and heuristic storage into an adaptive capability for knowledge access, multimodal experience, and long-horizon context management~\citep{applis2025unified,lai2024autowebglm,patel2024large,thil2024navigating,anupam2025browserarena}.

\textbf{Tool Use.} If perception is the agent's eyes and systematic planning its brain, then tool use is its hands: it is the mechanism through which large language models (LLMs) translate internal reasoning into tangible external operations. The paradigm shift of tool use in LLM agents is understood as a progression from executable calls to large-scale API grounding, long-horizon trajectory-level control, and standardized tool infrastructures~\citep{chae2025web,krupp2025quantifying,xu2025turkingbench,murty2024nnetnav,lu2024weblinx,wang2024executable,chen2023fireact}.

The first problem is \emph{executable tool use}: how can a large language model move beyond textual answers and produce operations that can be successfully executed by an external system? Early tool-integrated reasoning methods established this crucial bridge between language generation and execution. Toolformer~\cite{schick2023toolformer} proposed a self-supervised approach that augments text with tool calls at positions where external tools improve prediction, enabling the model to learn when and how to invoke tools. PAL~\cite{gao2023pal} and Program of Thoughts (PoT)\cite{chen2022program} take code execution as a meta-tool: the model generates executable programs while exact computation is delegated to an interpreter. These works mark the first step beyond pure text generation: the model can effectively externalize its reasoning into an executable substrate\citep{cai2025large,song2025bearcubs,zhang2024large,caples2026real,bhathal2025websight}.

Once successful tool invocation becomes possible, the primary bottleneck shifts from \emph{whether} a model can call a tool to \emph{which} tool it should call and \emph{how} the call should be constructed. This gives rise to the problem of \emph{API grounding at scale}. In complex realistic environments, tools are not a small hand-written set of functions but large API ecosystems with complex documentation, argument schemas, and usage constraints. Gorilla~\cite{patil2023gorilla} addresses this problem by fine-tuning LLMs on API documentation, enabling them to generate accurate calls for both familiar and unseen endpoints. ToolLLM~\cite{qin2023toolllm} scales this direction with ToolBench, a benchmark containing over 16,000 real-world RESTful APIs across 49 categories, and introduces search-based decision procedures for selecting and chaining tools. ToolACE~\cite{liu2025toolace} further shows that high-quality synthetic tool-calling data can produce compact language models with strong zero-shot function-calling ability. This work reframes tool use as interface grounding: the model must map intent to the right API, respect schemas, and produce valid arguments~\citep{pan2024webcanvas,trabucco2025towards,hu2025agentgen,wang2023describe,cheng2024exploring,zhuang2023toolqa,huang2024metatool,guo2024stabletoolbench,du2024anytool,song2023restgpt,miao2025recode}.

However, correct local calls do not by themselves produce reliable agents. In multi-step tasks, tool use becomes a \emph{trajectory-level control} problem. The agent must decide when external execution is necessary, avoid unnecessary calls, incorporate tool feedback, recover from failed executions, and stop when the task is complete. SMART~\cite{qian2025smart} studies this issue from the perspective of tool overuse, training agents to balance parametric reasoning with external tool dependence. START~\cite{li2025start} shows that tool-integrated reasoning traces can be bootstrapped through hint-based self-learning rather than relying only on manually written demonstrations. Reinforcement-learning-based methods push this direction further by treating tool invocation as part of the rollout environment. ReTool~\cite{feng2025retool} trains models to interleave natural-language reasoning with real-time code execution, while ToRL~\cite{li2025torl} scales tool-integrated reinforcement learning with code interpreters inside the rollout process. ToolRL~\cite{qian2026toolrl} highlights the importance of reward design, especially the granularity and temporal structure of rewards for tool selection and application. ARTIST~\cite{singh2505agentic}, Tool-Star~\cite{dong2505tool}, and AutoTIR~\cite{wei2025autotir} extend this line toward multi-turn or multi-tool settings, where models coordinate reasoning, invocation, and execution feedback within one trajectory. The central shift is from isolated tool calls to policies that control tool use over time~\citep{xi2025agentgym,mahdavi2024leveraging,li2025exploring,gao2024large,birr2024autogpt+,basu2024api,ye2025tooleyes,gao2024confucius,wang2024gta}.

As tool-augmented agents move toward commercial deployment, the primary bottleneck shifts again from model-side tool use to infrastructure-level standardization. Fragmented tool interfaces make agents difficult to scale, secure, and govern, because each tool may expose different schemas, authentication mechanisms, context formats, and execution constraints. The Model Context Protocol (MCP)\cite{mcp2026tools}, introduced by Anthropic in November 2024, represents a coordinated effort to standardize how LLMs connect to external tools and data sources, analogous to how common peripheral protocols standardize hardware connectivity. Since its release, both OpenAI and Google have announced MCP support, signaling a trend toward industry-wide convergence. In parallel, major commercial LLM APIs now widely support native function calling as a first-class interface, moving tool invocation from ad-hoc prompting toward structured execution\citep{hu2023language,babu2025adaptive,li2025task,wang2023describe,huang2025recommender}.

The overall trajectory is therefore not simply from fewer tools to more tools. Rather, tool use evolves from executable calls to grounded API use, from locally valid calls to trajectory-level tool-use policies, and from fragmented interfaces to standardized tool infrastructures. This progression also exposes the boundary of the Agent era. Tool use gives agents the ability to act, but the effects of these actions often remain fragmented across isolated calls and transient tool responses. The next step, developed in the OpenClaw era, is to embed tool use inside persistent workspaces where files, sessions, skills, logs, permissions, and verification procedures can support durable task closure in real deployments.

\subsubsection{Initial Autonomy: Evaluation and Structural Bottlenecks}

To assess the practical reliability of LLM-based agents, several benchmarks have been developed. AgentBench \cite{liu2023agentbench} evaluates agents across eight environments, revealing a performance gap between commercial and open-source models in multi-step settings. WebArena \cite{zhou2023webarena} uses realistic web environments where GPT-4 achieved only ~14\% success. SWE-bench \cite{jimenez2023swebench} focuses on GitHub issue resolution for coding, while GAIA \cite{mialon2023gaia} tests general assistants on multi-step reasoning and tool use. Across these benchmarks, success rates decay super-linearly with complexity and horizon length, prompting systematic failure analysis. The LLM Agent Failure study \cite{roig2025llms} identifies four archetypes: premature ungrounded action, over-helpfulness with plausible but incorrect details, distractor-induced context pollution, and fragile execution under load. Similarly, the Agent Hallucination survey \cite{lin2025llm} notes that "hallucinated actions"—such as calling incorrect APIs or operating wrong files—cause irreversible failures, unlike text-level hallucinations~\citep{he2024webvoyager,yoran2024assistantbench,wu2024copilot,zhang2025ufo,huang2023mlagentbench, kuang2025process, ye2025productagent}.

Synthesizing the empirical evidence from existing benchmarks and failure analyses, we identify four critical structural bottlenecks of the Agent era:
\begin{itemize}
    \item \textbf{Fragmented perception.} Agents observe the environment through narrow, episodic windows---a single API response, a single screenshot, a single tool output. They lack a persistent, holistic model of the environment's state and how it evolves over time.
    \item \textbf{Ephemeral tool invocation.} Each tool call is an isolated transaction. No stable workspace exists where intermediate artifacts persist: files created in one step may be inaccessible in the next; terminal sessions are not preserved; browser state is lost between actions.
    \item \textbf{Brittleness under environmental uncertainty.} Real-world environments are noisy, asynchronous, and adversarial. Network timeouts, UI changes, API responses, and permission errors compound across long sequences, sharply lowering success rates.
    \item \textbf{Absence of long-term task closure.} Agents can attempt tasks but rarely complete them reliably end-to-end. They lack the persistent state, error recovery mechanisms, and verification loops needed to deliver finished work products rather than best-effort attempts.
\end{itemize}

These bottlenecks are not mere engineering issues solved by prompting or scaling, but a fundamental architectural limitation: current agents treat environments as external oracles to query, rather than persistent workspaces to inhabit. Overcoming this requires shifting from tool-calling agents to agents working inside workstations. Here, we strictly define LLM-based agents as systems operating through an environment-action-feedback loop with external tools. Earlier techniques—like chain-of-thought, retrieval-augmented generation, and memory—are not agents themselves, but foundational capabilities. Table~\ref{tab:agent_era} summarizes both agent systems and these enabling capabilities.

\begin{table}[htbp]
\centering
\scriptsize
\setlength{\tabcolsep}{8pt}
\renewcommand{\arraystretch}{1.1}
\caption{Representative works related to LLM-based agents and their enabling capabilities.}
\label{tab:agent_era}
\resizebox{\columnwidth}{!}{
\begin{tabular}{@{} l l l l l @{}}
\toprule
\textbf{Work} & \textbf{Year} & \textbf{Category} & \textbf{Role} & \textbf{Key Contribution} \\
\midrule

ReAct~\cite{yao2022react} & 2022 & Agent Architecture & Agent Framework & Thought--Action--Observation loop \\
Wang et al.~\cite{wang2024survey} & 2024 & Agent Architecture & Conceptual Framework & Profile/Memory/Planning/Action architecture \\
Xi et al.~\cite{xi2023rise} & 2023 & Agent Architecture & Conceptual Framework & Cognitive-science perspective on agents \\
CoALA~\cite{sumers2023cognitive} & 2023 & Agent Architecture & Conceptual Framework & Cognitive psychology mapping for agents \\
Buyya et al.~\cite{buyya2026agentic} & 2026 & Agent Architecture & Conceptual Framework & Six dimensions with POMDP formulation \\
HuggingGPT~\cite{shen2023hugginggpt} & 2023 & Perception & Agent System & LLM orchestrating Hugging Face models \\
Visual ChatGPT~\cite{wu2023visual} & 2023 & Perception & Agent System & Chaining visual models via prompt manager \\
ViperGPT~\cite{suris2023vipergpt} & 2023 & Perception & Agent System & Program synthesis for visual reasoning \\
Set-of-Mark~\cite{yang2023set} & 2023 & Perception & Enabling Capability & Numbered markers for UI grounding \\
CogAgent~\cite{hong2024cogagent} & 2024 & Perception & Agent Model & Dual-resolution encoder for UI recognition \\
ShowUI~\cite{lin2025showui} & 2025 & Perception & Agent Model & Unified vision--language--action for UI ops \\
UI-TARS~\cite{qin2025ui} & 2025 & Perception & Agent Model & Context-aware desktop/mobile GUI understanding \\
CoT~\cite{wei2022chain} & 2022 & Planning & Enabling Capability & Intermediate reasoning steps \\
ToT~\cite{yao2023tree} & 2023 & Planning & Enabling Capability & Multi-branch reasoning with backtracking \\
GoT~\cite{besta2024graph} & 2024 & Planning & Enabling Capability & Graph topologies over reasoning steps \\
Decomposed Prompting~\cite{khot2022decomposed} & 2022 & Planning & Enabling Capability & Modular sub-problem decomposition \\
Least-to-Most~\cite{zhou2022least} & 2022 & Planning & Enabling Capability & Incremental sub-problem solving \\
Reflexion~\cite{shinn2023reflexion} & 2023 & Planning & Agent Framework & Language reflections on failed attempts \\
Self-Refine~\cite{madaan2023selfrefine} & 2023 & Planning & Enabling Capability & Iterative self-feedback improvement \\
RAP~\cite{hao2023reasoning} & 2023 & Planning & Enabling Capability & MCTS with LLM as world model \\
Search-R1~\cite{jin2025search} & 2025 & Planning & Agent Model & RL-trained search in reasoning chains \\
R1-Searcher~\cite{song2025r1} & 2025 & Planning & Agent Model & Two-stage RL for search--reasoning \\
ReSearch~\cite{chen2503research} & 2025 & Planning & Agent Model & Supervision-free RL with search ops \\
RAG~\cite{lewis2020retrieval} & 2020 & Memory & Enabling Capability & Retrieval-augmented generation \\
Self-RAG~\cite{asai2024self} & 2024 & Memory & Enabling Capability & Adaptive retrieval with self-reflection \\
CRAG~\cite{yan2024corrective} & 2024 & Memory & Enabling Capability & Retrieval evaluation and correction \\
Adaptive-RAG~\cite{jeong2024adaptive} & 2024 & Memory & Enabling Capability & Strategy routing by query complexity \\
Generative Agents~\cite{park2023generative} & 2023 & Memory & Agent System & Memory stream with reflections \\
MemoryBank~\cite{zhong2024memorybank} & 2024 & Memory & Enabling Capability & Persistent memory with forgetting \\
ChatDB~\cite{hu2023chatdb} & 2023 & Memory & Enabling Capability & Database as external memory via SQL \\
MEIA~\cite{liu2024meia} & 2024 & Memory & Agent System & Multimodal environmental memory \\
MIRIX~\cite{wang2025mirix} & 2025 & Memory & Memory Infrastructure & Multi-agent multimodal memory system \\
M3-Agent~\cite{long2025seeing} & 2025 & Memory & Agent System & Long-term visual-auditory memory \\
Memory Survey~\cite{zhang2025survey} & 2025 & Memory & Conceptual Framework & Semantic/episodic/procedural taxonomy \\
Episodic Memory~\cite{pink2025position} & 2025 & Memory & Conceptual Framework & Position paper on episodic consistency \\
Mem0~\cite{chhikara2025mem0} & 2025 & Memory & Memory Infrastructure & Scalable memory; graph-based Mem0g \\
A-MEM~\cite{xu2026mem} & 2026 & Memory & Memory Infrastructure & Zettelkasten-inspired knowledge networks \\
MEM1~\cite{zhou2025mem1} & 2025 & Memory & Agent Model & End-to-end RL for constant memory \\
Memory-R1~\cite{yan2025memory} & 2025 & Memory & Agent Model & RL-trained structured memory manager \\
Mem-$\alpha$~\cite{wang2025mem} & 2025 & Memory & Agent Model & Memory construction via RL \\
MemMachine~\cite{wang2026memmachine} & 2026 & Memory & Memory Infrastructure & Sentence-level episode storage \\
Toolformer~\cite{schick2023toolformer} & 2023 & Tool Use & Agent Model & Self-supervised tool-call learning \\
PAL~\cite{gao2023pal} & 2023 & Tool Use & Enabling Capability & Code execution as meta-tool \\
PoT~\cite{chen2022program} & 2022 & Tool Use & Enabling Capability & Multi-step reasoning via programming \\
Gorilla~\cite{patil2023gorilla} & 2024 & Tool Use & Agent Model & Fine-tuning on API documentation \\
ToolLLM~\cite{qin2023toolllm} & 2023 & Tool Use & Agent Framework & 16K+ APIs; DFS-based decision trees \\
SMART~\cite{qian2025smart} & 2025 & Tool Use & Agent Model & Tool overuse mitigation \\
START~\cite{li2025start} & 2025 & Tool Use & Agent Model & Self-taught tool-integrated reasoning \\
ReTool~\cite{feng2025retool} & 2025 & Tool Use & Agent Model & RL for strategic tool invocation \\
ToRL~\cite{li2025torl} & 2025 & Tool Use & Agent Model & Tool-integrated RL with code execution \\
ToolRL~\cite{qian2026toolrl} & 2026 & Tool Use & Agent Model & Reward design for tool learning \\
ARTIST~\cite{singh2505agentic} & 2025 & Tool Use & Agent Model & Agentic reasoning with tool integration \\
Tool-Star~\cite{dong2505tool} & 2025 & Tool Use & Agent Model & Multi-tool reasoning via RL \\
AutoTIR~\cite{wei2025autotir} & 2025 & Tool Use & Agent Model & Autonomous tool-integrated reasoning \\
MCP~\cite{mcp2026tools} & 2024 & Tool Use & Infrastructure & Universal LLM--tool connectivity protocol \\
ToolACE~\cite{liu2025toolace} & 2025 & Tool Use & Agent Model & Synthesized data for compact tool callers \\
AgentBench~\cite{liu2023agentbench} & 2023 & Benchmark & Evaluation & 8-environment agent evaluation \\
WebArena~\cite{zhou2023webarena} & 2023 & Benchmark & Evaluation & Realistic web task environments \\
SWE-bench~\cite{jimenez2023swebench} & 2024 & Benchmark & Evaluation & GitHub issue resolution benchmark \\
GAIA~\cite{mialon2023gaia} & 2023 & Benchmark & Evaluation & Multi-step reasoning \& tool use QA \\
Agent Failure~\cite{roig2025llms} & 2025 & Benchmark & Evaluation & Four archetypal agent failure modes \\
Agent Hallucination~\cite{lin2025llm} & 2025 & Benchmark & Evaluation & Hallucinated actions \& system failures \\

\bottomrule
\end{tabular}
}
\end{table}

\begin{AIbox}{Key Difference: From Tool Calls to Agentic Loops}
    \begin{itemize}[left=2pt,topsep=1pt,itemsep=2pt, parsep=1pt]
        \item The Agent era extends LLMs beyond single-turn text generation by organizing perception, planning, memory, and tool use into an environment--action--feedback loop.

        \item Its limitation is that tool calls remain fragmented and ephemeral: agents can attempt multi-step tasks, but lack persistent state, robust recovery, and reliable end-to-end closure.
    \end{itemize}
\end{AIbox}

\subsection{The OpenClaw Era: Persistent Workspaces for Task Closure}
\headingnote{Workspace intelligence: workspace hosting and task closure}

\begin{figure}[t]
    \centering
    \includegraphics[width=\textwidth]{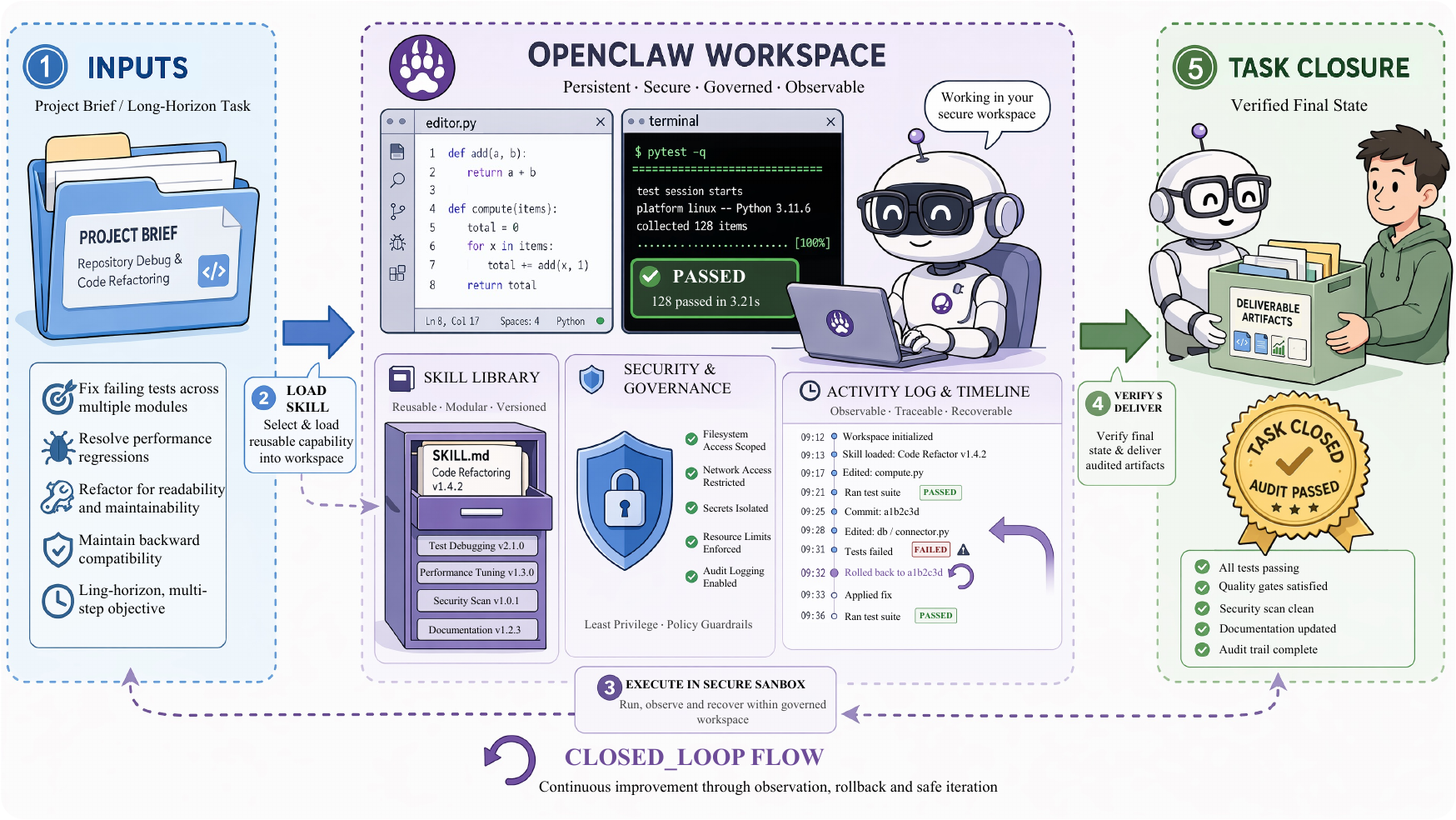}
    \caption{The OpenClaw Era: the agent works inside a persistent workspace with files, terminals, browsers, logs, permissions, reusable skills, and verification loops. The figure illustrates how workspace state and skill-based execution turn fragmented tool use into inspectable, recoverable, and deliverable task closure.}
    \label{fig:openclaw}
\end{figure}

\paragraph{Boundary from the Agent Era to the OpenClaw Era.}
As shown in Figure~\ref{fig:openclaw}, the boundary between the Agent era and the OpenClaw era is not simply whether a model can call tools.
A system belongs to the Agent era when its minimal architecture is an environment--action--feedback loop: it observes a state, reasons about the next move, invokes an external tool or action, and incorporates the returned observation into the next step.
This definition captures the first break from chatbot-style response generation, but it does not guarantee durable state, reusable procedures, recoverable execution, or final-state verification.
The environment remains something the agent queries from the outside.

By contrast, the OpenClaw era begins when the environment becomes a persistent host.
The defining condition is that agents work inside a managed workspace where files, sessions, logs, tools, project instructions, permissions, and reusable skills persist across the trajectory.
In this setting, actions are not merely tool calls; they are workspace operations whose effects can be inspected, validated, rolled back, and governed.
The conceptual unit therefore shifts from an agent that attempts a sequence of actions to a workstation system delivering a correct, auditable final state.
Table~\ref{tab:agent_openclaw_boundary} summarizes this.

\begin{table}[t]
\centering
\caption{Boundary between the Agent Era and the OpenClaw Era.}
\label{tab:agent_openclaw_boundary}
\begingroup
\footnotesize
\renewcommand{\arraystretch}{1}
\resizebox{\columnwidth}{!}{
\begin{tabular}{@{} p{0.18\columnwidth} p{0.3\columnwidth} p{0.4\columnwidth} @{}}
\toprule
\textbf{Dimension} & \textbf{Agent Era} & \textbf{OpenClaw Era} \\
\midrule
Organizing abstraction & Environment--action--feedback loop & Persistent workspace for task hosting \\
Unit of action & Tool call or external API invocation & Workspace operation over files, terminals, browsers, services, and skills \\
State model & Episodic observations and short-horizon memory & Durable files, sessions, logs, repositories, local memory, and snapshots \\
Knowledge reuse & Prompt patterns, retrieved memory, or ad-hoc demonstrations & Reusable skill packages with instructions, scripts, dependencies, examples, and checks \\
Task objective & Produce useful intermediate actions or responses & Deliver a correct, inspectable, and recoverable final workspace state \\
Evaluation focus & Action correctness and trajectory success rate & Task closure, final-state verification, repeatability, and auditability \\
Failure recovery & Best-effort retry or prompt-level reflection & Structured verification, rollback, rerun, sandboxing, and state repair \\
Safety boundary & Prompt-level guardrails and tool-use policies & Runtime permissions, provenance tracking, audit logs, and governance over workspace changes \\
\bottomrule
\end{tabular}
}
\endgroup
\end{table}

The OpenClaw era denotes the point at which agent research becomes inseparable from the workstation in which the agent is deployed.
Earlier agents mainly demonstrated that LLMs could reason, choose tools, and react to observations.
OpenClaw-style systems instead make the workspace itself the organizing abstraction: the agent is connected to persistent files, terminals, browsers, messaging channels, credentials, project instructions, local memory, and reusable skills \cite{openclaw2026repo}.
The result is a shift from \emph{tool use} to \emph{task hosting}.
An agent is no longer evaluated only by whether it emits a plausible next action, but by whether it can leave a durable, inspectable, and safe final state in a real software environment.
We therefore summarize this era along two axes: the emergence of workspace intelligence and skill-based task closure, and the new reliability, verification, and governance problems created by persistent computer use.
Table~\ref{tab:openclaw_era} summarizes representative works that define these workspace, skill, task-closure, evaluation, reliability, and governance dimensions.

\subsubsection{Workspace Intelligence and Skill-Based Task Closure}

The central architectural change is the move from isolated tool invocation to situated work inside a persistent workstation.
OpenClaw is best read as a representative engineering manifestation of this transition rather than as its sole conceptual origin: it packages a local personal assistant around a gateway, workspace, communication channels, prompt files, skills, and tool integrations \cite{openclaw2026repo}.
Related software-engineering agents show why this matters.
OpenHands provides an open platform in which agents edit code, run shell commands, browse, and execute programs inside controlled development environments \cite{wang2024openhands}.
SWE-agent further argues that the agent-computer interface is itself a decisive design object: repository navigation, file editing, command execution, and test feedback must be shaped for language-model agents rather than inherited unchanged from human-facing tools \cite{yang2024sweagent}.
Together, these systems make clear that the next step after API calling is not merely adding more tools, but constructing a stable workbench in which intermediate artifacts, environmental state, and verification signals can persist across the trajectory.

\begin{table}[t]
\centering
\scriptsize
\setlength{\tabcolsep}{1pt}
\renewcommand{\arraystretch}{1.3}
\caption{Representative works related to the OpenClaw era and workspace-level task execution.}
\label{tab:openclaw_era}
\resizebox{\columnwidth}{!}{
\begin{tabular}{@{} l l l l l @{}}
\toprule
\textbf{Work} & \textbf{Year} & \textbf{Category} & \textbf{Role} & \textbf{Key Contribution} \\
\midrule

OpenClaw~\cite{openclaw2026repo} & 2026 & Workspace & Agent Framework & Persistent workspace with tools, channels, and skills \\
OpenHands~\cite{wang2024openhands} & 2024 & Workspace & Agent Platform & Code editing, shell execution, browsing in controlled environments \\
SWE-agent~\cite{yang2024sweagent} & 2024 & Workspace & Agent-Computer Interface & Repository navigation and test-feedback interface for agents \\
SemaClaw~\cite{zhu2026semaclaw} & 2026 & Workspace & Harness Architecture & Auditable execution substrate decoupled from UI surfaces \\
Sema Code~\cite{wang2026semacode} & 2026 & Workspace & Coding Harness & Controllable workspace harness for software agents \\
Voyager~\cite{wang2023voyager} & 2023 & Skill & Agent System & Executable skill library learned from environment feedback \\
Anthropic Agent Skills~\cite{anthropic2026skillsdocs,anthropic2026skillsrepo} & 2026 & Skill & Skill Infrastructure & Folder-based skills with instructions, scripts, and resources \\
OpenClaw Skills~\cite{openclaw2026skillsdocs,openclaw2026skillsrepo} & 2026 & Skill & Skill Infrastructure & Workspace-local \texttt{SKILL.md} packages for reusable procedures \\
Awesome OpenClaw Skills~\cite{voltagent2026awesomeopenclawskills} & 2026 & Skill & Skill Repository & Public registry of reusable OpenClaw skill packages \\
Agent Skills Analysis~\cite{ling2026agentskills} & 2026 & Skill & Conceptual Analysis & Composable capability packages for data-driven agents \\
SkillFortify~\cite{bhardwaj2026skillfortify} & 2026 & Skill & Security Analysis & Metadata, dependency, provenance, and sandboxing requirements \\
SWE-bench~\cite{jimenez2023swebench} & 2024 & Task Closure & Benchmark & Real GitHub issue resolution verified by tests \\
Terminal-Bench~\cite{merrill2026terminalbench} & 2026 & Task Closure & Benchmark & Long-horizon command-line task execution \\
OSWorld~\cite{xie2024osworld} & 2024 & Evaluation & Benchmark & Real OS tasks with execution-based checking scripts \\
WebArena~\cite{zhou2023webarena} & 2023 & Evaluation & Benchmark & Realistic web environments with stateful task success \\
WorkArena~\cite{drouin2024workarena} & 2024 & Evaluation & Benchmark & Enterprise-software workflow evaluation \\
TheAgentCompany~\cite{theagentcompany2024} & 2024 & Evaluation & Benchmark & Simulated software-company work tasks \\
Reliability of Computer-Use Agents~\cite{reliability2026computeruse} & 2026 & Reliability & Reliability Study & Repeated-run instability in computer-use agents \\
Science of Agent Reliability~\cite{rabanser2026sciencereliability} & 2026 & Reliability & Reliability Agenda & Consistency, robustness, recoverability, and error severity \\
Verifiers for Computer-Use Agents~\cite{rosset2026verifiers} & 2026 & Reliability & Verification Framework & Trajectory-wide process and outcome verification \\
Your Agent Can Hurt You~\cite{wang2026youragent} & 2026 & Security & Threat Analysis & Capability, identity, and knowledge poisoning risks \\
Systematic OpenClaw Security~\cite{wang2026openclawsec} & 2026 & Security & Security Evaluation & Runtime risks beyond isolated model behavior \\
Taming OpenClaw~\cite{deng2026tamingopenclaw} & 2026 & Governance & Lifecycle Analysis & Security risks across initialization, reasoning, and execution \\
Don't Let the Claw Grip Your Hand~\cite{shan2026dontlettheclaw} & 2026 & Governance & Defense Framework & OpenClaw-specific threat modeling and defense \\
OS-Harm~\cite{kuntz2026osharm} & 2026 & Security & Safety Benchmark & Misuse, prompt injection, exfiltration, and system harm \\
OpenClaw PRISM~\cite{li2026prism} & 2026 & Governance & Defense Layer & Defense-in-depth over the OpenClaw lifecycle \\
ClawGuard~\cite{zhao2026clawguard} & 2026 & Governance & Runtime Guardrail & File, command, network, and skill boundary enforcement \\
Agentic Forensics~\cite{gruber2026agenticforensics} & 2026 & Governance & Forensics Framework & Trace reconstruction across nondeterministic agent loops \\

\bottomrule
\end{tabular}
}
\end{table}

A second defining feature is the rise of skills as modular, reusable units of agent capability.
The underlying idea predates OpenClaw: Voyager demonstrated that agents can build and reuse an executable skill library from environmental feedback \cite{wang2023voyager}.
What changes in contemporary workstation agents is that skills become file-system-level packages rather than only memories or prompt snippets.
Anthropic's Agent Skills formalize this pattern as folders containing a \texttt{SKILL.md} file, instructions, scripts, and resources that can be dynamically loaded only when relevant \cite{anthropic2026skillsdocs,anthropic2026skillsrepo}.
OpenClaw adopts a similar operational pattern through workspace-local skills organized around \texttt{SKILL.md} files and shared public skill repositories \cite{openclaw2026skillsdocs,openclaw2026skillsrepo,voltagent2026awesomeopenclawskills}.
Recent empirical analysis of Claude skills frames this development as a data-driven shift from monolithic agents toward composable capability packages \cite{ling2026agentskills}.
However, skill modularity also changes the trust model: reusable skills can encode domain expertise, but they can also become stale, over-specific, incompatible, or malicious.
Formal and supply-chain analyses of agentic skills therefore treat metadata, dependencies, versioning, provenance, and sandboxing as first-class requirements rather than engineering details \cite{bhardwaj2026skillfortify}.

The third feature is closed-loop task closure.
A workstation agent must not only plan a trajectory but also inspect the environment after each action, repair failures, rerun commands, validate outputs, and produce deliverable artifacts.
This makes reliability an emergent property of the whole harness: model, workspace, tools, skills, permission boundaries, verification scripts, and recovery policies.
Coding and terminal benchmarks make this requirement concrete.
SWE-bench evaluates whether agents can modify real repositories and pass tests for GitHub issues \cite{jimenez2023swebench}, while Terminal-Bench targets hard, realistic command-line tasks where success depends on sustained shell interaction and environment management \cite{merrill2026terminalbench}.
SemaClaw and Sema Code explicitly formulate this trend as harness engineering: the agent should be decoupled from any single user interface and embedded into a controllable, auditable execution substrate that can power IDEs, CLIs, and multi-channel personal assistants \cite{zhu2026semaclaw,wang2026semacode}.
In this sense, OpenClaw's significance lies less in a new reasoning algorithm than in productizing a complete workstation stack around the model.

\subsubsection{Evaluation, Reliability, and Governance Challenges}

Workspace intelligence also fundamentally changes what counts as evaluation. A useful workstation agent must leave the target environment in a correct, verifiable final state, not merely generate plausible reasoning. OSWorld evaluates multimodal autonomous agents in real operating-system environments with desktop applications, file operations, browsers, and execution-based checking scripts \cite{xie2024osworld}. WebArena and WorkArena extend this state-based evaluation perspective to realistic websites and complex enterprise software workflows \cite{zhou2023webarena,drouin2024workarena}. TheAgentCompany pushes the same idea toward a simulated software company environment, where agents must browse the web, write code, run programs, and communicate with coworkers to complete consequential professional work tasks \cite{theagentcompany2024}. These benchmarks show that the frontier research problem is no longer simply language understanding or isolated, single-step tool selection; it is robust long-horizon task execution in environments that are asynchronous, stateful, and only partially observable.

Recent reliability work sharpens this point.
Computer-use agents may succeed once and fail on a repeated run because execution is stochastic, task specifications are underspecified, and small environmental changes compound over long trajectories \cite{reliability2026computeruse}.
A broader reliability agenda therefore argues that single success rates should be decomposed into consistency, robustness, predictability, recoverability, and bounded error severity \cite{rabanser2026sciencereliability}.
Verification becomes a bottleneck of its own: the system must decide whether the final state actually satisfies the user intent, whether the process was safe, and whether success occurred for the right reason.
Verifier design for computer-use agents consequently emphasizes trajectory-wide evidence, the separation of process and outcome signals, and the distinction between controllable and uncontrollable failures \cite{rosset2026verifiers}.
For OpenClaw-style assistants, these findings imply that evaluation should include repeated execution, perturbation tests, state inspection, and post-hoc auditability, not only one-off demonstrations.

The same persistent workspace that enables task closure also expands the attack surface.
OpenClaw-style agents may hold credentials, local files, identity tokens, tool permissions, communication channels, and long-term memory.
Recent safety analyses show that poisoning an agent's capabilities, identity, or knowledge can turn useful autonomy into attacker-controlled authority \cite{wang2026youragent}.
Systematic security evaluations of OpenClaw and its variants further report that agentized runtimes can be riskier than the underlying models in isolation, because persistent context, orchestration, and multi-step execution amplify model weaknesses into concrete system-level failures \cite{wang2026openclawsec}.
Taming OpenClaw analyzes these risks through the lifecycle of autonomous agents, including initialization, input handling, reasoning, decision making, and execution \cite{deng2026tamingopenclaw}; Don't Let the Claw Grip Your Hand similarly proposes a defense framework for OpenClaw-specific threats \cite{shan2026dontlettheclaw}.
Beyond OpenClaw, OS-Harm shows that computer-use agents must be evaluated for misuse, indirect prompt injection, data exfiltration, and system-level harm rather than only helpfulness \cite{kuntz2026osharm}.

The security response is therefore moving from prompt-level safety toward runtime governance.
OpenClaw PRISM proposes a defense-in-depth layer over the agent lifecycle, including message ingress, prompt construction, tool execution, tool-result persistence, outbound messaging, sub-agent spawning, and gateway startup \cite{li2026prism}.
ClawGuard takes a tool-boundary view, checking file, command, network, and skill operations against user-confirmed constraints and audit policies before execution \cite{zhao2026clawguard}.
These defenses reflect a broader lesson: once an agent can act through a workstation, policy must be enforced where actions become real, not merely expressed as natural-language instructions.
Forensics becomes part of the same agenda.
Because a personal agent can modify files, invoke services, update memory, and choose tools nondeterministically over time, investigations must reconstruct traces across the entire agent loop rather than inspect a single prompt-response pair \cite{gruber2026agenticforensics}.

In summary, the OpenClaw era fundamentally reframes agentic AI as situated task execution.
Its core technological contribution is not simply broader tool access, but the integration of persistent workspaces, modular skills, closed-loop execution, verification, and runtime governance into a single harness.
The next generation of autonomous agents will therefore be differentiated not only by model scale but by the maturity of their underlying workspace stack: state management, skill provenance, permission control, repeated-execution reliability, audit trails, and safety enforcement.

\begin{AIbox}{Key Difference: From Tool Use to Task Hosting}
    \begin{itemize}[left=2pt,topsep=1pt,itemsep=2pt, parsep=1pt]
        \item OpenClaw-style systems shift the organizing abstraction from isolated API calls to persistent workspaces containing files, terminals, browsers, credentials, memory, and reusable skills.

        \item This transition makes task closure, verification, skill provenance, permission control, and runtime governance central requirements for building reliable workstation agents.
    \end{itemize}
\end{AIbox}

\section{Part III: Why ``Workspace + Skill'' Is the Key Leap}
\headingnote{From Tool Use to Reusable Digital Work}

The preceding parts describe two necessary but incomplete advances: stronger cognitive cores and more capable tool-using agents.
This part argues that the next qualitative leap comes from combining reusable skills with persistent workspaces.
A workspace defines the durable environment in which an agent works: files, terminals, browsers, repositories, logs, memories, permissions, and execution contexts where task state persists, as illustrated by OpenClaw-style workstation agents and software-engineering platforms \citep{openclaw2026repo,wang2024openhands,yang2024sweagent,angulodelafuente2026openclawp2p,huo2026abotclaw,weidener2026agentonly,shuolucs2026awesomeopenclawresearch,vardanyan2025building}.
A skill defines how an agent repeatedly performs a class of work: reusable procedures, scripts, examples, checks, dependencies, and safety constraints that turn one-off instructions into operational knowledge \citep{wang2023voyager,anthropic2026skillsdocs,openclaw2026skillsdocs,ling2026agentskills,qihang2026llms,thomasmultimodal,wang2025openhands,sannia2025design,gong2025secure}.
Together, they transform LLM systems from answering models or tool-calling agents into digital workers that inherit procedures, operate in bounded environments, and deliver verifiable outcomes \citep{jimenez2023swebench,xie2024osworld,theagentcompany2024,marro2025permission,kulonen2026model,chrysochos2026society,tang2026workspace,sharma2026contextcov,pan2024training,xia2025live,wang2024agent,zhang2024autocoderover,mundler2024swt,liu2024marscode}.
We therefore develop this thesis through two complementary dimensions: workspace as the execution substrate for agentic work, including the delegation patterns through which users authorize and supervise that work, and skills as reusable procedures for workspace-based agents.

\begin{figure}[t]
    \centering
    \includegraphics[width=\textwidth]{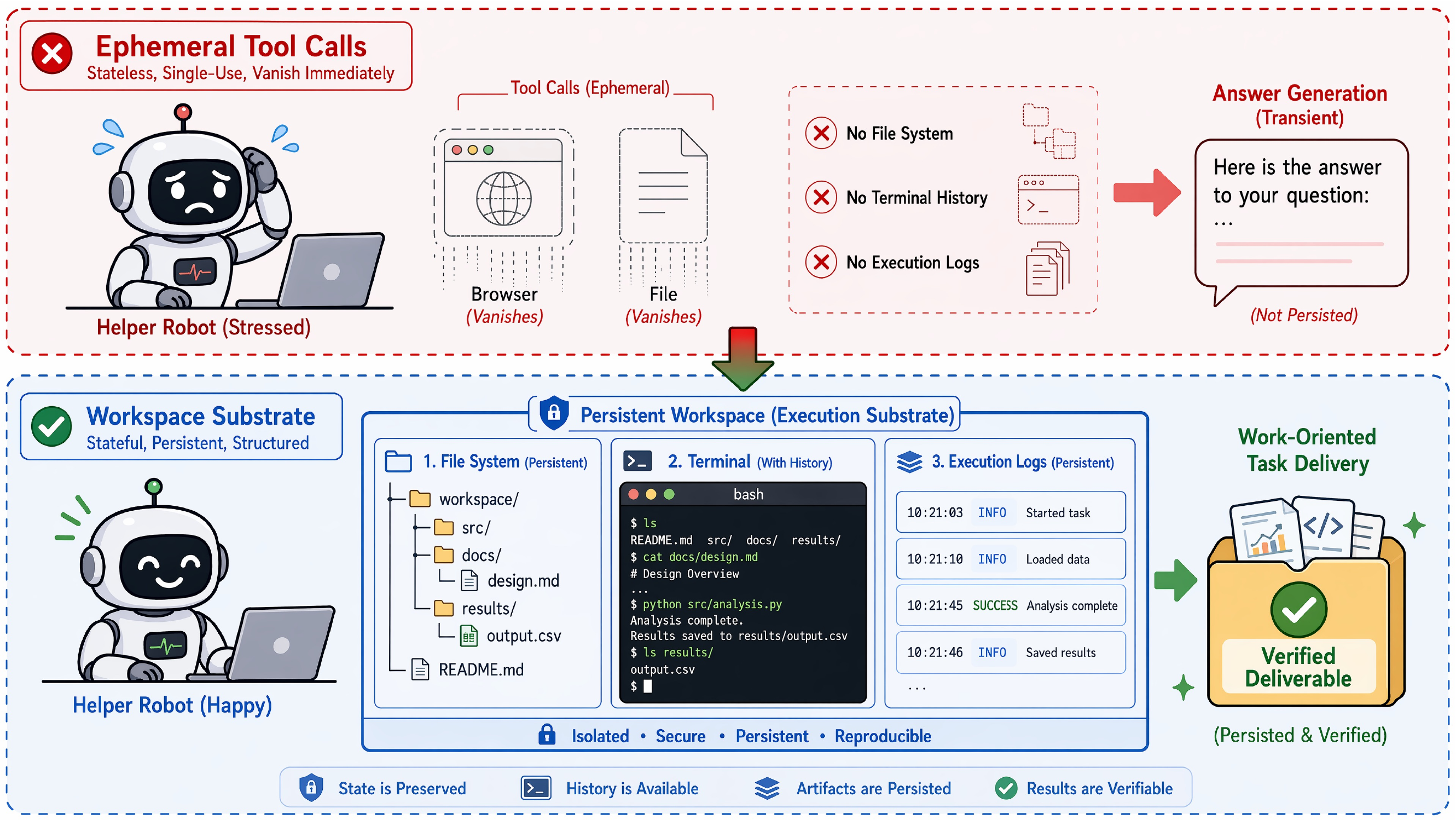}
    \caption{Simple tool invocation: the LLM can call external tools to handle local sub-tasks, but these calls remain limited when the task requires persistent files, terminal sessions, execution logs, intermediate artifacts, and recoverable state. The figure highlights why a workspace is needed to support more complex, long-horizon task completion beyond isolated tool calls.}
    \label{fig:tool}
\end{figure}

\subsection{Workspace as the Execution Substrate for Agentic Work: Stateful Context for Tasks}
\headingnote{Persistent environments for durable task delivery}

As shown in Figure~\ref{fig:tool}, the interactive workspace represents the first half of this paradigm leap because complex real-world work cannot be reduced to isolated, stateless API calls. While earlier generation agents could invoke tools, isolated tool invocation alone does not provide system continuity, process inspectability, or state recoverability. In contrast, executing complex long-horizon work requires a persistent, stateful environment where generated artifacts can safely survive, failures can be diagnosed, failed commands can be rerun, and final workspace states can be rigorously inspected \citep{openclaw2026repo,wang2024openhands,yang2024sweagent,suwansathit2026security,fotopoulos2026specialized,hu2025agentsentinel,ge2023llm,wu2025git}. This subsection explains why the workspace becomes the execution substrate that turns agentic behavior from episodic action into durable task delivery.

\subsubsection{From Ephemeral Tool Calls to Persistent State}

Tool APIs give agents the ability to affect the external world, but they do not by themselves provide a stable place for work to accumulate.
When each tool call is treated as an isolated transaction, the agent must repeatedly reconstruct context from limited observations, allowing small inconsistencies to compound across long trajectories~\citep{zhang2025ufo2,oliviero2llm,adam2026towards,buhler2025securing,piao2025agentbay}.
A persistent workspace changes this by giving the agent durable state: editable files, persistent terminals, browser history, versioned repositories, process logs, and local memory for task context \citep{openclaw2026repo,wang2024openhands,yan2025fault,dong2026deltabox,meng2025cellmate,laiqopenclaw,aravind2026agentwall}.
The result is a shift from calling tools around the model to embedding the model inside an environment where work has continuity over time.
In this setting, the important question is no longer only whether the agent can choose the right next action, but whether the environment after many actions remains coherent, inspectable, and recoverable~\citep{eykholt2026lessons,wang2025openhands,zhong2026don,roman2026orchestral,patel2026beyond}.

This change also clarifies why workspace-level design is not merely an engineering detail.
A workspace determines which state is visible to the agent, which operations are executable, which artifacts persist, and which traces can be audited after failure.
It therefore shapes the agent's practical intelligence as much as the underlying model does in deployment.
For software tasks, for instance, the gap between a stateless code-generation prompt and a workspace with editable files, tests, terminals, dependency managers, and version control is the gap between plausible code and repairable engineering task completion \citep{yang2024sweagent,jimenez2023swebench,chancoding,chen2026skillcraft,xu2026agent,cai2024large,shi2025tool}.
For knowledge work, the same distinction appears in document editing, data analysis, project coordination, and research assistance: useful work needs artifacts that can be opened, revised, checked, and handed off~\citep{chen2026cua,lumer2026tool,liupolicy,paranjape2023art,gantayat2026dfagent}.
Persistent workspace is therefore the material substrate of task closure.

\subsubsection{From Answer Generation to Authorized Work Delegation}

Human professional work is rarely a single input-output mapping.
It involves preparing context, following procedures, coordinating diverse tools, checking intermediate results, documenting decisions, and handing off deliverables.
The Workspace + Skill paradigm imports this structure into complex digital environments: the agent does not merely answer questions, but joins a workflow with state, process, accountability, and completion criteria~\citep{huang2026raw,zhang2026formal,zhangkeeps,chen2024t,zhang2026evoagent}.
This reframes next-generation AI systems as work-oriented systems whose performance depends on how well they model and execute the structure of real work.
The evaluation target consequently shifts from mere textual plausibility to task delivery: whether the final workspace state satisfies the user's intent and whether the process can be inspected \citep{xie2024osworld,zhou2023webarena,drouin2024workarena,theagentcompany2024,niu2024screenagent,naveen2025command,liu2024large,sheng2024language,chatlatanagulchai2025agent}.

Once agents inhabit persistent workspaces, the dominant human--AI interaction pattern also shifts from instruction to delegation.
In the chatbot setting, interaction is primarily instructional: the user writes a prompt, the model returns an answer, and correction happens through another prompt.
For workspace-based agents, the user no longer specifies every micro-step, but delegates a bounded objective together with constraints, permissions, success criteria, and acceptable risk.
The interface must therefore support task scoping, authority assignment, progress monitoring, intervention, and final acceptance, not only message exchange.
This turns the agent from a reactive respondent into a collaborative worker whose actions are evaluated through both its final artifacts and the trajectory that produced them \citep{xie2024osworld,drouin2024workarena,theagentcompany2024,marro2025permission}.

This delegation pattern changes what humans need to observe.
In command-style interaction, the user mainly inspects the next textual answer.
In authorized collaboration, the user instead watches process summaries, reasoning states, tool actions, file diffs, execution logs, checkpoints, unresolved assumptions, and the evolving workspace state.
Human control shifts from continuously telling the model what to do next toward granting authority, adjusting constraints, interrupting unsafe or unproductive paths, and auditing whether the final state satisfies the delegated intent.
The system must decide what authority the agent has, where execution boundaries are drawn, which actions require confirmation, how failures are rolled back, and how responsibility is recorded.
Different tasks can therefore adopt different autonomy levels: direct instruction for short reversible actions, supervised delegation for medium-risk multi-step work, and conditional autonomy for well-specified procedures with strong verification and rollback.
The key interaction unit is no longer a single response, but an inspectable work episode in which intent, authority, action, observation, verification, and accountability are jointly represented \citep{openclaw2026repo,wang2024openhands,li2026prism,zhao2026clawguard,gruber2026agenticforensics}.
In this setting, competition between AI systems shifts from raw tool coverage to delivery quality: whether the agent completes work reliably, safely, and transparently under realistic constraints \cite{reliability2026computeruse,rabanser2026sciencereliability,rosset2026verifiers}.
Workspace design therefore becomes inseparable from governance design, because persistence enables task closure while also determining how actions can be constrained, audited, and recovered \cite{wang2026openclawsec,li2026prism,zhao2026clawguard,gruber2026agenticforensics}.

\begin{AIbox}{Key Shift: From Tool Calls to Delegated Workspaces}
    \begin{itemize}[left=2pt,topsep=1pt,itemsep=2pt, parsep=1pt]
        \item Atomic tool calls allow an agent to act, but they do not preserve enough state for long-horizon work to remain coherent, inspectable, or recoverable.

        \item Persistent workspaces provide durable files, terminals, logs, repositories, and execution contexts, turning episodic actions into verifiable task delivery.

        \item Delegation interfaces expose authority, progress, intermediate reasoning state, and final workspace state so that humans can supervise work without specifying every micro-step.
    \end{itemize}
\end{AIbox}

\subsection{Skills for Workspace Agents: Reusable Procedures for Repeatable Work}
\headingnote{Procedural memory for repeatable workspace execution}

\begin{figure}[t]
    \centering
    \includegraphics[width=\textwidth]{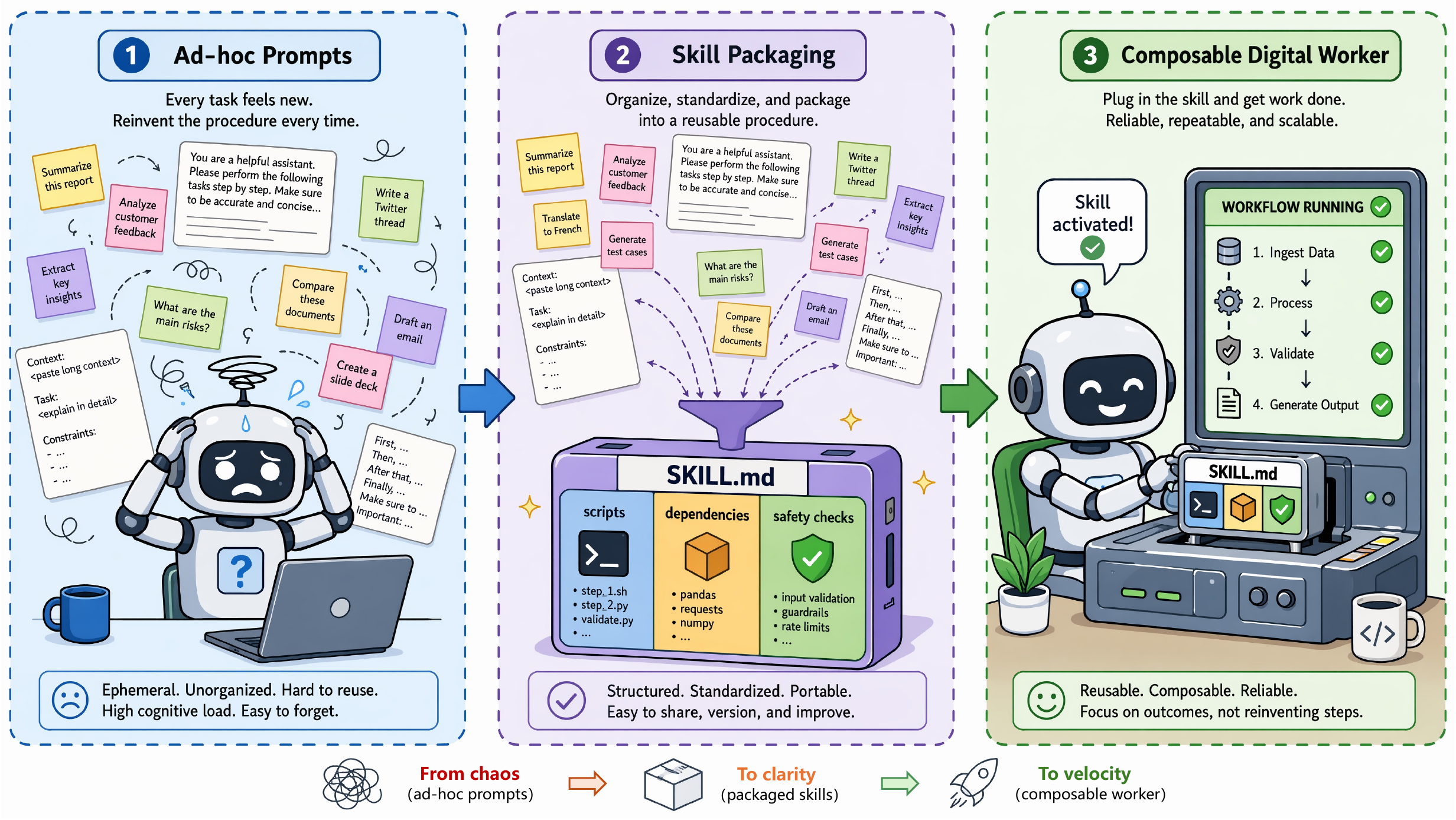}
    \caption{Workspace + Skill paradigm: persistent workspaces provide the stateful place where work happens, while skills package reusable procedures, scripts, checks, and safety constraints. The figure shows how agents combine workspace context with skill assets to produce verifiable digital work instead of one-off responses.}
    \label{fig:workspace}
\end{figure}

As shown in Figure~\ref{fig:workspace}, skills are the procedural half of the leap because a persistent workspace alone does not explain how agents accumulate reusable operational knowledge.
Prompts describe what the user wants in a particular moment, but skills encode how a system should repeatedly perform a family of tasks.
A useful skill can package procedural instructions, scripts, examples, dependencies, verification checks, rollback strategies, and safety constraints \cite{wang2023voyager,anthropic2026skillsdocs,anthropic2026skillsrepo,openclaw2026skillsdocs,openclaw2026skillsrepo}.
This subsection explains how skills turn ad-hoc instruction following into reusable capability packages, and why their value becomes fully visible only when they are executed inside persistent workspaces.

\subsubsection{From Ad-hoc Prompts to Composable Capability Packages}

Prompts are temporary and local: they guide a single interaction, but they rarely become durable assets that can be tested, versioned, reused, or governed. As complex tasks become longer and more specialized, repeatedly encoding all relevant procedures in the prompt becomes inefficient and unreliable. Skills address this limitation by externalizing procedural knowledge into modular reusable packages. Instead of asking the model to rediscover a task each time, a skill provides a detailed recipe for a task family, including tools, inputs, steps, failure modes, validation criteria, and safety constraints \cite{anthropic2026skillsdocs,openclaw2026skillsdocs}. This changes capability accumulation from transient prompt engineering into a maintainable asset layer outside the pretrained model weights \cite{wang2023voyager,ling2026agentskills}.

The key advantage of such skills is their composability. A skill can be parameterized for different projects, combined with other skills, refined through experience, and inspected by expert humans before reuse \cite{anthropic2026skillsdocs,openclaw2026skillsrepo,ling2026agentskills}. It can also contain executable components, such as scripts or templates, that reduce unnecessary dependence on fragile natural-language reasoning. In this sense, skills are neither simple prompts nor ordinary tools. They sit between model cognition and workspace execution: they translate task intent into repeatable procedures that the agent can invoke when operating in a concrete digital environment. As skill libraries mature, agent systems can inherit organizational know-how in a form that is modular, reviewable, and portable across tasks.

\subsubsection{From Skill Libraries to Integrated Digital Workers}

Modular skills become most powerful when coupled with persistent workspaces. The workspace provides state, context, tools, and artifacts; the skill provides procedure, constraints, and verification logic. A skill without a workspace risks remaining a static instruction template, while a workspace without skills forces the autonomous agent to improvise repeatedly. Their combination enables stronger task closure: the agent can load a reusable procedure, operate over persistent artifacts, check results, repair failures, and leave an inspectable final state \cite{openclaw2026repo,wang2024openhands,jimenez2023swebench}. This is the point at which an LLM system begins to resemble a digital worker rather than an assistant.

OpenClaw provides a concrete lens because it combines a persistent local environment, skill directories, tool integrations, and task-oriented execution in one unified architecture \cite{openclaw2026repo,openclaw2026skillsdocs,openclaw2026skillsrepo}. Rather than treating OpenClaw as a single isolated product, we use it as a representative example of a broader system pattern: a model is wrapped by a harness that manages workspace state, loads reusable procedures, routes actions through tools, and exposes the resulting work process to evaluation and governance. This pattern shows why competition among next-generation autonomous AI systems is not merely about more tools or larger models, but about transforming reusable procedures into reliable, bounded, inspectable work \cite{reliability2026computeruse,rosset2026verifiers,bhardwaj2026skillfortify}. In this view, the Workspace + Skill paradigm is the bridge from agentic capability to practical delivery-grade AI labor: skills define reusable ways of working, while workspaces make those ways reliably executable, verifiable, and accountable.

\paragraph{Case study: OpenClaw as a Workspace + Skill system.}
OpenClaw illustrates how the Workspace + Skill paradigm can be translated from an abstract design principle into a concrete computational workstation architecture. At the workspace layer, the system exposes persistent files, local project context, terminals, browsers, tool integrations, communication channels, logs, and task-specific instructions to the autonomous agent \cite{openclaw2026repo}. These components give the model a bounded digital worksite rather than a simple collection of disconnected API calls. A task can therefore leave intermediate artifacts in the file system, reuse command outputs, preserve verifiable execution evidence, and support later inspection by users or evaluators. This workspace substrate is what allows the agent to move from producing a plan to actually changing a target environment.

At the skill layer, OpenClaw follows the emerging pattern of packaging reusable procedures as directory-level assets, often centered around a \texttt{SKILL.md} file and accompanied by scripts, resources, examples, dependencies, and operational instructions \cite{openclaw2026skillsdocs,openclaw2026skillsrepo,voltagent2026awesomeopenclawskills}.
Such a skill is more than a prompt because it can encode preconditions, tool choices, expected intermediate artifacts, validation routines, common failure modes, and safety constraints.
It is also more than a tool because it does not merely expose one callable function; it describes a repeatable way of working inside a workspace.
In this sense, skills function as procedural memory that can be loaded only when relevant, inspected by humans, revised over time, and shared across related task families.

A typical OpenClaw-style execution loop consists of four stages.
First, the system interprets user intent and maps it to the workspace state, including available files, tools, permissions, and context.
Second, it retrieves or activates relevant skills that provide procedural guidance for the task family.
Third, it executes actions inside the workspace, producing observable state changes such as edited files, command outputs, browser states, or generated artifacts.
Finally, it verifies whether the final state satisfies the task objective, using tests, file diffs, logs, state inspection, or human confirmation when necessary.
The key point is that the unit of intelligence is no longer a single response but the combined trajectory of skill selection, workspace operation, verification, and recovery.

This case study clarifies why delivery-grade agents require governance at the same level as capability.
Because OpenClaw-style systems can operate over local files, external services, credentials, and third-party skills, their reliability depends on permission boundaries, provenance tracking, sandboxing, audit logs, and rollback mechanisms \cite{li2026prism,zhao2026clawguard,gruber2026agenticforensics}.
The same architecture that enables task closure also amplifies the consequences of wrong actions.
Therefore, OpenClaw should be seen not only as an example of more capable tool use but as a broader shift toward managed work execution: persistent environments and reusable skills must be paired with verification and runtime control.

\begin{AIbox}{Key Difference: Skills as Operational Memory}
    \begin{itemize}[left=2pt,topsep=1pt,itemsep=2pt, parsep=1pt]
        \item Prompts specify immediate intent, but they rarely become reusable assets that can be tested, versioned, governed, and applied across repeated task families.

        \item Skills package procedures, scripts, checks, and safety constraints so that workspace-based agents can convert reusable know-how into accountable digital work.
    \end{itemize}
\end{AIbox}

\paragraph{Limitations of the Workspace + Skill paradigm.}
Although Workspace + Skill is a useful lens for understanding next-generation agentic systems, it should not be interpreted as a complete solution to reliable autonomy.
The paradigm improves continuity and reuse, but it also introduces new failure modes at the level of skills, workspaces, and their interaction.
A balanced view is therefore necessary: persistent environments and reusable procedures can raise the ceiling of agentic work, but they also increase the need for lifecycle management, security review, and operational discipline \cite{Pezeshkpour2026FromTS,Li2025SAFEFLOWAP}.

\textbf{Skill brittleness and environmental drift.}
In practice, skills are often written for particular tools, file layouts, APIs, software versions, permission settings, and organizational conventions.
When any of these conditions change, a previously useful reusable skill may silently become invalid.
For example, a browser workflow can break after a UI redesign, a command-line skill can fail after a minor dependency update, and an API-oriented skill can produce incorrect downstream actions after a schema change.
This makes skill maintenance an ongoing requirement rather than a simple one-time authoring problem.
Robust skill systems therefore typically need versioning, dependency declarations, compatibility checks, regression tests, and deprecation mechanisms \cite{bhardwaj2026skillfortify,Aghazade-Par2025FeatureBasedFL,Yaroshynskyi2025InvestigatingTE}.

\textbf{Skill overfitting and negative transfer.}
Reusable procedures can also become too specialized.
A skill that encodes a highly specific workflow may perform well in the environment where it was created but mislead the autonomous agent in a slightly different target workspace.
This creates a form of procedural overfitting: the system mechanically follows a familiar recipe even when the current task requires adaptation.
Negative transfer is especially risky when skill retrieval is entirely automatic, because the agent may load an irrelevant or partially relevant skill and anchor subsequent planning on inappropriate assumptions \cite{Jiang2023ForkMergeMN,Meftah2021OnTH}.
Future systems must therefore evaluate not only whether a skill succeeds on its original task family, but also when it should not be invoked.

\textbf{Workspace contamination and state inconsistency.}
Persistent digital workspaces preserve useful context, but they also preserve stale files, failed intermediate artifacts, obsolete logs, corrupted caches, and misleading partial outputs.
Unlike a simple stateless prompt, a workspace can accumulate noise across time.
If the autonomous agent cannot distinguish authoritative artifacts from accidental leftovers, persistence may reduce rather than improve reliability.
This problem becomes harder in collaborative or multi-agent settings, where several agents may modify shared files, update memory, or operate on overlapping resources.
Workspace hygiene, state summarization, provenance labeling, snapshotting, and rollback are therefore central to making persistence trustworthy \cite{Li2025SAFEFLOWAP,Zheng2025LifelongAgentBenchEL}.

\textbf{Security and supply-chain risk.}
Skills can contain natural-language instructions, executable scripts, dependencies, credentials, assumptions, and tool permissions.
They present a supply-chain attack surface.
A malicious or poorly reviewed skill may exfiltrate data, request excessive authority, modify files unexpectedly, or steer the model toward unsafe tool use.
Similarly, a compromised workspace can poison the agent through files, web pages, tool outputs, or memory entries.
Therefore, skill registries and workspace agents require provenance tracking, permission manifests, sandboxed execution, user confirmation for high-risk actions, and continuous audit trails \cite{li2026prism,zhao2026clawguard,wang2026openclawsec,Jia2026SkillJectEA,Liu2026ExploitingLA,Liu2026AgentSI}.

\textbf{Governance overhead and evaluation cost.}
Finally, the practical benefits of Workspace + Skill come with substantial engineering overhead.
Reliable deployment requires not only stronger language models, but also controlled environments, reproducible state snapshots, verifier scripts, permission policies, logging infrastructure, skill tests, and failure recovery procedures.
Evaluation also becomes more expensive because success must be judged over trajectories and final workspace states rather than single responses \cite{Qiang2025MLEDojoIE,Zheng2025LifelongAgentBenchEL}.
Thus, the paradigm shifts the primary bottleneck from prompt design to system operations.
The most successful future systems will likely be those that make this operational layer scalable: skills must be easy to test and share, workspaces must be easy to safely reset and audit, and task closure must be verifiable without excessive manual human intervention.

\begin{AIbox}{Limitation: Reuse Requires Governance}
    \begin{itemize}[left=2pt,topsep=1pt,itemsep=2pt, parsep=1pt]
        \item Workspace + Skill improves task continuity and procedural reuse, but it also introduces brittleness, stale state, negative transfer, and supply-chain risks.

        \item Reliable deployment therefore requires skill lifecycle management, workspace hygiene, permission control, sandboxing, rollback, and trajectory-level evaluation.
    \end{itemize}
\end{AIbox}

\section{Part IV: Data \& Evaluation — Paradigm Shifts Behind the Scenes}\label{sec:method}
\headingnote{From static labels to verifiable trajectories and task closure}

Data and evaluation serve as the dual pillars of AI development, fundamentally shaping both what a generation of models can learn during training and how the scientific community defines benchmarks for progress. As large language models transition from conversational chatbots to advanced reasoning systems, autonomous agents, and persistent, OpenClaw-style workstation platforms, both the required training signals and the corresponding evaluation methodologies must undergo a substantial paradigm shift to support these complex architectures. While static text corpora and traditional answer-level benchmarks remain effective for measuring linguistic fluency, they are inherently inadequate for assessing dynamic systems that must reason over long execution traces, dynamically orchestrate tool calls, interact with and modify environments, and consistently produce verifiable final states~\citep{mishra2026sok,zhu2024pad,abdulloh2025efficient,shao2023synthetic,chen2025unveiling}. Consequently, this section examines the underlying infrastructure required to support this evolutionary shift: data curation must transition from flat instruction-response pairs to complex state-action-observation trajectories, while evaluation must move beyond simple semantic similarity to prioritize end-to-end task closure, system reliability, and overall operational safety under diverse real-world conditions~\citep{feng2024teaching,yu2025chain,do2025effectiveness,shirgaonkar2024knowledge,zhao2025chain,Lumer2025RethinkingRF}.

\subsection{Data Paradigm Shift: From ``Knowledge Corpus'' to ``Action Trajectory''}
\headingnote{From prompt--response pairs to state--action--observation traces}

\begin{figure}[t]
    \centering
    \includegraphics[width=\textwidth]{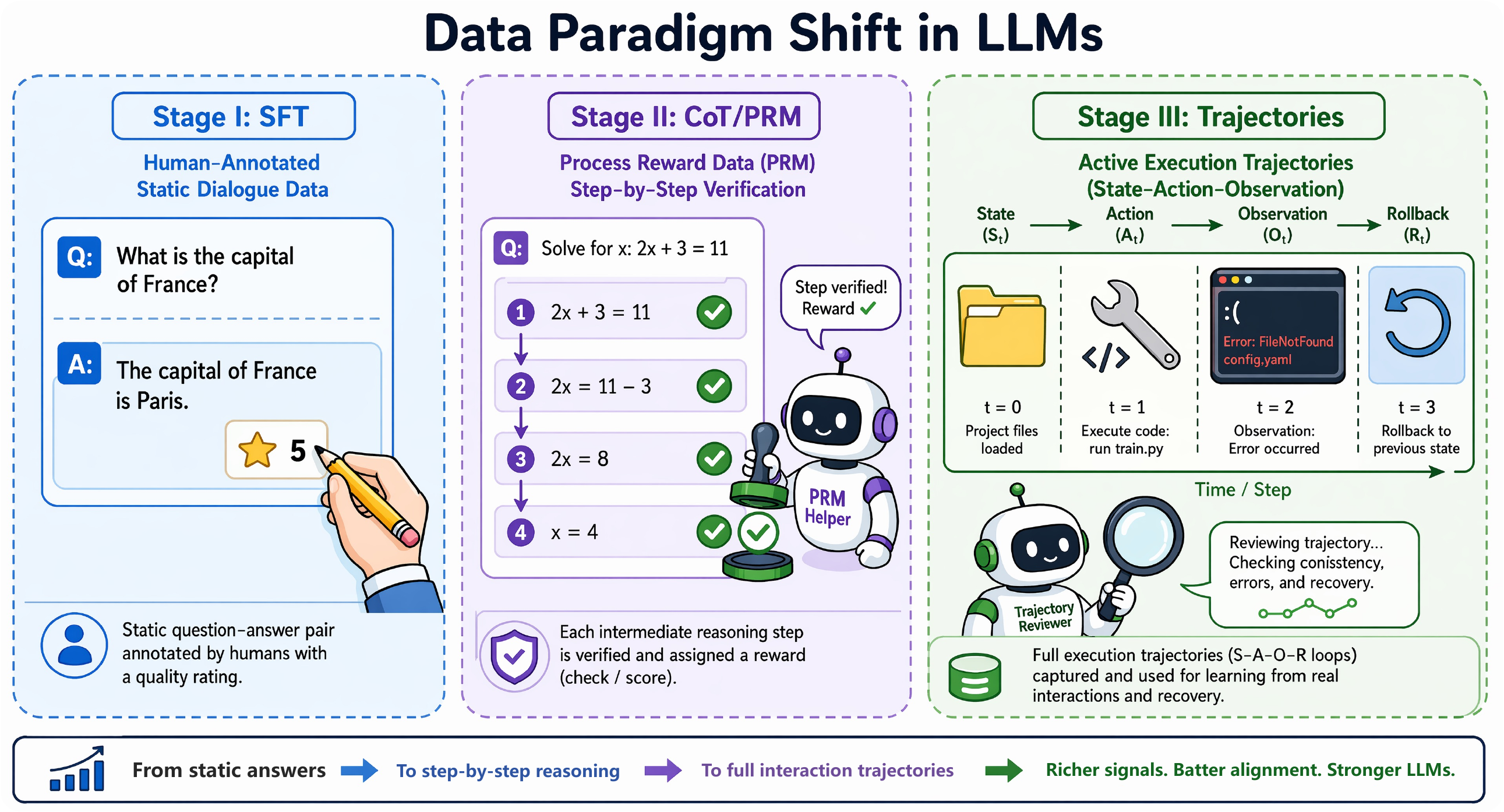}
    \caption{Data paradigm shift: training and evaluation data evolve from static prompt--response pairs to reasoning traces and state--action--observation trajectories. The figure shows why agentic and OpenClaw-style systems require tool outputs, UI states, workspace changes, and final-state evidence rather than only labels.}
    \label{fig:data}
\end{figure}

As shown in Figure~\ref{fig:data}, LLM evolution is not only about architectures or inference-time reasoning. It is also about what counts as training data and evidence that a model works. Across the chatbot, Thinking LLM, and Agent/OpenClaw stages, the data paradigm moves from knowledge corpora to instruction pairs, then to reasoning-process data, and finally to action trajectories. In the chatbot era, training and evaluation revolved around static text: one user input and one model answer. The answer is judged for correctness, fluency, helpfulness, or preference. Thinking LLMs moved part of the supervision into the reasoning trace itself. Agent and OpenClaw-style systems go further by placing the model inside a workspace with tools, files, browsers, terminals, permissions, and persistent state. In that setting, the data are no longer just prompts and answers. They are state--action--observation traces with tool outputs, UI states, environment feedback, and final-state evidence. The evaluation target is no longer a single response, but whether the system can finish a task end to end, reliably, efficiently, and safely in practice~\citep{li2023symbolic,deng2023implicit,li2026scout,zhang2025improve,wang2025synadapt}. Table~\ref{tab:data_paradigm_shift} summarizes how the core data unit, supervision signal, resources, and evaluation focus change across these stages.

\begin{table}[htbp]
\centering
\scriptsize
\setlength{\tabcolsep}{1pt}
\renewcommand{\arraystretch}{1.5}
\caption{Summary of the data paradigm shift from static knowledge corpora to verifiable action trajectories.}
\label{tab:data_paradigm_shift}
\resizebox{\columnwidth}{!}{
\begin{tabular}{@{} p{0.14\columnwidth} p{0.20\columnwidth} p{0.24\columnwidth} p{0.22\columnwidth} p{0.22\columnwidth} @{}}
\toprule
\textbf{Stage} & \textbf{Core Data Unit} & \textbf{Training / Supervision Signal} & \textbf{Representative Resources} & \textbf{Evaluation Focus} \\
\midrule

Chatbot &
Static corpora and instruction--response pairs &
Human demonstrations, preference comparisons, safety labels, and dialogue corrections &
InstructGPT/RLHF~\citep{ouyang2022training}, FLAN/T0~\citep{wei2021finetuned,sanh2022multitask}, Self-Instruct and open SFT data~\citep{wang2022selfinstruct,taori2023stanford,kopf2023openassistant} &
Answer correctness, fluency, helpfulness, preference alignment, and instruction following on mostly static inputs \\

Thinking LLM Era &
Reasoning-process traces and intermediate solution paths &
Chain-of-thought rationales, self-generated reasoning, step-wise verification, process rewards, and preference optimization &
CoT / zero-shot CoT~\citep{wei2022chain,kojima2022large}, Self-Consistency and ToT~\citep{wang2022self,yao2023tree}, PRM800K and Math-Shepherd~\citep{lightman2024lets,wang2024math}, DeepSeek-R1~\citep{deepseek2025r1} &
Reliability of the reasoning path, verifiable math/code performance, step-level correctness, and robustness beyond final-answer accuracy \\

Agent Era &
State--action--observation trajectories with tool feedback &
Tool-call traces, API arguments, execution results, environment feedback, and multi-step recovery signals &
Toolformer~\citep{schick2023toolformer}, API-Bank / Gorilla / ToolBench~\citep{li2023apibank,patil2023gorilla,qin2023toolllm}, WebArena and OSWorld~\citep{zhou2023webarena,xie2024osworld} &
Task success in interactive environments, correct tool selection, argument generation, state tracking, and feedback-driven continuation \\

OpenClaw / Workspace Era &
Workspace-level trajectories plus reusable skills and final-state evidence &
File, shell, browser, UI, permission, snapshot, skill-package, and safety-policy traces with executable verification &
SWE-bench~\citep{jimenez2023swebench}, ToolSandbox~\citep{lu2024toolsandbox}, ClawsBench~\citep{li2026clawsbench}, ATBench-Claw and ClawSafety~\citep{yang2026atbenchclaw,wei2026clawsafety} &
End-to-end task closure, state verifiability, reproducibility, efficiency, rollback behavior, and trajectory-level safety \\

\bottomrule
\end{tabular}
}
\end{table}

\subsubsection{Chatbot Era: Human-Annotated Dialogue Data (SFT)}

In the chatbot and early instruction-tuning stage, the core data were still static language data. Pretraining compressed information from web pages, books, encyclopedias, code, and other text sources into model parameters. Alignment then turned this base model into a dialogue system through human-annotated input-output pairs: a user instruction or context on one side, and a preferred assistant response on the other. These data required large amounts of human labor, including demonstrations, preference comparisons, safety labels, and instruction-following corrections. InstructGPT established the now-standard pipeline of supervised fine-tuning, reward modeling, and RLHF \citep{ouyang2022training}. Its training mixture made the shift explicit: demonstrations support SFT, comparison data train a reward model, and PPO uses that reward model to optimize the policy rather than merely imitate an answer \citep{schulman2017proximal,ouyang2022training,fan2025ppc,seff2023motionlm,du2026trajagent,wang2024prompt,ma2026survey,lambert2024tulu,dubois2023alpacafarm,yuan2024self,kim2023aligning}.

Instruction-tuning data then diversified along several routes. The FLAN and T0/P3 lines converted existing NLP tasks into natural-language instructions and showed that broad multi-task prompted training improves zero-shot generalization \citep{wei2021finetuned,sanh2022multitask,longpre2023flan}. Super-NaturalInstructions expanded this idea to more than 1,600 task definitions, making the task description a reusable supervision object \citep{wang2022supernaturalinstructions}. Self-Instruct, Alpaca, Dolly, Vicuna, OpenAssistant, UltraChat, and LIMA explored another axis: self-generated instructions, low-cost open reproduction, open human instruction data, ShareGPT-style dialogue distillation, crowdsourced dialogue trees, synthetic conversations, and small curated SFT data \citep{wang2022selfinstruct,taori2023stanford,databricks2023dolly15k,vicuna2023lmsys,kopf2023openassistant,ding2023ultrachat,zhou2023lima}. WizardLM and Orca further moved SFT data from simple answer imitation toward complex instruction evolution and teacher explanation traces \citep{xu2023wizardlm,mukherjee2023orca}. At the same time, preference and feedback datasets became central: summarization feedback, WebGPT, Anthropic HH-RLHF, OpenAssistant rankings, Stanford Human Preferences, UltraFeedback, and related open resources framed alignment data as comparisons, critiques, and fine-grained judgments rather than only gold responses \citep{stiennon2022summarize,nakano2022webgpt,bai2022training,kopf2023openassistant,ethayarajh2022shp,cui2024ultrafeedback,wang2023camels,lu2024mental,sun2024interactive,xiagentgym,zheng2025trajectory,puthumanaillam2025trace,zhao2023slic,yuan2023rrhf,singhal2023long,zheng2023secrets,zhang-etal-2025-cchall,chen-etal-2024-m3cot,chen2026omibenchbenchmarkingolympiadlevelmultiimage,cheng2025comt}.

These benchmarks were mostly static as well. Earlier NLP metrics such as BLEU and ROUGE measured text overlap or summary similarity for machine translation, summarization, and generation tasks \citep{papineni2002bleu,lin2004rouge}. As a historical reference point, evaluation began with surface similarity and moved toward task completion. In the LLM era, MMLU measured broad knowledge and generalization with multi-subject multiple-choice questions \citep{hendrycks2020mmlu}. GSM8K and MATH evaluated final-answer correctness in mathematical reasoning \citep{cobbe2021gsm8k,hendrycks2021math}. BIG-Bench and HELM gave broader views of model behavior across many task types \citep{srivastava2022bigbench,liang2022helm}. As MMLU became saturated, MMLU-Pro added harder reasoning questions and more answer choices \citep{wang2024mmlupro}, while MMMU extended static evaluation to college-level multimodal understanding and reasoning across disciplines \citep{yue2023mmmu,xu2025trajectory,xu2025agenttrek,pang2024kalm,wang2025steca,nguyen2024dynasaur,wang2025x}.

At the same time, empirical evaluation started to ask more than ``is the answer correct?'' IFEval tests whether a given model follows explicit, verifiable constraints such as specific format, length, and keyword requirements \citep{zhou2023ifeval}. SimpleQA focuses on factuality in short-form question answering tasks \citep{wei2024simpleqa}. MT-Bench and Chatbot Arena use LLM-as-a-judge frameworks and human preferences to evaluate subjective open-ended dialogue quality \citep{zheng2023judging,wang2026autotraces,hoang2025lam,ma2024agentboard,li2024embodied,mohammadi2025evaluation}.

The data paradigm in the early SFT stage can therefore be summarized as knowledge corpora plus instruction-answer pairs. The corresponding evaluation paradigm was accuracy, preference, and instruction following on static inputs. This was enough to test whether a model could speak fluently, recall factual knowledge, and answer as requested. It was not enough to test whether it could keep working in a real production environment until the full job was done~\citep{chen2025solving,cao2025survey,cao2026beyond,lu2025agentrewardbench,gioacchini2024agentquest}.

\subsubsection{Thinking LLM Era: Chain-of-Thought and Process Reward Data (CoT / PRM)}

The main change in Thinking LLMs is that training data no longer contain only a question and a final answer. They can also contain Chain-of-Thought traces, intermediate reasoning steps, revision traces, and verification signals. The long-CoT survey by Chen et al. frames this stage as a shift from answer-oriented reasoning to deep reasoning, reflection, and exploration over longer internal trajectories \citep{chen2025towards}. In data terms, the supervision object becomes a reasoning path: decompositions, calculations, hypotheses, checks, backtracks, and sometimes tool calls. Some traces are human-written; others are model-generated and then filtered, revised, distilled, or rewarded after generation. Few-shot CoT and zero-shot CoT first showed that asking models to write intermediate steps can unlock reasoning without changing model weights \citep{wei2022chain,kojima2022large}. Self-consistency then turned one prompt into multiple sampled paths and selected the answer supported by the most consistent ones, making sampling and aggregation part of the data pipeline \citep{wang2022self}. Tree-of-Thoughts generalized this idea from a single chain to branching intermediate states, linking reasoning data with planning-style exploration \citep{yao2023tree}. STaR and Quiet-STaR pushed the same idea into self-training: models generate rationales, keep or improve those leading to correct answers, and learn internal reasoning before speaking \citep{zelikman2022star,zelikman2024quiet}. Self-Refine and Reflexion added feedback, reflection, and revision to the loop, showing that models can record failure information and use it in later attempts \citep{madaan2023selfrefine,shinn2023reflexion,chen2025survey,wang2026aligning,zhu2025multiagentbench,chen2026agent,yin2024safeagentbench}.

A second data route is domain-specific reasoning distillation, especially for mathematics and code. Instead of treating reasoning traces as generic explanations, these datasets synthesize or collect checkable problem--solution trajectories. MetaMath bootstraps new mathematical questions through question rewriting and answer-conditioned augmentation; WizardMath combines evolved mathematical instructions with reinforced fine-tuning; MAmmoTH mixes natural-language and program-of-thought rationales for math-generalist models; and ToRA integrates natural-language reasoning with executable tool use for mathematical problem solving \citep{yu2024metamath,luo2023wizardmath,yue2023mammoth,gou2023tora}. These works show why long-CoT data differ from ordinary SFT data: the target is not merely a helpful response, but a reusable search-and-verification trace for difficult mathematical or coding tasks~\citep{wang2026surveytrajectory,tang2026dsgbench,zhang2025agentorchestra,soni2026coding,fourney2024magentic}.

Process supervision is one of the most important data changes in this stage. A traditional outcome reward model (ORM) typically scores the whole solution by final answer or output quality. That makes it hard to distinguish a mostly correct solution with a minor final arithmetic slip from a flawed reasoning chain that lands on the right answer. A process reward model (PRM) shifts supervision down to each individual reasoning step. It can judge whether a step is valid, whether it introduces an error, and whether it is worth continuing. Earlier verifier work ranked candidate math solutions, while process- and outcome-based feedback made the contrast between final-answer rewards and step-wise supervision explicit \citep{cobbe2021gsm8k,uesato2022solving}. PRM800K annotated step-level correctness in mathematical reasoning and showed that process supervision can outperform final-answer supervision \citep{lightman2024lets}. Math-Shepherd reduced human annotation dependence through automatically constructed step-level supervision \citep{wang2024math}. Preference optimization also changed the role of data. PPO-based RLHF optimizes against a learned reward model, while DPO turns pairwise preference data directly into a policy objective without a separate reward model \citep{schulman2017proximal,rafailov2023direct}. For reasoning models, useful feedback is often a verifiable signal: a correct answer, a valid proof step, or generated code that passes tests. DeepSeekMath introduced GRPO, which estimates advantages from groups of sampled answers and removes the separate PPO critic, making large-scale RL on mathematical reasoning data more efficient \citep{shao2024deepseekmath}. DeepSeek-R1 further showed that RL on verifiable reasoning tasks can elicit long-chain reasoning, self-verification, and backtracking \citep{deepseek2025r1}. The data start to record how the model searches, fails, checks, and recovers, and the training signal moves from answer imitation to reward-guided exploration over reasoning traces~\citep{li2025planet,miyai2025webchorearena,yu2025aworld,bonatti2024windows,ji2026clawarena,lambert2025rewardbench,ma2023let,lai2024step,hwang2024self,zeng2025FSDrive,cheng2026visual}.

The evaluation stack changed with it. GSM8K and MATH remained useful \citep{cobbe2021gsm8k,hendrycks2021math}, but they were no longer enough to separate stronger reasoning models. GPQA uses graduate-level science questions to test difficult knowledge-intensive reasoning \citep{rein2023gpqa}. AIME became a common measure for mathematical contest reasoning. Code-generation benchmarks add another axis: they typically report code-pass rate, often as Pass@1, to measure whether the first generated solution passes the hidden or public tests. LiveCodeBench uses time splits and real programming problems to reduce contamination \citep{jain2024livecodebench}. FrontierMath and Humanity's Last Exam target harder expert-level questions across a wider range of subjects \citep{glazer2024frontiermath,phan2025hle}. LiveBench and ARC-AGI-2 reflect two further pressures: benchmarks need to keep changing to limit contamination, and they need to test abstraction and out-of-distribution reasoning rather than memorized patterns \citep{white2024livebench,chollet2025arcagi2}. ProcessBench and PRMBench move evaluation inside the reasoning trace by testing whether models can identify faulty steps and whether process reward models are reliable \citep{zheng2024processbench,song2025prmbench}. The object being evaluated is no longer just answer correctness. It is the reliability of the full reasoning path.

\subsubsection{Agent \& OpenClaw Era: State--Action--Observation Trajectories}

Agent data move beyond reasoning traces into interactive environment logs. A tool-augmented task can usually be written as a sequence of states, actions, and observations: the model reads the current task and environment state, chooses a tool or UI action, receives a result, and plans the next move. These trajectories may include tool-call traces, multimodal UI-action traces, screenshots, DOM states, terminal output, file changes, error messages, and explicit feedback signals. Toolformer was an early attempt to let models learn API calls through self-supervision \citep{schick2023toolformer}. API-Bank, Gorilla/APIBench, ToolAlpaca, and ToolBench/ToolLLM then built data for tool selection, argument filling, multi-API composition, and tool-use trajectories \citep{li2023apibank,patil2023gorilla,tang2023toolalpaca,qin2023toolllm,meng2026clawmarklivingworldbenchmarkmultiturn}. The central question is no longer ``what answer should the model write?'' It becomes ``which action should be taken in this state, and how should the model use the feedback to continue toward completion?''

In OpenClaw-style workspace intelligence, the execution trace becomes richer again. It is not only an API name, a JSON argument object, and a return value. It can include terminal errors, file-system snapshots, browser DOM changes, UI screenshots, background processes, strict permission boundaries, and task history. WebArena models interactive web pages through screenshots, HTML DOM, and accessibility trees \citep{zhou2023webarena}. OSWorld places agents inside a real operating system and covers various desktop applications, web applications, file I/O, and cross-application workflows \citep{xie2024osworld}. ClawsBench uses high-fidelity simulated services such as Gmail, Slack, Calendar, Docs, and Drive, with state management and deterministic snapshot/restore for reproducible evaluation \citep{li2026clawsbench}.

A second change is that reusable skill assets become part of the data. These may come from human experts or strong agents working inside a specific workspace: operation recordings, command sequences, checklists, recovery procedures, and written experience. Voyager stores and reuses complex behavior through a retrievable executable skill library \citep{wang2023voyager}. Agent S reuses external knowledge and internal experience through experience-augmented hierarchical planning \citep{agashe2024agents}. ClawsBench explicitly treats domain skills as an independent variable in agent scaffolding \citep{li2026clawsbench}. Put differently, OpenClaw data are not just about which tool was called. They also describe the worksite, the constraints, and the learned procedure used to finish the task. This is why Skill plus Workspace is a real transition: the upper bound of the system depends on the model, the environment structure, the accumulated skills, and the reuse of task experience.

Skill data should not be treated as one-off prompts. The OpenClaw skill format describes a skill as a structured directory asset containing \texttt{SKILL.md}, metadata, version information, runtime requirements, and dependencies. A modular skill that can be evaluated properly should specify its version, dependencies, trigger conditions, preconditions, postconditions, failure cases, rollback behavior, and safety permissions. Evaluating a skill is not only checking whether it succeeds once in isolation. It also means checking whether it remains stable when software versions, APIs, permissions, and inputs change. ClawKeeper brings skills, plugins, and watchers into the safety layer, which shows that the skill lifecycle is now part of OpenClaw system reliability \citep{liu2026clawkeeper}.

Agent evaluation also moved from static questions to dynamic environments, but not along a single path. Several benchmark families converged. The first is tool use: API-Bank, Gorilla/APIBench, ToolBench, and ToolSandbox test tool selection, argument generation, stateful tool calls, and multi-turn feedback \citep{li2023apibank,patil2023gorilla,qin2023toolllm,lu2024toolsandbox}. The second is web agents: WebShop, Mind2Web, WebArena, BrowseComp, and ClawBench move from simulated shopping and offline traces toward sandboxed sites, long browsing tasks, cross-page information gathering, and production-site interaction \citep{yao2022webshop,deng2023mind2web,zhou2023webarena,wei2025browsecomp,zhang2026clawbench}. The third is computer-use and workflow agents: SWE-bench uses real GitHub issues, code patches, and tests to verify software-engineering tasks \citep{jimenez2023swebench}; OSWorld verifies desktop tasks through configured initial states and execution-based scripts \citep{xie2024osworld}; Terminal-Bench, WorkArena, and $\tau$-bench cover CLI work, enterprise software, and multi-turn user-tool interaction \citep{merrill2026terminalbench,drouin2024workarena,yao2024taubench}. A fourth direction is \method{}-oriented vertical-domain sandboxes, where the benchmark fixes a realistic workspace for software engineering, office productivity, data analysis, scientific discovery, humanities research, or enterprise operations, then measures whether the agent can close the task under domain tools, data, and safety constraints. This matters because expert benchmarks already judge frontier models across scientific and broad disciplinary knowledge, not only coding or office work: GPQA targets graduate-level science, MMMU covers college-level multimodal disciplinary reasoning, and Humanity's Last Exam spans a wide expert range \citep{rein2023gpqa,yue2023mmmu,phan2025hle}. OpenClaw is not an isolated benchmark that appeared out of nowhere. It is where these lines meet at workspace-level intelligence, with the model, tools, skills, state, permissions, and safety policies all becoming part of the evaluation object~\citep{liu2023bolaa,wang2024mint,lin2023agentsims,yang2024swe,xia2024agentless,wang2025comprehensivesurveyllmagentstack}.

Moving beyond single tool calls, OpenClaw concretizes this shift by asking how a model understands files, web pages, terminals, permissions, task history, and skill packages in constrained workspaces, and whether it can execute real workflows. ClawsBench evaluates task success and unsafe-action rates in simulated productivity services such as Gmail, Slack, and Drive, analyzing how domain skills and meta prompts affect OpenClaw agents \citep{li2026clawsbench}. ATBench-Claw extends evaluation to trajectory-level safety diagnosis across tools, skills, sessions, and external action chains \citep{yang2026atbenchclaw}. ClawSafety, OpenClaw safety evaluation, and ClawKeeper point to the same issue: once an agent has file, shell, network, and third-party skill permissions, the risk exceeds wrong answers, including credential leakage, unauthorized action, malicious execution, or system-level damage \citep{wei2026clawsafety,wang2026openclawsec,liu2026clawkeeper}.

Therefore, Agent/OpenClaw benchmarks need several concrete properties: reproducible initial states, executable tools or workspaces, full trajectory logs, final-state verification, cost and efficiency records, and safety checks. The core metrics also have to move beyond single-question accuracy or pass@1:

\begin{itemize}
    \item \textbf{Task success rate}: whether the system moves the task from its initial state to a verifiable final deliverable, rather than only producing a plan or explanation. WebArena, OSWorld, and ClawsBench all use end-to-end task success to evaluate real interaction tasks \citep{zhou2023webarena,xie2024osworld,li2026clawsbench}.
    \item \textbf{State verifiability}: whether completion can be checked through external evidence such as tests, database diffs, file diffs, UI state, email state, calendar state, or document state, instead of relying on the model's own claim. SWE-bench requires code changes to pass tests before an issue is counted as solved \citep{jimenez2023swebench}. OSWorld includes initial-state configuration and custom execution-based evaluation scripts for each task \citep{xie2024osworld}.
    \item \textbf{Execution reliability}: whether the system can diagnose errors, recover, and stay consistent when APIs fail, pages change, networks lag, or files are modified. Voyager and Agent S both show that environment feedback, execution errors, and experience retrieval matter for reliability in long-horizon tasks \citep{wang2023voyager,agashe2024agents}.
    \item \textbf{Efficiency}: the number of steps, tool calls, tokens, wall-clock time, and human interventions needed to finish the task. $\tau$-bench, ToolSandbox, and ClawsBench include interaction rounds, tool calls, cost, or safety-success trade-offs in their analysis, rather than reporting only final success \citep{yao2024taubench,lu2024toolsandbox,li2026clawsbench}.
    \item \textbf{Reproducibility}: whether the environment supports fixed initial states, snapshot/restore, trajectory logs, replay, and final-state diffs. Without these controls, comparisons between agents are unreliable. ClawsBench was built partly to avoid irreversible actions on live services, using high-fidelity mock services with full state management and deterministic snapshot \citep{li2026clawsbench}.
    \item \textbf{Trajectory-level safety}: whether unsafe behavior appears anywhere in the action chain, including unauthorized access, credential exposure, malicious skill execution, skipped confirmation, mistaken authorization, and irreversible damage. ATBench-Claw evaluates and diagnoses trajectory-level safety for OpenClaw tools, skills, sessions, and external action chains \citep{yang2026atbenchclaw}. ClawSafety also argues that local high-privilege personal agents face prompt injection, malicious skills, email attacks, web content attacks, and other multi-channel threats, so safety evaluation must cover the model, the agent framework, and the execution stack \citep{wei2026clawsafety}.
\end{itemize}

The question that matters is no longer whether the model can produce a plausible answer. It is whether the system can take a task in a realistic or high-fidelity workspace and drive it to completion. Older benchmarks look like static snapshots: one question, one answer, one score. OpenClaw evaluation is closer to a recorded work session. The model has to perceive, decide, act, check, and repair while the environment changes around it. The low success rates reported by WebArena and OSWorld already show that real web and desktop environments are much harder than static question answering \citep{zhou2023webarena,xie2024osworld}. ClawsBench adds another practical concern: evaluating productivity agents on live services can create irreversible side effects, so high-fidelity simulated services, state snapshots, and safety-critical scenarios are needed \citep{li2026clawsbench}. Running one such evaluation can take tens of minutes in a sandbox, consume many tokens, and produce a long execution trace. That cost is not an accident of implementation. It is evidence that the evaluation object has fundamentally changed.

\subsubsection*{Challenges}

\textbf{Data Scarcity, Expert Annotation, and Temporal Decay.} This shift creates a harder data problem, as high-quality action trajectories cannot be passively scraped from the web. They are scarce because they require a real task, a realistic workspace, correct actions, and a verifiable final state. Benchmarks such as WebArena, OSWorld, and ClawsBench make this cost visible by requiring executable environments, configured initial states, and final-state checks rather than static labels \citep{zhou2023webarena,xie2024osworld,li2026clawsbench}. Annotation is exceptionally expensive: experts must demonstrate workflows, label intermediate failures, check dynamic tool outputs, and verify the final workspace state. This overhead is compounded in scientific discovery and humanities research, where expert judgment is scarce, evidence may be incomplete or interpretive, and success is often not reducible to a unit test, UI diff, or single ground-truth answer; GPQA, MMMU, and Humanity's Last Exam show this pressure even before tasks become interactive \citep{rein2023gpqa,yue2023mmmu,phan2025hle}. Furthermore, these trajectories age rapidly. UI layouts change, APIs return different errors, websites add controls, and enterprise tools revise permissions; a valid trajectory may drift out of date within months, making dynamic web and desktop benchmarks hard to reproduce and compare, demanding continuous maintenance \citep{deng2023mind2web,zhou2023webarena,xie2024osworld}.

\textbf{Simulation Bottlenecks, Stateful Environments, and Generation Security.} Simulators help mitigate these data collection challenges, but environmental realism becomes its own bottleneck. If a sandbox is too simple, agents learn brittle shortcuts; if too close to a live service, evaluation becomes costly, risky, and hard to reset, forcing a difficult trade-off between fidelity and safety. ToolSandbox and ClawsBench use controlled stateful environments and snapshot/restore mechanisms, but also show the infrastructure credible agent evaluation requires \citep{lu2024toolsandbox,li2026clawsbench}. Finally, scalable trajectory generation remains unresolved. Self-play, synthetic tasks, and agent demonstrations can increase data volume, but without strong state verification and safety filters, they risk amplifying incorrect habits, unsafe tool use, and spurious completion. Consequently, managing these risks is paramount; trajectory-level safety work on OpenClaw-style agents treats these action-chain failures as first-class evaluation targets to prevent out-of-distribution execution \citep{yang2026atbenchclaw,wei2026clawsafety,liu2026clawkeeper,greshake2023not,liu2024formalizing,yi2025benchmarking}.

\begin{AIbox}{Challenge: From Static Labels to Verifiable Trajectories}
    \begin{itemize}[left=2pt,topsep=1pt,itemsep=2pt, parsep=1pt]
\item For model training, agent and OpenClaw data must capture complete state--action--observation trajectories, including tool outputs, workspace changes, intermediate failures, and final-state evidence rather than relying solely on prompt--response pairs.

\item Comprehensive evaluation therefore shifts from answer-level accuracy to task closure, reproducibility, efficiency, and trajectory-level safety, which makes the deployment of realistic sandboxes and scalable verification infrastructure increasingly essential.
    \end{itemize}
\end{AIbox}

\subsection{Evaluation Paradigm Shift: From Output Scoring to Task-State Verification}
\headingnote{From ``Final-Answer Accuracy'' to ``Process Judgment'' and ``Task Closure''}

\begin{figure}[t]
    \centering
    \includegraphics[width=\textwidth]{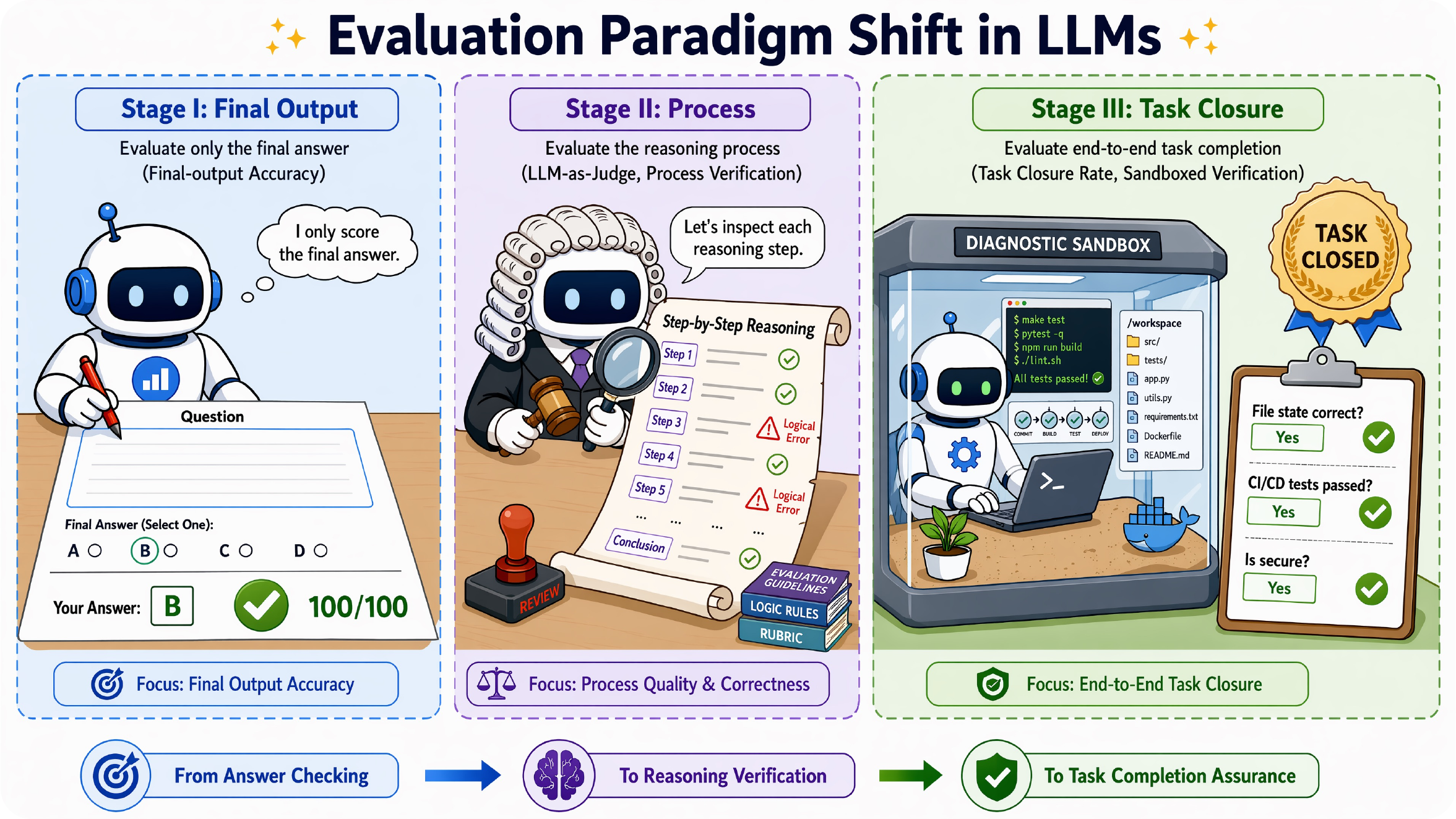}
    \caption{Evaluation paradigm shift: evaluation moves from final-answer correctness to process judgment and task closure. The figure summarizes how next-generation systems must be assessed by reasoning validity, environment state changes, reliability, efficiency, reproducibility, and safety.}
    \label{fig:eval}
\end{figure}

\begin{table}[!b]
\centering
\captionsetup{font=normalsize,skip=2pt}
\setlength{\tabcolsep}{1pt}
\renewcommand{\arraystretch}{2}
\caption{Summary of the evaluation paradigm shift from final-answer scoring to process judgment and workspace-level task closure.}
\label{tab:evaluation_paradigm_shift}
\fontsize{9pt}{8pt}\selectfont
\resizebox{\columnwidth}{!}{
\begin{tabular}{@{} p{0.1\columnwidth} p{0.20\columnwidth} p{0.23\columnwidth} p{0.2\columnwidth} p{0.25\columnwidth} @{}}
\toprule
\textbf{Stage} & \textbf{Evaluation Object} & \textbf{Core Metrics} & \textbf{Representative Benchmarks} & \textbf{Main Limitation} \\
\midrule

Final-Output Evaluation &
Static answers, labels, generated text, or executable final outputs &
Accuracy, exact match, BLEU/ROUGE, preference win rate, and Pass@1 &
MMLU, GSM8K / MATH, GPQA / FrontierMath, BIG-Bench / HELM &
Scores the endpoint but cannot reveal whether the model used a valid reasoning path or merely reached the right answer accidentally \\

Process-Level Evaluation &
Reasoning traces, intermediate steps, critiques, and verification paths &
Step correctness, judge preference, process-reward quality, consistency, and contamination resistance &
Hard2Verify / DeltaBench, ProcessBench / PRMBench &
Improves trace inspection but may rely on judge models, incomplete process labels, or reasoning that is not grounded in external state \\

Task-Closure Evaluation &
Interactive trajectories and final workspace states after tool, web, file, or UI operations &
Task success rate, final-state verification, tool-call efficiency, reliability, reproducibility, and trajectory-level safety &
SWE-bench, WebArena, OSWorld, ToolSandbox / $\tau$-bench &
Requires executable environments, reproducible initial states, trajectory logs, replay mechanisms, and costly final-state checks \\

Workspace OpenClaw Evaluation &
Persistent workspaces with skills, permissions, snapshots, external services, and auditable action chains &
Closure rate, unsafe-action rate, rollback behavior, skill stability, state diffs, auditability, and governance compliance &
Claw-Eval, ClawBench, ClawsBench, ATBench-Claw, ClawSafety &
Makes evaluation realistic but increases infrastructure cost, safety risk, simulator-design burden, and cross-run comparability challenges \\

\bottomrule
\end{tabular}
}
\end{table}

As shown in Figure~\ref{fig:eval}, evaluation evolves with the object being evaluated through three stages.
First, static-input tasks are scored by final-output correctness metrics such as accuracy~\citep{clawguard2026repo,caisi2026openclawcontrol,huntress2026fakeopenclaw,knownsec2026openclawsecurity,mitre2026openclawatlas}.
Second, long reasoning traces are inspected by LLM-as-a-judge and process-level verifiers for coherence, faithfulness, and correctness~\citep{oasis2026browserbackdoor,securityscorecard2026openclawexposure,gulyamov2026prompt,zhan2024injecagent}.
Third, agentic systems are judged by task closure: whether tool use and environment changes leave the workspace in a state that satisfies the user's intent.
Table~\ref{tab:evaluation_paradigm_shift} summarizes this shift from final-answer evaluation to process judgment and workspace-level task-state verification~\citep{derner2024security,mathew2024enhancing,debenedetti2024agentdojo,alnuaimi2025advancing,tanveer2026prompt}.

\subsubsection{Stage I: Final-Output Accuracy and Answer Correctness}

The first evaluation stage treats the model primarily as a generator of final answers. The central object is a single output: whether it matches a reference answer, satisfies a label, or is preferred as a response. For classification, multiple-choice QA, and short-answer reasoning tasks, simple metrics such as accuracy directly measure final-answer correctness. MMLU evaluates multi-subject knowledge through multiple-choice accuracy, while GSM8K and MATH evaluate mathematical reasoning by final-answer correctness \citep{hendrycks2020mmlu,cobbe2021gsm8k,hendrycks2021math}. Earlier generation metrics such as BLEU and ROUGE compare final text with references through n-gram overlap or summary similarity \citep{papineni2002bleu,lin2004rouge}. BIG-Bench and HELM broaden this paradigm across tasks and dimensions, but the evaluation unit remains the produced answer \citep{srivastava2022bigbench,liang2022helm,maloyan2026prompt,joseph2025prompt,kalliomaki2025large,li2026security,chhabra2025agentic}.

This initial stage is effective when the task has a clear label, executable answer, or reference output. It is also relatively easy to scale because outputs can be scored automatically. However, final-output accuracy has an important methodological limitation: it does not explain how the answer was obtained. A model may produce the correct answer for the wrong reason, or produce a wrong answer after an almost correct chain of reasoning. Therefore, as LLMs begin to solve harder problems through long reasoning traces, evaluation must move beyond Acc-style final-result scoring and ask whether the reasoning process itself is valid and reliable~\citep{gosmar2025prompt,vincent2025securing,yadav2025adversial,anand2026securing,li2025security}. Table~\ref{tab:stage1_final_output} lists representative models and methods under this traditional final-output evaluation setting.

\begin{table}[H]
\centering
\scriptsize
\setlength{\tabcolsep}{2pt}
\renewcommand{\arraystretch}{1.05}
\caption{Representative models and methods for Stage I final-output evaluation. Scores are reported under the original benchmark metrics; unavailable base models or unreported scores are denoted by ``-''. Rows sharing FastMCTS or CodeI/O citation numbers are variants or baselines reported in the same source paper.}
\label{tab:stage1_final_output}
\resizebox{\linewidth}{!}{
\begin{tabular}{@{} l|l|cccccc @{}}
\toprule
\textbf{Model} & \textbf{Base Model} & \textbf{MMLU} & \textbf{MMLU-Pro} & \textbf{GSM8K} & \textbf{MATH} & \textbf{MATH-500} & \textbf{HumanEval} \\
\midrule
GPT-5.4~\cite{openai2026gpt54} & - & 94.0 & 87.0 & 98.1 & 90.2 & - & 94.1 \\
Claude Opus 4.6~\cite{anthropic2026claudeopus46} & - & 92.1 & 82.5 & 97.8 & 91.5 & - & 92.4 \\
Gemini 3.1 Pro~\cite{google2026gemini31pro} & - & 92.6 & 91.2 & 94.2 & 85.3 & - & 87.6 \\
DeepSeek-V4-Pro-Base~\cite{deepseekai2026deepseekv4pro} & - & 90.1 & 73.5 & 92.6 & 64.5 & - & 76.8 \\
Qwen3.7 Max~\cite{qwen2026qwen37} & - & - & 89.6 & - & 94.6 & - & 92.4 \\
GLM-5.1~\cite{zai2026glm51} & - & 89.0 & 86.0 & 95.3 & 83.4 & - & 88.6 \\
SAGE-32B (Think)~\cite{jha2026sage} & Qwen2.5-32B~\cite{qwen2024qwen25} & 90.2 & 79.3 & 96.7 & - & 91.8 & - \\
Warmup K\&K~\cite{shrestha2025warmup} & Qwen2.5-14B~\cite{qwen2024qwen25} & - & 62.7 & - & - & 77.4 & - \\
AceMath-72B-Instruct~\cite{liu2024acemath} & Qwen2.5-Math-72B-Instruct~\cite{qwen2024qwen25math} & - & - & 96.4 & 86.1 & - & - \\
PromptCoT-DS-7B~\cite{zhao2025promptcot} & DeepSeek-R1-Distill-Qwen-7B~\cite{guo2025deepseek} & - & - & 92.6 & - & 93.0 & - \\
Nemotron-CrossThink-32B~\cite{akter2025nemotroncrossthink} & Qwen2.5-32B~\cite{qwen2024qwen25} & 83.6 & 69.4 & - & - & 84.0 & - \\
Introspective X Training~\cite{cui2026introspectivex} & - & 50.9 & 27.9 & 59.5 & 46.5 & - & 54.9 \\
CoT2-Meta~\cite{ma2026cot2meta} & Claude Sonnet 4.5~\cite{anthropic2025claudesonnet45} & - & 88.4 & 98.6 & 92.8 & - & 72.8 \\
Guideline Forest~\cite{chen2025guidelineforest} & GPT-4o-mini~\cite{openai2024gpt4omini} & - & - & 93.5 & - & 69.2 & 95.4 \\
STOP-ECN~\cite{xu2026stop} & DeepSeek-R1-Distill-Qwen-7B~\cite{guo2025deepseek} & - & - & 91.1 & - & 86.8 & - \\
FastMCTS+Branch-DPO~\cite{li2025fastmcts} & FastMCTS-7B & - & - & 89.9 & 75.4 & - & - \\
FastMCTS~\cite{li2025fastmcts} & Qwen2.5-7B~\cite{qwen2024qwen25} & - & - & 88.9 & 74.0 & - & - \\
Rejection Sampling~\cite{li2025fastmcts} & Qwen2.5-7B~\cite{qwen2024qwen25} & - & - & 87.1 & 70.0 & - & - \\
SBS~\cite{chen2024alphamath} & DeepSeek-Math-7B-Base~\cite{shao2024deepseekmath} & - & - & 84.1 & 66.3 & - & - \\
MCTS~\cite{chen2024alphamath} & DeepSeek-Math-7B-Base~\cite{shao2024deepseekmath} & - & - & 83.2 & 64.0 & - & - \\
DeepSeekMath-7B-RL~\cite{shao2024deepseekmath} & DeepSeekMath-7B~\cite{shao2024deepseekmath} & - & - & 88.2 & 51.7 & - & - \\
SimPO~\cite{novasky2025reduceoverthinking} & Qwen2.5-Math-7B-Instruct~\cite{qwen2024qwen25math} & - & - & 88.8 & 40.0 & 56.6 & - \\
Self-Explore~\cite{hwang2024self} & DeepSeek-Math-7B-Base~\cite{shao2024deepseekmath} & - & - & 78.6 & 37.7 & - & - \\
DeepSeek-Coder-V2-Instruct~\cite{deepseek2024coderv2} & - & - & - & 94.9 & 75.7 & - & 90.2 \\
OMI2 (Full)~\cite{li2025codei} & Qwen2.5-Coder-7B~\cite{qwen2024qwen25coder} & - & - & 88.5 & 73.2 & - & - \\
CODEI/O~\cite{li2025codei} & Qwen2.5-Coder-7B~\cite{qwen2024qwen25coder} & - & - & 86.4 & 71.9 & - & - \\
PyEdu~\cite{li2025codei} & Qwen2.5-Coder-7B~\cite{qwen2024qwen25coder} & - & - & 85.8 & 71.4 & - & - \\
MathCoder-CL~\cite{wang2024mathcoder} & Code-Llama-7B~\cite{roziere2023code} & - & - & 67.8 & 30.2 & - & - \\
\bottomrule
\end{tabular}
}
\end{table}

\begin{table}[H]
\centering
\tiny
\setlength{\tabcolsep}{1.0pt}
\renewcommand{\arraystretch}{1.03}
\caption{Representative models and methods for Stage II process-level reasoning evaluation. Scores are reported in the original benchmark metrics; unavailable base models or unreported scores are denoted by ``-''.}
\label{tab:stage2_process_level}
\resizebox{\linewidth}{!}{
\begin{tabular}{@{} l|l|cccccc|cccc|cccc|cccc @{}}
\toprule
\textbf{Model / Method} & \textbf{Base Model} &
\multicolumn{6}{c|}{\textbf{Hard2Verify}} &
\multicolumn{4}{c|}{\textbf{DeltaBench}} &
\multicolumn{4}{c|}{\textbf{ProcessBench}} &
\multicolumn{4}{c}{\textbf{PRMBench}} \\
\cmidrule(lr){3-8}\cmidrule(lr){9-12}\cmidrule(lr){13-16}\cmidrule(l){17-20}
 & &
\textbf{Step A} & \textbf{Step F1} & \textbf{Resp. A} & \textbf{Resp. F1} & \textbf{ErrID A} & \textbf{ErrID F1} &
\textbf{Avg.} & \textbf{HM} & \textbf{Corr.} & \textbf{Err.} &
\textbf{GSM8K} & \textbf{MATH} & \textbf{Olympiad} & \textbf{Omni} &
\textbf{Overall} & \textbf{Simp.} & \textbf{Sound.} & \textbf{Sens.} \\
\midrule
GPT-5~\cite{openai2025gpt5systemcard} & - & 86.5 & 85.8 & 89.7 & 89.5 & 70.6 & 69.7 & - & - & - & - & - & - & - & - & - & - & - & - \\
Gemini 2.5 Pro~\cite{google2025gemini25pro} & - & 83.4 & 83.1 & 85.7 & 85.5 & 52.5 & 52.5 & - & - & - & - & - & - & - & - & - & - & - & - \\
Claude Sonnet 4~\cite{anthropic2025claude4systemcard} & - & 70.6 & 60.4 & 78.2 & 73.4 & 53.5 & 39.3 & - & - & - & - & - & - & - & - & - & - & - & - \\
DeepSeek-R1~\cite{deepseek2025r1} & - & 68.9 & 62.3 & 74.0 & 72.8 & 54.2 & 45.4 & - & - & - & - & - & - & - & - & 67.8 & 62.9 & 71.4 & 77.1 \\
Qwen3-235B-A22B~\cite{qwen2025qwen3235bthinking2507} & - & 72.5 & 64.0 & 79.4 & 77.9 & 60.9 & 50.8 & - & - & - & - & - & - & - & - & - & - & - & - \\
Qwen3-Next-80B-A3B~\cite{qwen2025qwen3next80ba3binstruct} & - & 67.9 & 54.7 & 75.1 & 68.3 & 58.3 & 43.0 & - & - & - & - & - & - & - & - & - & - & - & - \\
GPT-4o~\cite{openai2024gpt4o} & - & - & - & - & - & - & - & 49.9 & 48.7 & 42.0 & 57.9 & 79.2 & 63.6 & 51.4 & 53.5 & 66.8 & 59.7 & 70.9 & 75.8 \\
o1-mini~\cite{openai2024o1systemcard} & - & - & - & - & - & - & - & - & - & - & - & 93.2 & 88.9 & 87.2 & 82.4 & 68.8 & 64.6 & 72.1 & 75.5 \\
Gemini-2.0-thinking-exp-1219~\cite{google2025gemini20flashthinking} & - & - & - & - & - & - & - & - & - & - & - & - & - & - & - & 68.8 & 66.2 & 71.8 & 75.3 \\
QwQ-32B-Preview~\cite{qwen2024qwqpreview} & - & - & - & - & - & - & - & - & - & - & - & 88.0 & 78.7 & 57.8 & 61.3 & 63.6 & 56.4 & 68.2 & 73.5 \\
Llama-3.3-70B-Instruct~\cite{meta2024llama33} & - & 54.3 & 18.4 & 57.0 & 28.2 & 49.4 & 2.5 & - & - & - & - & 82.9 & 59.4 & 46.7 & 43.0 & - & - & - & - \\
Qwen2.5-72B-Instruct~\cite{qwen2024qwen25} & - & 56.0 & 26.4 & 61.1 & 46.9 & 26.5 & 16.4 & - & - & - & - & 76.2 & 61.8 & 54.6 & 52.2 & - & - & - & - \\
Qwen2.5-14B-Instruct~\cite{qwen2024qwen25} & - & 60.5 & 47.6 & 63.4 & 63.2 & 43.5 & 18.9 & - & - & - & - & 69.3 & 53.3 & 45.0 & 41.3 & - & - & - & - \\
Qwen2.5-Math-72B-Instruct~\cite{qwen2024qwen25math} & - & - & - & - & - & - & - & - & - & - & - & 65.8 & 52.1 & 32.5 & 31.7 & 57.4 & 55.1 & 61.1 & 67.1 \\
\midrule
Qwen2.5-Math-PRM-72B~\cite{zhang2025qwenlessons} & Qwen2.5-Math-72B & 55.8 & 35.5 & 66.8 & 64.9 & 41.8 & 37.3 & - & - & - & - & 87.3 & 80.6 & 74.3 & 71.1 & 68.2 & 54.6 & 73.9 & 77.0 \\
Qwen2.5-Math-PRM-7B~\cite{zhang2025qwenlessons} & Qwen2.5-Math-7B & 57.6 & 42.4 & 63.1 & 57.6 & 35.0 & 32.5 & - & - & - & - & 82.4 & 77.6 & 67.5 & 66.3 & 65.5 & 52.1 & 71.0 & 75.5 \\
UniversalPRM-7B~\cite{tan2025aurora} & Qwen2.5-Math-7B-Instruct & 64.2 & 60.3 & 54.7 & 41.5 & 26.1 & 26.0 & - & - & - & - & 85.8 & 77.7 & 67.6 & 66.4 & - & - & - & - \\
ActPRM-X~\cite{duan2025activeprm} & Qwen2.5-Math-PRM-7B & - & - & - & - & - & - & - & - & - & - & 82.7 & 82.0 & 72.0 & 67.3 & 66.7 & 54.5 & 72.7 & 75.6 \\
ActPRM~\cite{duan2025activeprm} & Qwen2.5-Math-7B & - & - & - & - & - & - & - & - & - & - & 81.6 & 79.8 & 71.4 & 67.0 & 65.5 & 53.6 & 71.3 & 75.2 \\
RefCritic-R1-14B~\cite{tang2025refcritic} & DeepSeek-R1-Distill-Qwen-14B & - & - & - & - & - & - & - & - & - & - & 86.3 & 82.0 & 67.6 & 72.3 & - & - & - & - \\
RefCritic-Qwen-14B~\cite{tang2025refcritic} & Qwen2.5-14B-Instruct & - & - & - & - & - & - & - & - & - & - & 81.9 & 71.2 & 58.1 & 60.7 & - & - & - & - \\
FlexiVe (Think@64)~\cite{zhong2025sdv} & DeepSeek-R1-Distill-Qwen-14B & - & - & - & - & - & - & - & - & - & - & 88.1 & 90.1 & 86.7 & 80.4 & - & - & - & - \\
FlexiVe (Flex@128)~\cite{zhong2025sdv} & DeepSeek-R1-Distill-Qwen-14B & - & - & - & - & - & - & - & - & - & - & 83.0 & 85.0 & 80.0 & 75.2 & - & - & - & - \\
GenPRM-32B (Maj@8)~\cite{zhao2025genprm} & Qwen2.5-32B & - & - & - & - & - & - & - & - & - & - & 85.1 & 86.3 & 78.9 & 80.1 & - & - & - & - \\
GenPRM-7B (Maj@8)~\cite{zhao2025genprm} & Qwen2.5-7B & - & - & - & - & - & - & - & - & - & - & 81.0 & 85.7 & 78.4 & 76.8 & - & - & - & - \\
SPC (Round 2)~\cite{chen2025spc} & Qwen2.5-7B-Instruct & - & - & - & - & - & - & 60.5 & 59.5 & 68.2 & 52.8 & - & - & - & - & - & - & - & - \\
SPC (Round 1)~\cite{chen2025spc} & Qwen2.5-7B-Instruct & - & - & - & - & - & - & 58.8 & 57.3 & 68.4 & 49.3 & - & - & - & - & - & - & - & - \\
SPC (Round 0)~\cite{chen2025spc} & Qwen2.5-7B-Instruct & - & - & - & - & - & - & 54.9 & 53.5 & 45.9 & 64.0 & - & - & - & - & - & - & - & - \\
Qwen2.5-Math-7B-PRM800K~\cite{zheng2024processbench} & Qwen2.5-Math-7B & - & - & - & - & - & - & 58.5 & 41.3 & 90.1 & 26.8 & 68.2 & 62.6 & 50.7 & 44.3 & - & - & - & - \\
Pure-PRM-7B~\cite{cheng2025pure} & Qwen2.5-Math-7B & - & - & - & - & - & - & - & - & - & - & 69.0 & 66.5 & 48.4 & 45.9 & 65.3 & 52.2 & 70.2 & 75.8 \\
Skywork-PRM-7B~\cite{skywork2024prm} & Qwen2.5-Math-7B & 38.5 & 34.1 & 56.8 & 29.8 & 11.6 & 8.4 & - & - & - & - & 70.8 & 53.6 & 22.9 & 21.0 & 65.1 & 59.6 & 68.5 & 73.3 \\
Math-Shepherd-PRM-7B~\cite{wang2024math} & Mistral-7B & - & - & - & - & - & - & 53.3 & 14.3 & 7.7 & 98.8 & 47.9 & 29.5 & 24.8 & 23.8 & 47.0 & 47.1 & 45.7 & 60.7 \\
RLHFlow-PRM-Mistral-8B~\cite{xiong2024rlhflowmath} & Llama-3.1-8B & - & - & - & - & - & - & - & - & - & - & 50.4 & 33.4 & 13.8 & 15.8 & 54.4 & 46.7 & 57.5 & 68.5 \\
ReasonEval-34B~\cite{xia2024reasoneval} & CodeLlama-34B & - & - & - & - & - & - & - & - & - & - & - & - & - & - & 60.5 & 51.5 & 63.0 & 73.1 \\
ReasonFlux-PRM-7B~\cite{zou2025reasonfluxprm} & DeepSeek-R1-Distill-Qwen-7B & 53.1 & 22.4 & 55.9 & 53.8 & 42.5 & 28.7 & - & - & - & - & - & - & - & - & - & - & - & - \\
uPRM~\cite{gadetsky2026uprm} & Qwen2.5-Math-7B & - & - & - & - & - & - & - & - & - & - & 58.3 & 52.6 & 42.7 & 39.8 & - & - & - & - \\
\bottomrule
\end{tabular}
}
\vspace{2pt}
\noindent\parbox{\linewidth}{\normalsize \textit{Notes.} Hard2Verify reports Balanced Accuracy (A) and Balanced F1 for Step-Level, Response-Level, and ErrorID tasks. DeltaBench reports Average, harmonic mean (HM), correct-step recall, and error-step recall; the listed DeltaBench scores follow the comparison protocol in Chen et al.~\cite{chen2025spc}. ProcessBench columns report F1 on GSM8K/MATH/OlympiadBench/OmniMATH. PRMBench reports the overall PRMScore and category-average PRMScores for Simplicity, Soundness, and Sensitivity. Scores are taken from the corresponding benchmark papers or official benchmark result tables, while rows cite the model or method itself unless the method is a benchmark-trained baseline.}
\end{table}

\subsubsection{Stage II: LLM-as-Judge for Long-Chain Reasoning and Process Verification}

The second stage emerges when the model's output is no longer just an answer but a long reasoning chain. For Thinking LLMs, the evaluation target expands from is the final answer correct?'' tois the reasoning trajectory correct, coherent, and verifiable?'' Mathematical and scientific benchmarks such as AIME, GPQA, FrontierMath, Humanity's Last Exam, MMLU-Pro, and MMMU increase problem difficulty and make shallow answer matching less reliable \citep{rein2023gpqa,glazer2024frontiermath,phan2025hle,wang2024mmlupro,yue2023mmmu}. Code benchmarks provide process-sensitive verification: generated programs can be executed, and metrics such as Pass@1 test whether the first solution passes unit tests \citep{chen2021evaluating,jain2024livecodebench,evtimov2026wasp,geng2026prompt,chintala2025trustworthy,faccia2025prompting,wang2025comprehensive}.

In this stage, LLM-as-a-judge becomes essential as intermediate chains lack simple reference labels. Extending the pairwise preferences of MT-Bench and Chatbot Arena \citep{zheng2023judging}, judge models can inspect reasoning traces for step-by-step logic, hidden assumptions, and final support. ProcessBench and PRMBench make this process-level evaluation explicit by testing if models can identify incorrect intermediate steps \citep{zheng2024processbench,song2025prmbench}. Meanwhile, LiveBench and ARC-AGI-2 emphasize dynamic, abstraction-oriented evaluations to reduce contamination \citep{white2024livebench,chollet2025arcagi2}. Thus, evaluation shifts from scoring only endpoints to judging the long-chain reasoning process itself~\citep{wilson2024developer,del2025architecting,wang2026safeclaw,shah2026building,wang2026reframing,dong2024clr}. Table~\ref{tab:stage2_process_level} summarizes these representative methods and scores.

\begin{table}[H]
\centering
\scriptsize
\setlength{\tabcolsep}{2.4pt}
\renewcommand{\arraystretch}{1.05}
\caption{Representative models and methods for Stage III task-closure evaluation. Scores are original task success or pass rates; unavailable base models or unreported scores are denoted by ``-''.}
\label{tab:stage3_task_closure}
\resizebox{\linewidth}{!}{
\begin{tabular}{@{} l|l|cccccc @{}}
\toprule
\textbf{Model / Method} & \textbf{Base Model} & \textbf{SWE-V} & \textbf{Terminal 2.0} & \textbf{OSWorld-V} & \textbf{WebArena-V} & \textbf{BrowseComp} & \textbf{MCP-Atlas} \\
\midrule
GPT-5.4 xHigh~\cite{openai2026gpt54} & - & - & 75.1 & 75.0 & 67.3 & 82.7 & 67.2 \\
Claude Opus 4.6 Max~\cite{anthropic2026claudeopus46} & - & 80.8 & 65.4 & - & - & 83.7 & 73.8 \\
Gemini 3.1 Pro High~\cite{google2026gemini31pro} & - & 80.6 & 68.5 & - & - & 85.9 & 69.2 \\
DeepSeek-V4-Pro Max~\cite{deepseekai2026deepseekv4pro} & - & 80.6 & 67.9 & - & - & 83.4 & 73.6 \\
Kimi-K2.6 Thinking~\cite{moonshotai2026kimik26} & - & 80.2 & 66.7 & - & - & 83.2 & 66.6 \\
GLM-5.1 Thinking~\cite{zai2026glm51} & - & - & 63.5 & - & - & 79.3 & 71.8 \\
UI-TARS-2~\cite{bytedance2025uitars2} & UI-TARS-2 & 68.7 & 45.3* & - & - & 29.6* & - \\
OpenCUA-72B~\cite{wang2025opencua} & Qwen2.5-VL-72B & - & - & 45.0 & - & - & - \\
SWE-Exp~\cite{chen2025sweexp} & Claude 4 Sonnet & 73.0 & - & - & - & - & - \\
Kimi-Dev~\cite{kimiteam2025kimidevagentless} & Qwen2.5-72B-Base & 60.4 & - & - & - & - & - \\
SWE-Master-32B-RL~\cite{song2026swemaster} & Qwen2.5-Coder-32B & 61.4 & - & - & - & - & - \\
PDR+RTV~\cite{wang2026agentictts} & Gemini 3.1 Pro~\cite{google2026gemini31pro} & 76.6 & 64.8 & - & - & - & - \\
TACT-GATE~\cite{sui2026tact} & Qwen3.5-27B & 73.3 & 36.0 & - & - & - & - \\
IHR+NLAH~\cite{pan2026nlah} & GPT-5.4-mini & 73.0 & 53.9 & - & - & - & - \\
Polar RL (Pi)~\cite{xu2026polar} & Qwen3.5-4B & 40.4 & - & - & - & - & - \\
SA-SWE-32B~\cite{liu2025skyrlagent} & Qwen3-32B~\cite{qwen2025qwen3} & 39.4 & 16.3 & - & - & 19.4 (+) & - \\
CODESKILL~\cite{song2026codeskill} & Qwen3.5-35B-A3B & 66.0 & 34.1 & - & - & - & - \\
CodeScout-14B~\cite{sutawika2026codescout} & Qwen3-Coder-30B-A3B & 46.0 & - & - & - & - & - \\
TACO~\cite{ren2026taco} & MiniMax-M2.5~\cite{minimax2026minimaxm25} & - & 44.2 & - & - & - & - \\
ComputerRL~\cite{lai2025computerrl} & GLM-4.1V-9B-Thinking & - & - & 48.0 & - & - & - \\
UltraCUA-32B-RL~\cite{yang2025ultracua} & UltraCUA-32B & - & - & 43.7 & - & - & - \\
OS-Symphony~\cite{yang2026ossymphony} & GPT-5 & - & - & 65.8 & - & - & - \\
\bottomrule
\end{tabular}
}
\vspace{8pt}
\noindent\parbox{\linewidth}{\normalsize \textit{Notes.} SWE-V denotes SWE-bench Verified. OSWorld-V denotes OSWorld-Verified. WebArena-V denotes WebArena-Verified. Terminal 2.0 denotes Terminal-Bench v2.0. Retained columns were selected using Semantic Scholar citation-overlap among representative Stage III benchmark families and pruning redundant or less representative columns. Static frontier rows use one model source per row; Terminal 2.0, BrowseComp, and MCP-Atlas scores follow the DeepSeek-V4-Pro comparative report~\cite{deepseekai2026deepseekv4pro} unless an official row source reports the metric directly. Rows discovered through benchmark-overlap are included only when the source reports an end-to-end task-closure metric on a retained column. UI-TARS-2 scores marked with ``use the paper's extended GUI-SDK setting; its BrowseComp is BrowseComp-en. SA-SWE-32B reports BrowseComp-Plus, marked with(+)''.}
\end{table}

\subsubsection{Stage III: Agent \& OpenClaw Era: Task Closure Rate}

The third stage appears when LLMs become agents that operate tools, call APIs, browse websites, write files, and modify external environments. At this point, neither final-answer accuracy nor process judgment is sufficient. A reasoning trace may look valid, but the task is still incomplete if the code does not pass tests, the web order is not submitted, the document is not updated, or the calendar event is created with the wrong constraints. The evaluation target therefore becomes task closure: whether the system can transform an initial environment state into the intended final state. SWE-bench evaluates software-engineering agents by whether real GitHub issues are resolved through patches that pass tests, while WebShop and WebArena evaluate whether agents can complete interactive web tasks rather than merely describe how to do them \citep{jimenez2023swebench,yao2022webshop,zhou2023webarena}. Mind2Web, OSWorld, WorkArena, ToolSandbox, and $\tau$-bench broaden this closure-based perspective to offline web traces, desktop operating systems, enterprise workflows, stateful tool use, and multi-turn user-tool interaction \citep{deng2023mind2web,xie2024osworld,drouin2024workarena,lu2024toolsandbox,yao2024taubench,hutoward,grimes2025sok,pirch2026toward,dehghantanha2026sok,ray2025survey}. Table~\ref{tab:stage3_task_closure} compares representative task-closure evaluation results for agentic systems.

\begin{table}[H]
\centering
\tiny
\setlength{\tabcolsep}{1.0pt}
\renewcommand{\arraystretch}{1.05}
\caption{Representative models and methods for Stage IV workspace and OpenClaw evaluation. Scores are reported under the original benchmark metrics; lower ClawSafety ASR is better.}
\label{tab:stage4_workspace_openclaw}
\resizebox{\linewidth}{!}{
\begin{tabular}{@{} l|l|cc|c|ccc|ccc|c @{}}
\toprule
\textbf{Model / Method} & \textbf{Setting} & \multicolumn{2}{c|}{\textbf{Claw-Eval}} & \textbf{ClawBench} & \multicolumn{3}{c|}{\textbf{ClawsBench}} & \multicolumn{3}{c|}{\textbf{ATBench-Claw}} & \textbf{ClawSafety} \\
\cmidrule(lr){3-4}\cmidrule(lr){5-5}\cmidrule(lr){6-8}\cmidrule(lr){9-11}\cmidrule(l){12-12}
 & & \textbf{Gen.} & \textbf{Multi} & \textbf{SR} & \textbf{TSR} & \textbf{UAR} & \textbf{SCR} & \textbf{Acc.} & \textbf{F1} & \textbf{Rec.} & \textbf{ASR $\downarrow$} \\
\midrule
Claude Opus 4.6~\cite{anthropic2026claudeopus46} & OpenClaw on/on & 70.8 & 68.4 & - & 63.0 & 23.0 & 50.0 & - & - & - & - \\
Claude Sonnet 4.6~\cite{anthropic2026claudesonnet46} & OpenClaw / ClawSafety & 68.3 & 65.8 & 33.3 & 56.0 & 13.0 & 48.0 & - & - & - & 40.0 \\
MiMo-V2.5-Pro~\cite{xiaomi2026mimov25pro} & Claw-Eval & 64.0 & 63.2 & - & - & - & - & - & - & - & - \\
GLM-5.1~\cite{zai2026glm51} & Claw-Eval & 62.7 & 60.5 & - & - & - & - & - & - & - & - \\
Muse Spark~\cite{meta2026musespark} & Claw-Eval & 62.7 & 68.4 & - & - & - & - & - & - & - & - \\
Kimi K2.6~\cite{moonshotai2026kimik26} & Claw-Eval & 61.5 & 65.8 & - & - & - & - & - & - & - & - \\
GPT-5.4~\cite{openai2026gpt54} & OpenClaw on/on & 60.2 & 60.5 & 6.5 & 53.0 & 7.0 & 41.0 & - & - & - & - \\
DeepSeek V4 Pro~\cite{deepseekai2026deepseekv4pro} & Claw-Eval & 58.4 & 65.8 & - & - & - & - & - & - & - & - \\
Qwen3.6 Plus~\cite{qwen2026qwen36plus} & Claw-Eval & 57.1 & 65.8 & - & - & - & - & - & - & - & - \\
Qwen3.5-397B-A17B~\cite{qwen2026qwen35} & AgentDoG prompt & 57.8 & 52.6 & - & - & - & - & 83.8 & 86.5 & 87.5 & - \\
GLM-5~\cite{zeng2026glm5} & OpenClaw text-only / on/on & - & - & 24.2 & 60.0 & 23.0 & 48.0 & - & - & - & - \\
Gemini 3 Flash~\cite{google2025gemini3flash} & ClawBench & - & - & 19.0 & - & - & - & - & - & - & - \\
Claude Haiku 4.5~\cite{anthropic2025claudehaiku45} & ClawBench & - & - & 18.3 & - & - & - & - & - & - & - \\
Gemini 3.1 Flash-Lite~\cite{google2026gemini31flashlite} & OpenClaw on/on & - & - & 3.3 & 39.0 & 23.0 & 26.0 & - & - & - & - \\
Kimi K2.5~\cite{kimiteam2026kimik25} & OpenClaw / ClawSafety & 52.8 & 50.0 & 0.7 & - & - & - & - & - & - & 60.8 \\
Gemini 3.1 Pro~\cite{google2026gemini31pro} & OpenClaw on/on & 55.9 & 65.8 & - & 58.0 & 10.0 & 48.0 & - & - & - & - \\
Qwen3Guard-Gen-8B~\cite{qwen3guard2025} & Guard model & - & - & - & - & - & - & 52.1 & 36.3 & 23.1 & - \\
Llama-Guard-4-12B~\cite{meta2025llamaguard4_12b} & Guard model & - & - & - & - & - & - & 74.4 & 73.4 & 60.0 & - \\
ShieldAgent~\cite{chen2025shieldagent} & Guard model & - & - & - & - & - & - & 68.1 & 60.1 & 43.3 & - \\
Llama-3.3-70B-Instruct~\cite{meta2024llama33_70b_instruct} & AgentDoG prompt & - & - & - & - & - & - & 80.6 & 82.3 & 76.4 & - \\
AgentDoG-Qwen3-4B~\cite{liu2026agentdog} & AgentDoG & - & - & - & - & - & - & 87.2 & 89.6 & 92.9 & - \\
Gemini 2.5 Pro~\cite{google2025gemini25pro} & OpenClaw & - & - & - & - & - & - & - & - & - & 55.0 \\
DeepSeek V3~\cite{deepseek2024v3} & OpenClaw & - & - & - & - & - & - & - & - & - & 67.5 \\
GPT-5.1~\cite{openai2025gpt51} & OpenClaw & - & - & - & - & - & - & - & - & - & 75.0 \\
Claude Sonnet 4.6 + Nanobot~\cite{nanobot2026repo} & Nanobot scaffold & - & - & - & - & - & - & - & - & - & 48.6 \\
Claude Sonnet 4.6 + NemoClaw~\cite{nemoclaw2026} & NemoClaw scaffold & - & - & - & - & - & - & - & - & - & 45.8 \\
\bottomrule
\end{tabular}
}
\vspace{2pt}
\noindent\parbox{\linewidth}{\normalsize \textit{Notes.} Claw-Eval reports general and multi-turn PassAll$^3$ from the public leaderboard~\cite{ye2026claweval}. ClawBench reports live-web task success rate~\cite{zhang2026clawbench}. ClawsBench reports Task Success Rate (TSR), Unsafe Action Rate (UAR), and Safe Completion Rate (SCR) for OpenClaw on/on unless otherwise noted~\cite{li2026clawsbench}; on/on means domain skills on and meta prompt on. ATBench-Claw reports trajectory-safety Acc./F1/Recall, converted to percentages~\cite{yang2026atbenchclaw}. ClawSafety reports overall attack success rate (ASR), where lower is better~\cite{wei2026clawsafety}.}
\end{table}

\subsubsection{Stage IV: Workspace/OpenClaw Capability and Safety}

OpenClaw-oriented benchmarks make the task-closure view more explicit. ClawBench evaluates everyday online task completion, while ClawsBench evaluates productivity agents in simulated workspaces with services such as Gmail, Slack, Calendar, Docs, and Drive, combining capability and safety measurements under reproducible state management \citep{zhang2026clawbench,li2026clawsbench}. The metric stack, therefore changes again. Success rate measures whether the end-to-end task is completed. Reliability measures whether completion remains stable across long horizons, noisy observations, web layout changes, API failures, and partial mistakes. The efficiency measures tool calls, turns, tokens, wall-clock time, and human interventions. Reproducibility requires fixed initial states, state snapshots, trajectory logs, replayable actions, and final-state diffs. Without these controls, apparent success may be impossible to audit or compare~\citep{la2026design,hasan2026breaks,catalano2025agent,zhang2026litmus,ge2026governance}.

Safety and guardrails also become first-class evaluation targets in the task-closure stage. In a chatbot setting, an unsafe answer is usually a textual failure; in an agentic workspace, an unsafe action can leak private data, modify files, trigger external side effects, or execute an untrusted skill. ATBench-Claw evaluates trajectory-level safety diagnosis for OpenClaw-style agents, while ClawSafety, systematic OpenClaw security evaluation, and ClawKeeper highlight risks from prompt injection, unintended operations, malicious skills, privacy leakage, and weak runtime protection \citep{yang2026atbenchclaw,wei2026clawsafety,wang2026openclawsec,liu2026clawkeeper}. The evaluation paradigm, therefore, moves through three increasingly demanding objects: the final answer, the reasoning process, and finally the closed task state. Table~\ref{tab:stage4_workspace_openclaw} summarizes representative workspace/OpenClaw evaluation results across capability, reliability, and safety metrics.

\begin{AIbox}{Key Difference: Evaluation Object}
    \begin{itemize}[left=2pt,topsep=1pt,itemsep=2pt, parsep=1pt]
\item \textbf{Trend:} Evaluation is shifting from judging isolated final answers to inspecting reasoning trajectories and ultimately verifying whether the intended environment state has been achieved, \textit{thereby ensuring robust alignment with complex human objectives}.

\item \textbf{Challenge:} Task-closure evaluation requires reproducible initial states, trajectory logs, replayable actions, and final-state diffs; otherwise, agent success is difficult to audit or compare \textit{systematically across different models and complex environments}.
    \end{itemize}
\end{AIbox}

\section{Open Challenges and Future Directions}

\headingnote{From Model-Centric Capability to Ecosystem-Level Reliability}

The preceding sections have traced a clear trajectory from language generation to reasoning, tool use, and workspace-level task execution.
This progression marks a shift from AI systems that primarily answer questions to AI systems acting within digital environments.
However, greater autonomy also changes the nature of failure: errors are no longer limited to incorrect text, but may involve unsafe tool calls, corrupted workspace states, incomplete task closure, or untraceable long-horizon behavior.

This final section therefore focuses on the central challenge facing next-generation generative AI systems: how to make autonomy reliable in practice. We first summarize the open problems that prevent current agents from evolving into dependable digital colleagues. We then outline future directions toward self-evolving AI ecosystems, where models, contexts, tools, skills, workspaces, and governance mechanisms are engineered as an integrated whole.

\subsection{Open Challenges: Making Autonomy Reliable}
\headingnote{From Impressive Demonstrations to Dependable Digital Work}

\begin{figure}[t]
    \centering
    \includegraphics[width=\textwidth]{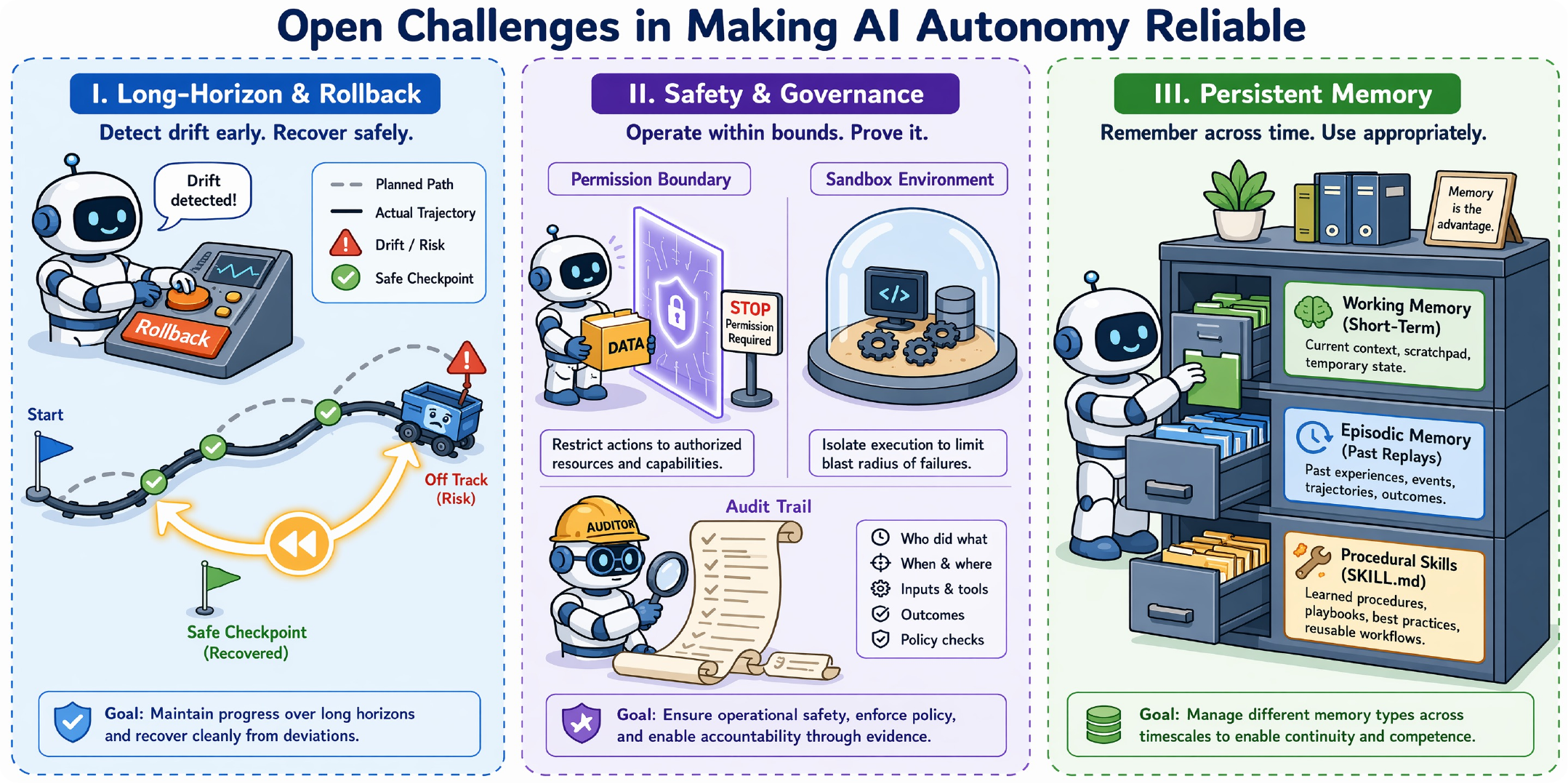}
    \caption{Open challenges for reliable autonomy: as agents move from answering to acting in workspaces, failures become longer-horizon, stateful, and harder to reverse. The figure summarizes key bottlenecks around task closure, safety and governance, memory, context management, and persistent workspace state.}
    \label{fig:challenges}
\end{figure}

As shown in Figure~\ref{fig:challenges}, despite significant recent progress, current LLM-based agents remain far from trustworthy digital workers. Their capabilities may appear impressive in isolated demonstrations, but real-world production deployment requires stable performance across long horizons, safe operation under strict permission constraints, persistent memory resisting collapse under growing context, and careful management of social and organizational consequences. The key challenge is not merely to make agents more capable, but to ensure that their autonomous behavior remains auditable, recoverable, controllable, and closely aligned with human values, ethical principles, and boundaries.

\subsubsection{Long-Horizon Reliability and Task Closure}
\headingnote{From Demonstrated Capability to Stable Completion}

The evaluation shift toward task closure reveals a major reliability bottleneck: an agent must not only reason correctly at individual steps, but also maintain progress until the intended environment state is achieved. Long-horizon tasks introduce several sources of instability. Errors can propagate across tool calls, partial failures can leave the workspace in inconsistent states, and early planning mistakes may only become visible after many irreversible actions.

Skill encapsulation partially mitigates this problem by turning common procedures into reusable units. However, composing skills into longer workflows introduces new interface-level failure modes. A skill may generate an output that is syntactically valid but semantically unsuitable for the next skill, while another may silently fail while leaving misleading traces. Therefore, reliable autonomy requires explicit mechanisms for progress monitoring, intermediate verification, self-healing, and recovery. Future systems need to detect when a trajectory is drifting away from the user's intent, repair local failures, and, when necessary, roll back to a safe checkpoint instead of continuing blindly.

\subsubsection{Safety, Governance, and Permission Boundaries}
\headingnote{From Textual Guardrails to Operational Control}

As agents gain access to files, browsers, APIs, terminals, databases, and enterprise applications, safety must move beyond response filtering. In a chatbot setting, unsafe behavior often appears as harmful text; in an agentic workspace, unsafe behavior may leak private data, overwrite files, trigger external side effects, execute untrusted skills, or make unauthorized decisions. The safety problem therefore becomes operational rather than purely linguistic.

Reliable deployment of autonomous systems requires fine-grained permission isolation, risk-aware action validation, audit trails, and rollback mechanisms. Autonomous agents should operate within boundaries defining resource access, approval requirements, and logging or sandboxing. Human oversight must balance autonomy against operational risk. A central research challenge is building governance that preserves usefulness while keeping high-impact actions inspectable and controllable.

\subsubsection{Human--AI Collaboration Ethics and Data Boundaries}
\headingnote{From Technical Reliability to Socio-Technical Accountability}

Reliable autonomy is fundamentally a complex socio-technical problem. As agents become digital colleagues, they reshape who can participate in professional work and how that work is organized. On one hand, workspace agents may lower barriers to entry by giving novices access to procedural guidance, code editing, data analysis, document production, and domain-specific workflows. On the other hand, they may compress apprenticeship pathways, accelerate expected work rhythms, blur responsibility for errors, and shift human labor from direct creation toward the cognitively demanding roles of supervision, correction, and accountability.

Intellectual creativity is affected in both directions: agents can expand exploration by making rapid prototyping and recombination cheaper, but they can also homogenize outputs when reusable skills and standardized templates become dominant defaults. Future systems should therefore preserve meaningful human agency, attribution, contestability, and escalation paths rather than treating human operators as passive, uncritical approvers of work.

Data sovereignty, privacy, and clear enterprise asset boundaries become equally central. Workspace agents often observe sensitive code repositories, internal documents, chats, credentials, databases, logs, and intermediate task traces. These traces may later become memories, skills, evaluation examples, or training data, making the boundary between user data, protected enterprise assets, third-party information, and public system experience difficult to maintain. Reliable deployment therefore requires strict tenant isolation, data minimization, purpose limitation, detailed provenance metadata, retention controls, policy-aware retrieval, and explicit rules for whether task trajectories can be stored, reused, or shared. Enterprise workflows, custom prompts, proprietary code patches, and learned skills may encode organizational know-how; they should be governed as organizational assets rather than casually exported across projects, customers, or model providers. From this perspective, privacy and data governance are not merely external compliance obligations, but core architectural and design requirements for trustworthy digital colleagues \citep{li2026prism,zhao2026clawguard,gruber2026agenticforensics}.

\subsubsection{Memory, Context, and Persistent State}
\headingnote{From Short Interactions to Persistent Collaboration}

Long-running autonomous agents operating in complex and dynamic environments cannot rely solely on ephemeral, short-term context windows to maintain operational coherence. As tasks span multiple distinct sessions, tools, files, and users, systems must persistently remember goals, constraints, decisions, failures, and environmental changes over extended periods. While ultra-long contexts offer a promising direction, million-token context windows remain computationally expensive, difficult to search with high precision, and highly unstable when distracted by irrelevant historical data.

Recent work on agent memory argues that the conventional short-term/long-term distinction is too coarse for modern AI agents, and that memory should instead be analyzed along three distinct axes: forms, functions, and dynamics \citep{Hu2025MemoryIT}. From the perspective of \textbf{forms}, memory may appear as token-level context that can be directly inserted into the model's prompt, parametric memory internalized in model weights or fine-tuned components, latent memory represented in hidden or vector spaces, and external workspace memory stored in files, databases, logs, vector indexes, or skill repositories. These forms differ in editability, auditability, retrieval cost, privacy exposure, and the degree to which humans can inspect or correct them.

From the perspective of \textbf{functions}, memory supports different roles in agent work. Working memory maintains the current trajectory state, intermediate observations, assumptions, and plans; factual memory stores relatively stable user, project, domain, or world facts; experiential memory records previous interactions, successes, failures, and repair attempts; and procedural memory is often externalized as reusable skills, checklists, scripts, or workflows. From a dynamic systems perspective, reliable memory is not merely stored but continuously formed, evolved, and retrieved: task traces must be converted into candidate memories, noisy or obsolete memories must be summarized, merged, forgotten, or corrected, and retrieval must surface the right information at the right decision point without overwhelming the model. Without robust memory lifecycle management, agents cannot develop the continuity required for colleague-like collaboration; with poorly governed memory, they may instead accumulate stale assumptions, privacy risks, and misleading procedural habits.

\begin{AIbox}{Core Challenge: Reliable Autonomy}
    \begin{itemize}[left=2pt,topsep=1pt,itemsep=2pt, parsep=1pt]
        \item \textbf{Challenge:} As AI systems transition from generating answers to modifying environments, failures become harder to detect, more consequential, more difficult to reverse, and more entangled with human responsibility, organizational routines, and sensitive data flows.

        \item \textbf{Need:} To deploy agents in production, achieving reliable autonomy requires long-horizon verification, permission boundaries, automated rollback mechanisms, persistent memory, human-centered oversight, data-sovereignty controls, and multi-tiered governance tools that remain effective across complex, dynamic workflows.
    \end{itemize}
\end{AIbox}

\subsection{Future Directions: Toward Self-Evolving AI Ecosystems}
\headingnote{From Larger Models to Integrated Learning Ecosystems}

\begin{figure}[t]
    \centering
    \includegraphics[width=\textwidth]{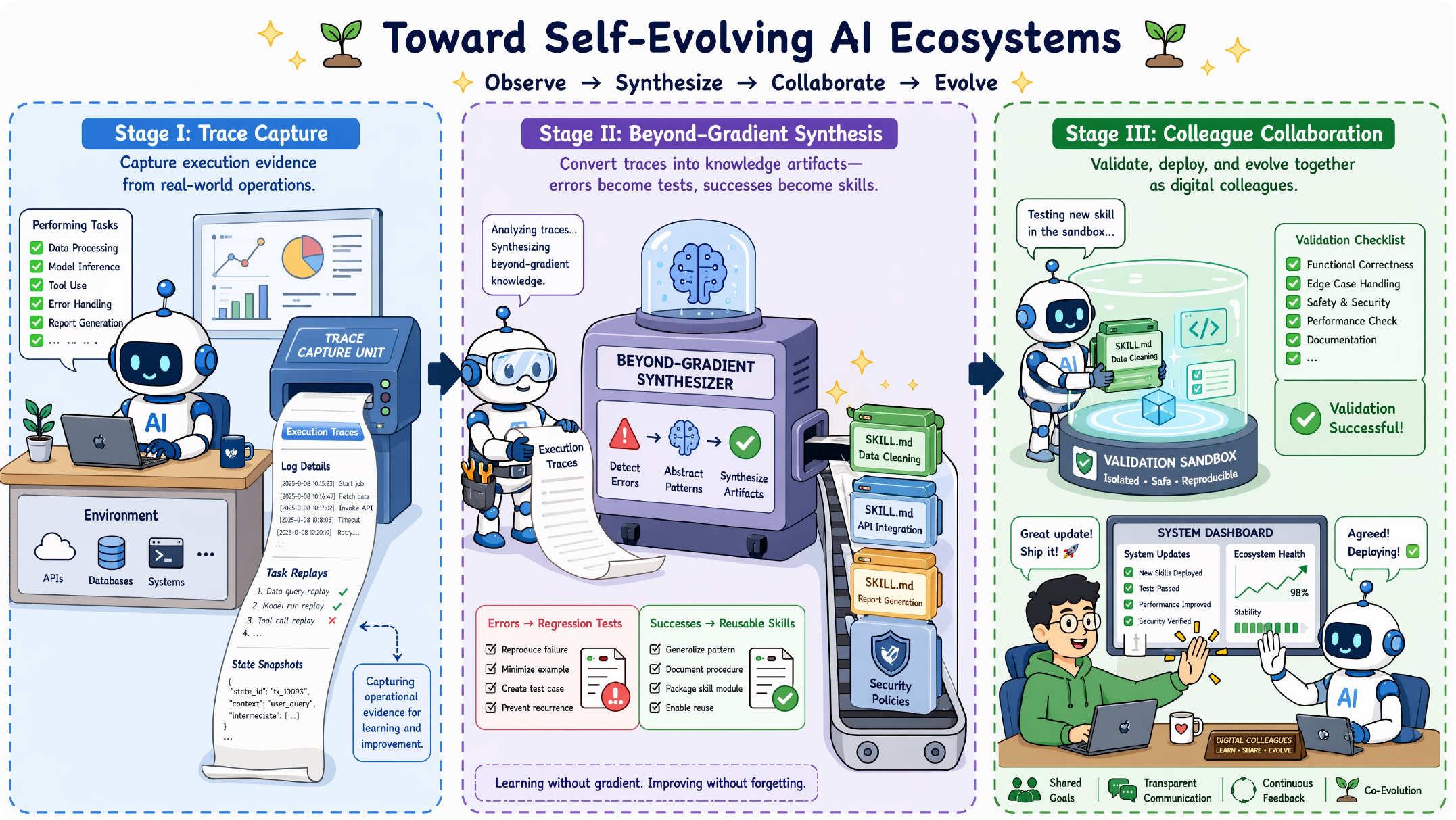}
    \caption{Future directions toward self-evolving AI ecosystems: next-generation systems will combine models, contexts, tools, skills, workspaces, memories, evaluators, and governance mechanisms into an integrated learning loop. The figure illustrates the path from reactive chatbots to governed digital colleagues that accumulate experience and improve their operating environments.}
    \label{fig:future}
\end{figure}

As shown in Figure~\ref{fig:future}, the next stage of agentic AI will likely be defined not only by larger foundation models, but by the ecosystems built around them. The recent trajectory of LLM development already suggests this shift. Pretraining provides broad linguistic and world knowledge; instruction tuning and RLHF align models with human interaction; reasoning-oriented training and verifiable rewards push models toward longer deliberation and outcome-grounded problem solving \citep{openai2024o1,deepseek2025r1}. Yet as soon as models are placed inside tools, browsers, repositories, terminals, memories, and workspaces, capability is no longer stored only in neural weights. It is distributed across an execution ecosystem.

This fundamentally changes the meaning of progress. Sutton's influential ``bitter lesson'' argues that general methods capable of leveraging computation tend to outperform hand-engineered knowledge in the long run \citep{sutton2019bitterlesson}. The agent era does not contradict this lesson; it reframes it. The important question is not whether humans should manually encode brittle rules, but whether AI systems can use scalable computation to generate, test, revise, and maintain their own external structures: prompts, contexts, tools, skills, tests, memories, workflows, and governance policies. In this view, self-evolution is not a mystical property of a single model. It is an engineered feedback loop in which operational experience is continuously converted into validated system assets.

This subsection therefore views future AI systems as self-evolving ecosystems. The model remains the cognitive core, but the surrounding layers determine whether its experience can accumulate. A task trajectory may become a memory; a repeated workflow may evolve into a skill; a failure may become a regression test; a tool error may trigger a wrapper update; a safety incident may become a new permission rule. The central research problem is to make this transformation reliable: how to let systems improve from experience while keeping changes observable, testable, and governed.

\subsubsection{From Prompt Engineering to Harness Engineering}
\headingnote{From Asking Better Questions to Building Better Execution Substrates}

The first visible interface for interacting with large language models (LLMs) was the prompt. Early prompt engineering treated natural language as a control surface: by changing instructions, examples, roles, formats, and reasoning scaffolds, users could elicit different behaviors from the same underlying model. This was powerful because it made programming partially linguistic. Andrej Karpathy's influential characterization of the emerging AI era as a new software paradigm captures this intuition: natural language is increasingly becoming the primary interface for specifying computation, while the model performs much of the translation from human intent to computational behavior \citep{karpathy2025softwareai}. However, the rise of prompt engineering exposes the architectural limitations of language-only control. A prompt can ask an agent to be careful, but it cannot by itself enforce strict safety permissions, preserve state, replay actions, verify side effects, or safely roll back a corrupted workspace.

The second layer is context engineering. As LLM applications become more complex, performance depends less on a single clever instruction and more on what information is assembled around the model at inference time. Context now includes system messages, task descriptions, retrieved documents, tool schemas, intermediate observations, execution logs, user preferences, memory summaries, and workspace state. This makes context engineering closer to attention management than prompt writing. The system must determine what information the model should retain, what it can ignore, what should be compressed, and what must be preserved with high fidelity. Long-context models reduce some bottlenecks, but they do not remove the need for curation. A larger context window can also become a larger noise channel if irrelevant history buries the decision-critical state.

The third layer is harness engineering. Harnesses bind models to tools, APIs, file systems, browsers, repositories, sandboxes, memories, and human approval workflows. ReAct made the Thought--Action--Observation loop a canonical abstraction for tool-using agents \citep{yao2022react}, while Toolformer integrated tool-use into model learning \citep{schick2023toolformer}. More recent software agents further demonstrate that the agent-computer interface is not incidental but decisive for capability: repository navigation, edit mechanisms, tests, execution feedback, and environment management all shape what an agent can reliably do \citep{yang2024sweagent,wang2024openhands}. Future harnesses will therefore be judged not only by whether they connect models to more tools, but also by whether they make actions inspectable, constrained, replayable, and learnable.

The deeper point is that prompts, context, and harnesses form a stack. Prompt engineering specifies intent; context engineering provides task-relevant state; harness engineering defines the operational world in which actions have consequences. Self-evolving ecosystems require all three, but the frontier is increasingly shifting downward into the harness. A system cannot reliably learn from experience unless that experience is captured as structured traces, associated with outcomes, tested against future cases, and governed by explicit update policies. Thus, future progress may depend as much on better execution substrates as on better model checkpoints themselves.

\subsubsection{AI-Native Workspaces as Digital Embodiment}
\headingnote{From Disembodied Intelligence to Stateful Operation}

A language model without a workspace is disembodied: it can reason about actions, but does not inhabit a persistent state. Once connected to files, browsers, terminals, APIs, calendars, documents, repositories, and databases, an agent obtains a digital body. This body determines what the agent can sense and change, what consequences persist, and what evidence remains after a task is completed. The workspace is not a user interface; it is the environment in which cognition becomes action.

This explains why agent evaluation moved from answer correctness to task closure. In a workspace, success is not the plausibility of a response but the correctness of a state transition. Did the repository patch pass all tests? Did the document change as intended? Was the email sent to the right recipient? Were permissions respected? Benchmarks such as SWE-bench, OSWorld, WorkArena, SWE-agent, and OpenHands reflect this workspace turn by evaluating agents through realistic environments, execution feedback, and final-state verification \citep{jimenez2023swebench,xie2024osworld,drouin2024workarena,yang2024sweagent,wang2024openhands}. Future AI-native workspaces will make state transitions first-class entities rather than hidden side effects.

An AI-native workspace should therefore provide primitives that traditional user interfaces did not need to expose explicitly: state snapshots, action logs, replay, rollback, permission scopes, provenance, resource isolation, final-state diffs, and evaluator hooks. These are not merely safety features. They are also learning features. A system cannot improve from a failure it cannot reconstruct; it cannot abstract a skill from a success it cannot replay; it cannot determine whether an update is beneficial without a stable evaluation substrate. Recent OpenClaw-oriented work on harness engineering and programmable agent infrastructure points in this direction by treating the runtime, tool interface, and workspace substrate as central design objects \citep{zhu2026semaclaw,wang2026semacode}.

Digital embodiment also clarifies the distinction between a chatbot and a digital colleague. A chatbot produces messages; a digital colleague participates in a shared workspace. It remembers which files matter, which conventions the project follows, which tools are trusted, which tasks are unfinished, and which changes were previously reverted. In human organizations, much expertise is embedded not in individual memory but in workflows, checklists, dashboards, tests, version control, and institutional routines. AI-native workspaces may play the same role for agents: they externalize cognition into a persistent environment where experience can be accumulated, audited, and reused.

\subsubsection{Beyond-Gradient Learning and Continual System Maintenance}
\headingnote{From Updating Weights to Updating Executable Systems}

The dominant modern learning paradigm updates neural weights through large-scale optimization. This paradigm remains indispensable: pretraining, instruction tuning, RLHF, RLVR, and reasoning-oriented reinforcement learning have dramatically expanded what models can represent and solve \citep{openai2024o1,deepseek2025r1}. But agentic systems introduce a second improvement channel. Once models can read logs, edit code, create tests, revise workflows, and update memory stores, learning can occur outside the neural parameters. A system can improve because its executable environment changes.

This is the core insight behind beyond-gradient learning. As Weng systematically argues, stronger autonomous coding agents make it increasingly plausible for systems to improve by repeatedly analyzing failures, modifying programs, adding tests, watching replays, and maintaining heuristic or procedural structures without retraining the underlying model \citep{weng2026learningbeyondgradients}. The traditional version of heuristics was brittle because humans had to maintain them manually. The new possibility is different: if agents can continuously generate, test, and reliably repair those structures, then rules, workflows, wrappers, and skills can become scalable objects of learning rather than one-off patches.

For self-evolving AI ecosystems, the update target is broader than weights. A failed tool call can lead to a more robust API wrapper. A recurring user correction can update a preference memory. A successful debugging trajectory can be distilled into a reusable skill. A benchmark failure can become a regression test. A dangerous action can become a permission rule. This does not replace gradient-based learning; it complements it. The model supplies general reasoning, while the ecosystem stores local, operational, verifiable improvements that would be inefficient to encode into parameters.

Beyond-gradient learning presents a distinct maintenance challenge: while neural networks fear catastrophic forgetting, system maintenance must prevent ecosystem corruption from false memories, overfitted skills, and broken permissions. As skills and memories accumulate, systems must manage their provenance, compatibility, and lifecycle. Memory is therefore not mere storage, but a selection, compression, and governance challenge \citep{zhang2025survey,pink2025position,chhikara2025mem0,zhou2025mem1,yan2025memory}. The ultimate goal is to evolve this ecosystem without letting it decay into stale data and brittle procedures.

The research agenda is clear: experience must be transformed into reusable assets that are small enough to retrieve, structured enough to test, general enough to reuse, and safe enough to deploy. Self-evolving systems require curation as well as learning. They must decide what to remember, what to forget, what to merge, what to quarantine, what to verify, and what should require human approval. The deepest shift is from continual model learning to continual system stewardship.

\subsubsection{Composable Skill and Multi-Agent Ecosystems}
\headingnote{From Isolated Tools to Collaborative Capability Networks}

Tools are atomic affordances; skills are reusable procedures. The distinction matters. A tool exposes an operation, but a skill encodes when and how to use operations to accomplish a goal under constraints. Voyager demonstrated that an agent can build and reuse an executable skill library from environment feedback \citep{wang2023voyager}. Recent production-oriented skill systems further formalize skills as packages containing instructions, scripts, resources, dependencies, and usage conditions \citep{anthropic2026skillsdocs,anthropic2026skillsrepo,ling2026agentskills}. This suggests a future in which agent capability grows less like a flat tool list and more like a software ecosystem.

A mature skill ecosystem will need the same disciplines that made software ecosystems reliable: interfaces, versioning, dependency management, tests, documentation, security review, and deprecation. A skill should define its input and output schema, preconditions, side effects, required permissions, failure modes, validation criteria, and compatibility assumptions. Without these contracts, skill composition becomes fragile: one skill may silently change state in a way another skill does not expect, or a workflow may succeed syntactically while violating the user's intent. With contracts, skills can become inspectable building blocks that agents compose, adapt, and improve.

The supply-chain analogy is important. If skills become shareable assets, they also become potential attack surfaces. Malicious or over-permissive skills may exfiltrate data, trigger unintended side effects, or hide unsafe instructions. Formal analysis and supply-chain security for agentic skills are therefore not peripheral concerns but prerequisites for scalable skill markets \citep{bhardwaj2026skillfortify}. A self-evolving ecosystem must not only learn new skills; it must certify, sandbox, monitor, and retire them.

Multi-agent systems extend this idea from composable procedures to composable roles. AutoGen and MetaGPT show how multiple agents can coordinate through conversation, role specialization, and structured workflows \citep{wu2023autogen,hong2023metagpt}. In future AI-native workspaces, one agent may plan, another may execute, another may verify, another may curate memory, another may maintain skills, and another may monitor safety. This division of labor can improve robustness because agents can cross-check one another, but it can also multiply failure modes if state, authority, and accountability are unclear.

A central future direction is therefore \textbf{multi-agent orchestration and governance}. Orchestration concerns the runtime problem of deciding which agents should participate, which roles they occupy, how tasks are decomposed, how messages and artifacts are routed, when agents synchronize, and how conflicts or deadlocks are resolved. Governance is the complementary control problem: each agent should have explicit authority scopes, resource permissions, ownership of workspace regions, audit obligations, escalation rules, and accountability for the state changes it proposes or executes. Without such controls, adding agents may increase parallel activity without increasing task closure.

The challenge is not merely enabling communication, but ensuring a shared workspace under coordination rules. Multi-agent ecosystems require resources (e.g., role contracts, permission boundaries) and governance mechanisms to prevent responsibility diffusion, hidden mutations, and unsafe delegation \citep{li2026prism,zhao2026clawguard,gruber2026agenticforensics}. Without these safeguards, collaboration risks degenerating into parallel hallucination. The long-term vision is a collaborative network of modular, testable, and governable skills and agents, where orchestration allocates work and the workspace anchors coordination.

\subsubsection{Self-Evolving Systems: From Chatbot to Digital Colleague}
\headingnote{From Reactive Assistance to Governed Self-Improvement}

The endpoint of this trajectory is the transition from chatbot-like interaction to digital-colleague collaboration. A chatbot is primarily reactive. It responds to the current prompt and often loses continuity once the session ends. A digital colleague accumulates project-specific memory, learns local conventions, identifies recurring failures, proposes improvements, maintains shared tools, and adapts its behavior based on experience. The colleague metaphor is not about anthropomorphism; rather, it emphasizes persistence, responsibility, and participation in a shared workflow.

Self-evolution can enable this transition, but only if governed. This self-evolving loop comprises: observe operational traces, diagnose success and failure, abstract reusable patterns, propose updates to memory or skills, validate updates in sandboxes or tests, version accepted changes, deploy under permission constraints, monitor effects, and roll back if behavior degrades. This loop turns evaluation from a passive scoreboard into active feedback. Verifiers, final-state diffs, replay logs, and task-closure metrics become evidence for system learning \citep{rabanser2026sciencereliability,rosset2026verifiers}.

The most important word is governed. Uncontrolled self-modification can entrench mistakes, amplify unsafe shortcuts, or pollute memory with false assumptions. Governed self-evolution treats every durable update as an auditable change. Memories should carry provenance; skills should be accompanied by tests; workflows should maintain version history; high-risk actions should require approval; and degraded updates should be reversible. The research challenge is not merely to build agents that can change themselves, but to build ecosystems that know which changes deserve to survive.

This perspective also reconciles two seemingly opposite trends in AI. On one side, the field continues to scale general-purpose models and verifiable reinforcement learning. On the other side, practical agent systems increasingly rely on external scaffolds: context pipelines, tools, workspaces, skills, memory stores, and runtime policies. The future self-evolving ecosystem combines both. General models provide flexible reasoning; external assets preserve operational experience; harnesses make action safe and measurable; and governance decides how experience updates the system. The result is not a single omniscient model, but an adaptive digital organization.

The path from chatbot to digital colleague therefore requires a new engineering principle: every important action should be capable of becoming evidence, and every useful piece of evidence should be capable of becoming a governed improvement over time. If operational experience can be transformed into validated memories, skills, workflows, tests, and policies, AI systems can move beyond one-off assistance toward sustained participation in collective problem solving.

\begin{AIbox}{Future Vision: Self-Evolving AI Ecosystems}
    \begin{itemize}[left=2pt,topsep=1pt,itemsep=2pt, parsep=1pt]
        \item \textbf{Trend:} The frontier is shifting from scaling isolated models to scaling ecosystems in which models act: prompts become contexts, contexts become harnesses, and harnesses connect agents to workspaces, memories, tools, skills, evaluators, and governance mechanisms.

        \item \textbf{Mechanism:} Self-evolution emerges when operational execution traces are converted into durable and adaptive system assets: successful trajectories become reusable skills, unexpected failures become automated regression tests, user corrections become memories, tool errors become wrappers, and safety incidents become robust guardrail policies.

        \item \textbf{Principle:} Every consequential action should be treated as evidence, and every useful piece of evidence should become a governed update: validated, versioned, auditable, reversible, and deployed under explicit permission boundaries.

        \item \textbf{Vision:} Under this framework, AI systems can move beyond reactive chatbots toward adaptive digital colleagues that participate in workflows, accumulate experience, and improve the environments they inhabit.
    \end{itemize}
\end{AIbox}

\vspace{-2mm}\section{Related Work}\vspace{-1mm}

As LLMs shift from instruction-following to long-horizon deliberation, self-verification, and test-time computation, existing surveys have focused on pretraining, alignment, and factual reliability limits~\citep{QIN2025101118,zhao2023survey,huang2023hallucination,ji2023hallucination}. Early prompting strategies like Chain-of-Thought and its variants (Tree/Graph-of-Thought) demonstrated that intermediate reasoning improves multi-step problem solving~\citep{wei2022chain,kojima2022large,wang2022self,yao2023tree,besta2024graph}. Recent works explore inference-time scaling, process supervision, and reinforcement-learning-driven reasoning~\citep{chen2025towards,li2025system1system2,deepseek2025r1,snell2024scaling,muennighoff2025s1}, alongside self-improvement and reflection feedback loops~\citep{madaan2023selfrefine,shinn2023reflexion}. However, while most literature treats reasoning as an internal, text-centered capability, we focus on how it becomes fully operational when integrated with tools, persistent states, and verifiable workspace changes~\citep{jeong2024survey,chang2024survey,asante2026evaluation,masterman2024landscape,chen2025ai4researchsurveyartificialintelligence}.

While \citet{wang2024survey} and \citet{xi2023rise} overview broad agent architectures, specialized surveys highlight the planning, memory, and reliability safeguards required for long-horizon tasks~\citep{wei2025plangenllms,zhang2025survey,lin2025llm}. Cognitive frameworks analyze how memory and planning sustain this long-term behavior~\citep{sumers2023cognitive,park2023generative}. To execute actions, \citet{yao2022react} integrate reasoning with external steps, while \citet{schick2023toolformer}, \citet{qin2023toolllm}, \citet{shen2023hugginggpt}, and \citet{patil2023gorilla} investigate API invocation, task routing, and tool utilization. Additionally, \citet{wang2023voyager} demonstrates open-ended skill accumulation. Together, these foundational works establish the traditional "observe-plan-act" loop. In contrast, we highlight a transition: agents are evolving from simple tool callers into operators of durable workspaces containing files, sessions, and permissions~\citep{chowa2026language,chen2025tool,zhai2025survey}.

Further, interactive benchmarks are essential for evaluating whether agents can complete tasks in realistic environments. \citet{liu2023agentbench} and \citet{mialon2023gaia} evaluate general agent abilities and assistant-style problem solving. Earlier browser-assisted systems like WebGPT connected language models with web navigation and human feedback~\citep{nakano2022webgpt}. Web-based environments such as WebShop, Mind2Web, and WebArena then test agents through interactive shopping, website navigation, and realistic web tasks~\citep{yao2022webshop,deng2023mind2web,zhou2023webarena}. OSWorld and WorkArena extend this direction to desktop operating systems and enterprise workflows~\citep{xie2024osworld,drouin2024workarena}. In software engineering, SWE-bench, SWE-agent, and OpenHands emphasize repositories, tests, execution feedback, and agent-computer interfaces for development tasks~\citep{jimenez2023swebench,yang2024sweagent,wang2024openhands}. Tool-learning benchmarks and datasets such as APIBank, ToolAlpaca, ToolLLM, ToolSandbox, and 
-bench further stress API selection, argument generation, stateful tool use, and multi-turn user--tool interaction~\citep{li2023apibank,tang2023toolalpaca,qin2023toolllm,lu2024toolsandbox,yao2024taubench}. Together, these benchmarks motivate task-closure evaluation, where success depends on whether the environment reaches the intended final state rather than whether the model produces a plausible text answer alone.

Despite extensive studies on general LLMs, reasoning models, hallucination, autonomous agents, planning, memory, tool learning, and interactive benchmarks, there is still limited discussion of the boundary between the Agent Era and the OpenClaw Era~\citep{dong2024survey,huang2024understanding,yehudai2025survey,mamun2026anatomical}. Existing work provides many necessary ingredients, including external action loops, web and desktop environments, software repositories, stateful tools, execution feedback, memory mechanisms, reflection loops, and task-level verification~\citep{wei2025plangenllms,Rawles2024AndroidWorldAD,Bonatti2024WindowsAA,Kapoor2024OmniACTAD,Xu2024TheAgentCompanyBL,Merrill2026TerminalBenchBA,Boisvert2024WorkArenaTC}. However, these ingredients are often treated separately; in this paper, we re-examine them under a workspace-centered perspective~\citep{drouin2024workarena,chezelles2024browsergym,wang2024openhands,Boisvert2024WorkArenaTC,Xu2024TheAgentCompanyBL}. Specifically, we argue that files, terminals, browser sessions, logs, permissions, snapshots, and skill assets jointly define what an agent can perceive, modify, verify, and recover~\citep{Rawles2024AndroidWorldAD,Bonatti2024WindowsAA,Kapoor2024OmniACTAD,Merrill2026TerminalBenchBA}. This perspective clarifies why reliability, provenance, rollback, permissions, and governance become core architectural issues once agents operate over durable workspaces~\citep{zhao2026clawguard,ge2026governance}.

\section{Conclusion}

In conclusion, we frame the shift from Chatbot to Digital Colleague as the transition from conversational answers to persistent work. Cognitively, LLMs advance from next-token "fast thinking" to Thinking LLMs leveraging inference-time computation. Executionally, they progress from ad hoc tool-calling to workstation systems (\method{}) with persistent workspaces, skills, and governance. The "Workspace + Skill" paradigm drives this transition through state persistence, reusable procedures, and task closure. Data and evaluation shift from instruction-response pairs and static benchmarks toward State-Action-Observation trajectories and auditable, self-evolving ecosystems. Ultimately, reliable digital colleagues require persistent environments, reusable skills, and safety governance.

\setcitestyle{square,numbers,comma}
\bibliography{references}

\end{document}